\documentclass[a4paper,12pt]{article}
\usepackage[utf8]{inputenc}
\usepackage{hanging}
\usepackage{graphicx}
\usepackage{caption}
\usepackage{subfigure}
\usepackage{subfloat}
\usepackage{amsmath}
\usepackage{mathtools}
\usepackage{algorithm}
\usepackage{algorithmic}
\usepackage{wrapfig}
\usepackage{verbatim}
\usepackage{titlesec}
\usepackage{indentfirst}

\usepackage[margin=1in,textwidth=20cm,textheight=24cm,bindingoffset=0.5in]{geometry}
\begin{document}

\begin{center}
\large\textbf{scGHSOM: Hierarchical clustering and visualization of single-cell and CRISPR data using growing hierarchical SOM}

\vspace{2mm}

\textit{Shang-Jung Wen, Jia-Ming Chang and Fang Yu}

\textit{National Chengchi University. Taipei, Taiwan ROC.}
\end{center}

\begin{abstract}
High-dimensional single-cell data poses significant challenges in identifying underlying biological patterns due to the complexity and heterogeneity of cellular states. We propose a comprehensive gene-cell dependency visualization via unsupervised clustering, Growing Hierarchical Self-Organizing Map (GHSOM), specifically designed for analyzing high-dimensional single-cell data like single-cell sequencing and CRISPR screens. 

GHSOM is applied to cluster samples in a hierarchical structure such that the self-growth structure of clusters satisfies the required variations between and within. We propose a novel Significant Attributes Identification Algorithm to identify features that distinguish clusters. This algorithm pinpoints attributes with minimal variation within a cluster but substantial variation between clusters. These key attributes can then be used for targeted data retrieval and downstream analysis. Furthermore, we present two innovative visualization tools: Cluster Feature Map and Cluster Distribution Map. The Cluster Feature Map highlights the distribution of specific features across the hierarchical structure of GHSOM clusters. This allows for rapid visual assessment of cluster uniqueness based on chosen features. The Cluster Distribution Map depicts leaf clusters as circles on the GHSOM grid, with circle size reflecting cluster data size and color customizable to visualize features like cell type or other attributes.

We apply our analysis to three single-cell datasets and one CRISPR dataset (cell-gene database). We evaluate clustering methods with internal and external CH and ARI scores. GHSOM performs well and is even the best performer in internal evaluation (CH=4.2). In external evaluation, GHSOM has the third-best performance of all methods. We also improve the visualization of GHSOM clustering results with cluster feature map and cluster distribution map. These visualizations allow us to observe features of clustering results and data more efficiently and clearly.
\end{abstract}
\newpage
\tableofcontents
\newpage{}
\section{Introduction}
Artificial intelligence (AI) applications in medicine are getting more and more. It is a trend and remarkable development in the field of health and healing. A cell is the smallest and the most fundamental unit of life. However, cells are complex and diverse. The developments regarding single-cell RNA sequencing (scRNA-seq) and genome editing technologies provide critical insight to investigate cell heterogeneity and determine the cell type and the functions of an individual cell. Genome editing technology is also a rising trend. DNA has 
played an essential role in the advancement of cancer treatment, so CRISPR, a genome-editing tool, became a revolution in cancer research. Same as scRNA-seq data, genome data is mass and complex. It is a vital step to identify the target gene as a therapy to do genome editing.
\par 
The clustering technique is an important and efficient way to label and analyze data. scRNA-seq data clustering provides the identification of a cell from numerous heterogeneous cells. Because of the importance of clustering scRNA-seq data, several clustering methods have been developed such as Seurat~\cite{seurat_2015}, SC3~\cite{SC3_2017}, and CIDR~\cite{CIDR_2017}. In general, the scRNA-seq data is high-dimensional. Also, one of the structures of cell heterogeneity is the hierarchical structure which represents nested cell types or status. Most of the non-hierarchical clustering methods might lose the hierarchical information of cells. 
\par
The self-organizing map (SOM) is one of the most common and valuable neural network clustering methodologies~\cite{som_1990}. It is superior to the hierarchical clustering methods and enables the improvement of the problems of hierarchical clustering methods. However, SOM still has a few limitations, which require neuron weights to be necessary and sufficient to cluster inputs. In this research, we use a growing hierarchical self-organizing map (GHSOM)~\cite{ghsom_2000}, which can address the two limitations of SOM.
\par
The hierarchical structure of GHSOM clustering results reveals more details of high dimensional data than non-hierarchical clustering methods. We compared GHSOM's clustering performance to the other seven clustering methods, such as ACCENSE, K-means, flowMeans. GHSOM has a superb performance among them. For external evaluation, the ARI score of GHSOM is 0.88, which is the third place of the eight clustering methods (the best score is 0.9250 by PhenoGraph and the second best score is 0.9208 by flowMeans). For internal evaluation, the CH score of GHSOM is 4.2, which is the best performance of all clustering methods (the second best is is 4.1028 by DEPECHE and the third best is 3.7380 by PhenoGraph). The evaluations show that GHSOM is competitive among other clustering methods.
\par
To visualize the GHSOM result, we propose \textit{Cluster Feature Map} and \textit{Cluster Distribution Map}. A cluster feature map is well used to represent hierarchical structure and the feature that we would like to observe. A cluster distribution map is clear to observe the leaf clusters position and relations of leaf clusters. Both of the visualizations can show features such as gene expression or cell type of every leaf cluster. Therefore, we can analyze clustering results without dimension reduction techniques such as UMAP to reflect data through these visualizations.
\par
Our cases study is regarding single cells to proteins and cancer cells to genes. We also developed the \textit{Significant Attributes Identification} method to determine what attributes in clusters are the keys for clustering. We identify the attributes that has small variation within its cluster and big variation between other clusters. As a result, we expect to find out the significant genes to cancers and proteins to single cells. 
\par
We propose a comprehensive gene-cell dependency visualization via unsupervised clustering. \textit{Cluster Feature Map} and \textit{Cluster Distribution Map} show clustering results without dimension reduction, such as UMAP. The visualizations help us to observe trends and features of clustering results. From our cases study on CyTOF-Wong, CyTOF-Samusik, scRNA-seq-PBMCs and CRISPR datasets, we achieve: 1. Competitive expression map of the given genes against Seurat and given proteins against UMAP. 2. Significant genes have been reported as bio-markers. For instance, in case study scRNA-seq-PBMCs, the significant attribute LYZ and CD74 are well-known bio-markers.
\newpage
\section{Related Work}\label{section:Related work}
\subsection{K-Nearest Neighbors (KNN)}
K-Nearest Neighbors algorithm (KNN)~\cite{KNN_1992} is a non-parametric classification method and is widely used for classification and regression. A supervised machine learning algorithm needs labeled input data to learn a function, which produces the label as output when given unlabeled data. The input consists of the $k$ closest training examples in the dataset. Two types of output depending on KNN are used for classification or regression. In classification, the output would be the class label. A majority vote of its neighbors classifies an object. The object is assigned to the most common class among its $k$ nearest neighbors. In regression, the output is the property value, the average of the values of $k$ nearest neighbors. 

\subsection{K-means}
K-means~\cite{kmeans_1982} is a popular clustering method based on unsupervised learning. K-means aims to find clusters in the data, with the number of clusters represented by the variable $K$. K-means assigns each data point to one of $K$ clusters based on the provided features. It tries to make the data points in the cluster as similar as possible while also keeping the clusters as different as possible. The algorithm workflow follows: First, initialize $K$ points, centroids. Then, categorize each item to its closest centroid based on the squared Euclidean distance, and update the centroid’s coordinates, which are the averages of the items categorized in that centroid so far. Then repeat the process for a given number of iterations. 
There are some limitations of K-means, the users have to specify the number of clusters at the beginning, and K-means assumes that users deal with spherical clusters and that each cluster has roughly equal numbers. Besides, K-means does not work well on large-scale and high dimension data. Before applying the K-means algorithm, users need to deal with outliners and missing values.

\subsection{Self-Organizing Map (SOM)}
The Self-Organizing Map (SOM)~\cite{som_1990} is an artificial neural network model adhering to unsupervised learning. It is widely used as a clustering tool for mapping high-dimensional data into a two-dimensional representation space. The model consists of several neural processing elements. Each of the units $i$ is assigned an $n$-dimensional weight vector $m_i$. The main benefit of SOM is that the similarity between input data measured in the input data space is preserved within the representation space.
However, SOM has two limitations. First, SOM is the static network architecture that needs to define the number and arrangement of neural processing elements before training. Second, it has limited capabilities to represent the data hierarchical relations, which may contain critical information. Furthermore, SOM may observe hierarchical relations in a vast spectrum of application domains. Identifying the result remains a data mining task that cannot be addressed conveniently by SOM.

\subsection{Automatic classification of cellular expression by nonlinear stochastic embedding}
Automatic classification of cellular expression by nonlinear stochastic embedding (ACCENSE)~\cite{ACCENSE} is a tool that combines nonlinear dimensionality reduction with density-based partitioning and displays multivariate cellular phenotypes on a 2D plot. The high-dimensional data generated by mass cytometry are challenging to interpret in biologically meaningful ways. ACCENSE uses t-distributed stochastic neighbor embedding (t-SNE)~\cite{tsne} for embedding and seeks corresponding 2D vectors such that T cells with similar phenotypes are embedded close to each other in the map, whereas phenotypically dissimilar cells are embedded far apart. It also computes a 2D distillation that faithfully preserves neighborhood relationships present in the high-dimensional data. Then ACCENSE uses a kernel-based estimation of the 2D probability density of cells in the embedding. Local maxima in the 2D probability density corresponded to phenotypic subpopulations and were identified using a 2D peak-finding algorithm. It can automatically identify cellular subpopulations based on all proteins analyzed, thus aiding the full utilization of powerful new single-cell technologies such as mass cytometry.

\subsection{X-shift}
X-shift~\cite{X-shift} is a population finding algorithm that can process large datasets using fast KNN estimation of cell event density and automatically arranges populations by a marker-based classification system. X-shift computes the density estimate for each data point and searches for the local density maxima in a nearest-neighbor graph, which become cluster centroids. All the remaining data points are then connected to the centroids via density-ascending paths in the graph, thus forming clusters. The algorithm checks for density minima on a straight line segment between the neighboring centroids and merges them as necessary. The resolution of X-shift clustering is defined by the number ($K$) of nearest neighbors used for the density estimate. Lower $K$ values allow for resolving small and closely-positioned populations, but the result becomes increasingly affected by stochastic variations. X-shift provides a rapid and reliable approach to managed cell subset analysis that maximizes automation.

\subsection{Phenograph}
Phenograph~\cite{Phenograph} is a clustering method designed for high-dimensional single-cell data. It creates a graph ("network") representing phenotypic similarities between cells and then identifies communities in this graph. It constructs a nearest-neighbor graph to capture the phenotypic relatedness of high-dimensional data points, and then it applies the $Louvain$~\cite{louvain} graph partition algorithm to dissect the nearest-neighbor graph into phenotypically coherent subpopulations. Phenograph provides an alternative paradigm for identifying primitive cancer cells that complements the immunophenotypic definitions of cancer stem cells traditionally used in acute myeloid leukemia and other systems.

\subsection{flowMeans}
flowMeans~\cite{flowmeans} is a time-efficient and accurate method for automated identification of cell populations in flow cytometry (FCM) data based on K-means clustering. Unlike the traditional K-means, flowMeans can identify concave cell populations by modeling a single population with multiple clusters. flowMeans addresses the initialization, shape limitation, and model-selection problems of K-means clustering and can be applied to FCM data. FlowMeans uses a change point detection algorithm to determine the number of sub-populations, enabling the method to be used in high throughput FCM data analysis pipelines. flowMeans automatically chooses the number of clusters ($K$) based on a reasonable maximum by using the number of modes found individually in every eigenvector of the data. They also introduced a new merging method, which extended the flowMerge~\cite{flowmerge} approach by replacing the statistical model with a faster clustering algorithm. The merging procedure iterates between the following two steps until all of the points are merged to a single cluster: calculate/update the distance between every pair of clusters, and identify and merge the closest pair of clusters. 

\subsection{FlowSOM}
FlowSOM~\cite{flowsom} is an algorithm that speeds time to analysis and quality of clustering with Self-Organizing Maps (SOMs) that can reveal how all markers are behaving on all cells and detect subsets that might otherwise be missed. FlowSOM clusters cells (or other observations) based on chosen clustering channels and generates a SOM clustering result, then produces a Minimum Spanning Tree (MST) of the clusters and assigns each cluster to a metacluster, effectively grouping them into a population. The FlowSOM algorithm outputs SOMs and MSTs showing population abundances and marker expression in various plots including pie charts, star plots, and channel-colored plots.

\subsection{DEPECHE}
Determination of Essential Phenotypic Elements of Clusters in High-dimensional Entities (DEPECHE)~\cite{depeche} is a parameter-free and sparse $k$-means-based algorithm for clustering of multi- and mega-variate single-cell data. DEPECHE uses a penalized version of the k-means algorithm. The k-means~\cite{kmeans_1982} algorithm clusters data by fitting a mixture of normal distributions to the data with $k$ equal mixture components and unit variance. To reduce the influence of uninformative dimensions that only contribute with noise, penalized $k$-means introduces an L1-penalty for each element of each cluster center to the optimization objective. In DEPECHE, the number of clusters, $k$, is always chosen to be so large that at least one cluster is eliminated. DEPECHE simultaneously clusters and simplifies the data by identifying the variables that contribute to separate individual clusters from the rest of the data.

\subsection{Seurat}
Seurat~\cite{seurat_2015} is a R package for Quality Control (QC), analysis, and exploration of single-cell RNA-seq data. It embeds an unsupervised clustering algorithm and dimension reduction with graph-based partitioning methods. At first, Seurat selects and filtrates the cells based on the QC metrics and uses LogNormalize to normalize the feature expression measurements for each cell by the total expression, multiplies this by a scale factor (10,000 by default), and log-transforms the result. Then find a set of the high-variable genes by calculating a subset of features that exhibit high cell-to-cell variation. As a standard preprocessing step before dimensional reduction techniques, Seurat applies a linear transformation that shifts each gene's expression to make the mean expression across cells is 0, and the variance across cells is 1. Dimension reduction is made by Principle Component Analysis (PCA). At the first of the workflow, it constructs a KNN graph with the euclidean distance in PCA space and purifies the edge weights between any two cells based on the shared overlap in their local neighborhoods. Seurat next applies modularity optimization techniques to cluster the cells so as the Louvain algorithm~\cite{louvain} (default) or SLM~\cite{slm} to iteratively group cells together to optimize the standard modularity function.

\subsection{SC3}
SC3~\cite{SC3_2017} is a user-friendly tool for unsupervised clustering, which achieves high accuracy and robustness by combining multiple clustering solutions through a consensus approach. It implements consensus clustering, gene- and cell-filtering, and summarizes the probability of each pair of cells from the same cluster. QC in SC3 is implemented by calculating a filtered expression matrix to detect potentially problematic genes and cells. PCA or Laplacian graphs perform dimension reduction. Then SC3 computes the consensus matrix of the results of clustering. The consensus matrix is a binary similarity matrix in which an entry is one if two cells belong to the same cluster, otherwise zero. It is obtained by averaging the individual clustering. Hierarchical clustering infers clusters by the $k$ level of the hierarchy, where the user can supply $k$. Then it randomly select cells and use them in the clustering approach. These subpopulations train a support vector machine to infer the cluster labels of the remaining cells.

\subsection{HGC}
A classic hierarchical clustering algorithm is only suitable for small-size datasets due to the high computational complexity.
HGC~\cite{hgc_2021} is a fast hierarchical graph-based clustering method for large-scale single-cell data and is implemented in R language. The key idea is to construct a dendrogram of cells on their shared nearest neighbor (SNN) graph. It combines the advantages of graph-based clustering methods and hierarchical clustering. The workflow of HGC contains graph construction and dendrogram construction. In the graph construction step, HGC conducts PCA on the expression data, then builds the KNN and SNN graphs in the PC space. Constructing a dendrogram on the graph is a recursive procedure of finding the nearest neighbor pair and updating the graph by merging node pairs. To find the nearest neighbor pair on a graph, HGC utilizes the node pair sampling distance by Node Pair Sampling~\cite{pairSampling_2018}. HGC reveals the hierarchical structure of the data that has been mostly ignored in classic clustering methods. 

\subsection{shinyDepMap}
shinyDepMap~\cite{shinyDepMap_2021} is a web tool to enable researchers to rapidly determine the essentiality and selectivity of a given gene across cell lines and find groups of functionally related genes with similar essentiality profiles. It integrates CRISPR~\cite{CRISPR_2019} and RNAi~\cite{rnai} datasets from Depmap Portal~\cite{depmap_2017}.
\par
They performed PCA on the two datasets and computed a linear combination of the value projected on the line parallel to the eigenvector corresponding to the primary PCs. They derive two measures for each gene: the degree to which loss of the gene reduces cell growth insensitive lines (‘efficacy’) and how its essentiality varies across lines (‘selectivity’). They assessed the similarity of original data and their perturbation score by computing the \textit{Spearman Correlation Coefficients} in the three measures (CRISPR, shRNA, perturbation score). The result shows that shRNA and the perturbation score are marginally correlated, and CRISPR and the perturbation were highly correlated.
\newpage
\section{Methodology}
\subsection{Analysis Workflow}
\begin{figure}[htb]
    \centering
    \includegraphics[width=3.5in]{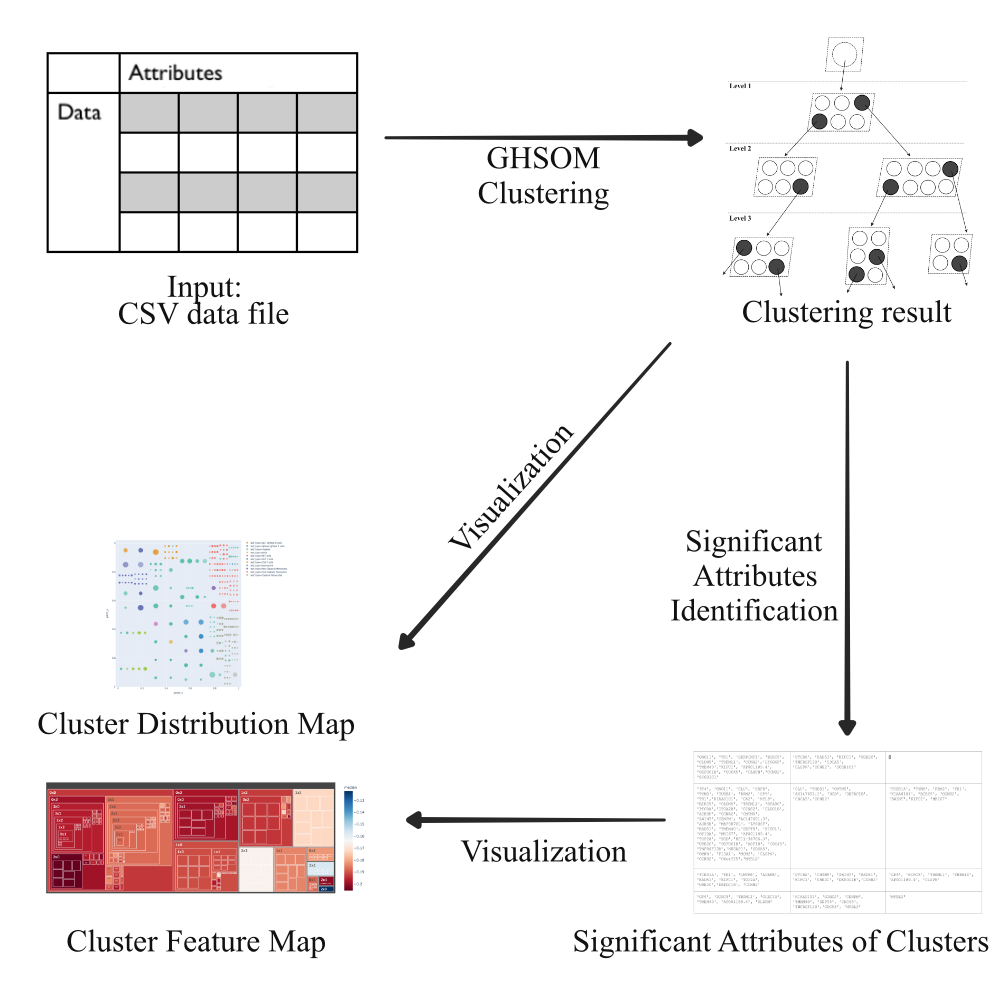}
    \caption{The workflow of our analysis.}
    \label{fig:scRNA-workflow}
\end{figure}
Our workflow contains three procedures: GHSOM clustering, Significant Attributes Identification, and Visualization Maps (Fig.~\ref{fig:scRNA-workflow}).

\subsubsection{GHSOM Clustering}
We cluster data with an unsupervised clustering data called Growing Hierarchical Self-Organizing Map (GHSOM)~\cite{ghsom_2000}. We will explain the details of GHSOM in the later chapter~\ref{section:GHSOM}. Depending on the case we are going to cluster, the input data might need to be preprocessed. We all did data preprocessing before GHSOM clustering except for CyTof case. We will discuss each of the data later in chapter~\ref{section:casestudy}.
\par
Sometimes, the number of attributes is excessive and lets GHSOM be overloaded. We will need to make dimension reductions to reduce the number of attributes. In the case PBMCs, we identified Highly Variable Genes (HVGs) as the attributes for clustering data.

\subsubsection{Significant Attributes Identification}
To know the marker genes of a cluster, we implement the \textit{Significant Attributes Identification} algorithm that is introduced in chapter \ref{section:FSF}. The marker genes are significantly practical attributes with high expressions to cluster cells. Therefore, we expect to find the vital cluster attributes and know the relation between attributes and clusters.

\subsubsection{Visualization}
We visualize the clustering result with \textit{cluster distribution map} and \textit{cluster feature map}. With a cluster distribution map, we can know how leaf clusters are relative to each other by cluster positions. Depending on what feature we are going to research and analyze, the color of the cluster distribution map can also be different features such as cell type, organ type, or expression value.
\par
Through the cluster feature map visualization, we observe the hierarchical clustering result in grids, coloring each cluster according to its value of significant attributes. By the different attributes that we use to be represented by the color on cluster maps, we can observe different situations of the clustering result. We take the median value of data to be represented by color. Then we can know which cluster has the highest or lowest median value at first sight. We also take the gene expression value to be represented. We can directly see how the clustering result is revealed with specific genes.

\subsection{Unsupervised Clustering with Growing Hierarchical Self-\\Organizing Map}\label{section:GHSOM}
\begin{figure}[htb]
    \centering
    \includegraphics[width=150pt]{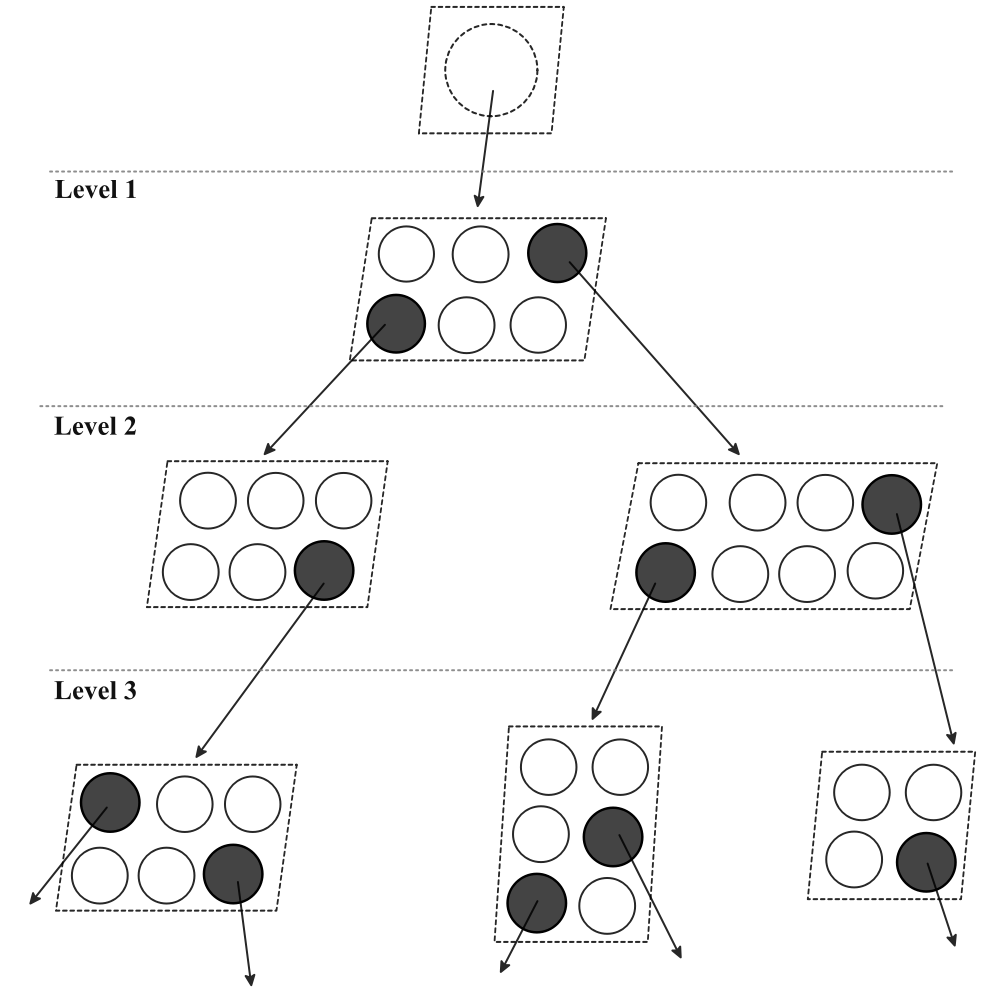}
    \caption{The structure of GHSOM}
    \label{fig:ghsom stucture}
\end{figure}

Growing Hierarchical Self-Organizing Map (GHSOM)~\cite{ghsom_2000} addresses two limitations of SOM~\cite{som_1990}. GHSOM uses a hierarchical structure of multiple layers where each layer consists of several independent SOMs (Fig. \ref{fig:ghsom stucture}). Initially, a 2x2 SOM is used as the root of the hierarchy and iteratively expanded until variation within and between clusters satisfying predefined requirements. GHSOM grows both directions: hierarchical and horizontal ways. According to the data distribution, GHSOM allows a hierarchical decomposition and navigation in sub-parts of the data. A horizontal split means that the size of each map adapts itself to the requirements of the input space. This principle is repeated with any further layers of the GHSOM. The basic idea of GHSOM is that each layer of the model is responsible for explaining some portion of the deviation of the input data as present in its previous layer.

\par
The starting point for the growth process is the overall deviation of the input data as measured with the single-unit SOM at layer 0. This unit is assigned a weight vector $w_0,$ computed as the average of all input data. Next, the deviation of the input data is computed as given in Eq.~\ref{eq:mqe0} by the mean quantization error of this single unit with $d$ that represents the number of input data $x$.
\begin{equation}
    \label{eq:mqe0}
    mqe_0 = \frac{1}{d}\cdot\parallel{w_0 - x}\parallel
\end{equation}

After the computation of $mqe_0$, the GHSOM starts with its first layer SOM, which initially consists of a small number of units, e.g., a 2 × 2 units grid, and $i$ is assigned an $n$-dimensional weight vector $w_i$. \\
The learning process of SOMs is similar to a competition among the units to represent the input patterns. The unit with the weight vector being closest to the presented input pattern in terms of the input space wins the competition. The weight vector of the winner and units in the vicinity of the winner are adapted in such a way as to resemble more closely the input pattern. The degree of adaptation is guided by means of a learning-rate parameter $\alpha$, decreasing in time. The number of units that are subject to adaptation also decreases in time such that at the beginning of the learning process a large number of units around the winner is adapted, whereas towards the end only the winner is adapted. These units are chosen by means of a neighborhood function $h_{ci}$ which is based on the units' distances to the winner as measured in the two-dimensional grid formed by the neural network. In combining these principles of SOM training, the learning rule is given in Eq.~\ref{eq:learning}, where $x$ represents the current input pattern, and $c$ refers to the winner at iteration $t$.
\begin{equation}
    \label{eq:learning}
    w_{i}(t+1)=w_{i}(t)+\alpha(t) \cdot h_{ci}(t) \cdot [x(t)-w_{i}(t)]
\end{equation}

In order to adapt the size of this first layer SOM, the mean quantization error of the map is computed ever after a fixed number $\lambda$ of training iterations as given in Eq.~\ref{eq:mqe}. We will refer to the \textit{mean quantization error} of a map as $MQE$ in upper case letters.

\begin{equation}
    \label{eq:mqe}
    MQE_m = \frac{1}{u}\cdot\sum_i mqe_i
\end{equation}

The SOM on each layer is allowed to grow until the deviation present in the unit of its preceding layer is reduced to at least a fixed percentage. For growing each SOM horizontally, each SOM grows until the mean value of the MQEs for all nodes on the SOM is smaller than the $mqe$ of the parent node multiplied by the parameter $\tau_1$. If the stop criterion is not satisfied, we find the error node that owns the largest $mqe$ and inserts one row or column of new nodes between the error node and its dissimilar neighbor.
\par
For expanding the hierarchical structure, the node with its $mqe_i$ greater than $\tau_2$ times $mqe$ on the last level will be used to develop the next SOM layer. The smaller the parameter $\tau_1$ or $\tau_2$ is chosen, the larger the emerging SOM will be. We implement GHSOM by the GHSOM toolbox~\cite{vienna_ghsom}. The segment of the GHSOM algorithm is shown in Algorithm \ref{algo:ghsom}~\cite{ghsom_algo}. By adjusting the values of breadth($\tau_1$) and depth($\tau_2$), we derive different shapes of GHSOMs and discover the most suitable one. 
\par 
The original visualization of GHSOM result is a table. We named clusters as row and column numbers of clusters, such as 0x0, 1x0. If the cluster is at a deeper level than L1 (L1), we will name it in the ascending order of the cluster name of levels. For example, the cluster is at L1 0x0 and L2 2x1. Then we name it 0x0-2x1.
\newpage
\begin{algorithm}
\caption{GHSOM Algorithm}
\label{algo:ghsom}
\begin{algorithmic}[1]
\STATE{Initialize the parameters $\tau_1$ and $\tau_2$}
\STATE{Initialize the set of Layer $L = \{l_0\}$}
\STATE{Initialize the set of maps $M = \{m_0\}$}
\STATE{Compute the initial mean quantization error $mqe_0$}
\FOR{$l$ in $L$}
\FOR{$m$ in $M$}
\REPEAT
\STATE{Train the map $m$ as a single SOM}
\STATE{Compute $mqe_i$ and $MQE_m$}
\IF{$MQE_m \geq \tau_1 \cdot mqe_u$}
\STATE{Select a neuron $e$ with the highest $mqe$ and its most dissimilar neighbor $d$.}
\STATE{Insert a row or column of neurons between $d$ and $e$ and initialize their weight vectors as the means of their respective neighbors.}
\ENDIF
\UNTIL{$MQE_m < \tau_1 \cdot mqe_u$}
\FOR{$w_i$ in $W$}
\IF{$mqe_i \geq \tau_2 \cdot mqe_0$}
\STATE{Create a new map in the lower layer from the neuron $i$}
\ENDIF
\ENDFOR
\ENDFOR
\ENDFOR
\end{algorithmic}
\end{algorithm}

\subsection{Significant Attributes Identification}\label{section:FSF}
After GHSOM clustering, we would like to analyze the clustered result. Since GHSOM is an unsupervised clustering method,  we try to figure out the significant attributes that affect the clustering result most. GHSOM shows that the relative and similar elements are clustered in one cluster. Therefore, the significant attributes are supposed to have similar values of data within cluster and dissimilar values between clusters. To figure out how similar significant attributes in the cluster are, we adopt the idea of Within-Group Sum of Squares (WGSS) and Between-Group Sum of Squares (BGSS) and refine them in attribute level. We calculated $\sigma_I(\textbf{c},\textbf{g})$ that is the standard deviation of the attributes \textbf{g} across data in the cluster \textbf{c}, and it is defined as
\begin{equation}
    \label{eq:sigmaI}
    \sigma_I(\textbf{c},\textbf{g}) = \sqrt{\frac{1}{|\textbf{c}|}\sum\limits_{data \in \textbf{c}} (data[\textbf{g}] - m^\textbf{c}_\textbf{g})^2}
\end{equation}
, where the $m^\textbf{c}_\textbf{g}$ is the mean of attribute $\textbf{g}$ across data in the cluster \textbf{c}, and $data_\textbf{g}$ is the data value of attribute \textbf{g}. Meanwhile, to know how different the attributes are in other clusters from the cluster \textbf{c}, we calculated $\sigma_B$ that is the standard deviation of the attributes \textbf{g}'s mean within every cluster, and it is defined as
\begin{equation}
    \label{eq:sigmaB}
    \sigma_B(\textbf{c},\textbf{g}) =  \sqrt{\frac{1}{|\textbf{C}|}\sum\limits_{\textbf{c'} \in  \textbf{C}}(m^\textbf{c}_\textbf{g} - m^\textbf{c'}_\textbf{g})^2}
\end{equation}
\par
In order to minimize the standard deviation $\sigma_I(\textbf{c},\textbf{g})$ inside the cluster and maximize the standard deviation $\sigma_B(\textbf{c},\textbf{g})$ between clusters, we implement with Equation \ref{equation:fg}, which takes the difference of $\sigma_B(\textbf{c},\textbf{g})$ and $\sigma_I(\textbf{c},\textbf{g})$. The larger $\text{diff}(\textbf{c},\textbf{g})$ means the larger $\sigma_B(\textbf{c},\textbf{g})$ and the smaller $\sigma_I(\textbf{c},\textbf{g})$. We consider the attributes with the top10 $\text{diff}(\textbf{c},\textbf{g})$ as the significant attributes($F^\textbf{c}$) for the cluster \textbf{c}.
\begin{equation}
    \text{diff}(\textbf{c},\textbf{g}) = \sigma_B(\textbf{c},\textbf{g}) - \sigma_I(\textbf{c},\textbf{g})
    \label{equation:fg}
\end{equation}
\begin{align*}
    F^\textbf{c} &= \{ fg \mid fg \in  \text{attributes have top 10  } \text{diff}(\textbf{c},\textbf{g}) for~\textbf{c}\}
\end{align*}
\par
The Significant Attributes Identification works as Algorithm \ref{algo:SFI}. Initially, we have the data of every clusters and the target cluster in dataframe type. Directly compute $m^\textbf{c}_\textbf{g}$ and $\sigma_I(\textbf{c},\textbf{g})$ in every cluster \textbf{c} inside dataframes (line 2~3). Then compute $\sigma_B(\textbf{c},\textbf{g})$ and $\text{diff}(\textbf{c},\textbf{g})$ with loops (line 6~14). Eventually, we take top10 significant attributes after sorting data by $\text{diff}(\textbf{c},\textbf{g})$.
\newpage
\begin{algorithm}[htb]
\caption{Significant Attributes Identification Algorithm}
\label{algo:SFI}
\begin{algorithmic}[1]
\STATE{Target cluster $\textbf{c}$ from which we are going to identify significant attributes $\in$ all cluster $\textbf{C}$ as input}
\STATE{Compute all $m^\textbf{c}_\textbf{g} = $ the mean value of each attribute $\textbf{g}$ across all data in cluster $c$ }
\STATE{Compute all $\sigma_I(\textbf{c},\textbf{g}) = $ the standard deviation value of each attribute $\textbf{g}$ across all data in cluster $\textbf{c}$ }
\FOR{$g$ in all attributes}
\STATE{$\sigma_B(\textbf{c},\textbf{g}) = 0$}
\FOR{$\textbf{c'}$ in all cluster $\textbf{C}$}
\IF{$\textbf{c'} \neq $ target cluster $\textbf{c}$}
\STATE{$\sigma_B(\textbf{c},g) += (m^\textbf{c}_\textbf{g} - m^\textbf{c'}_\textbf{g})^2$}
\ENDIF
\STATE{$\sigma_B(\textbf{c},\textbf{g}) = \sqrt{\sigma_B(\textbf{c},\textbf{g})\div\text{(len(\textbf{C})-1)}}$}
\ENDFOR
\ENDFOR
\FOR{$\textbf{g}$ in all attributes}
\STATE{diff(\textbf{c},\textbf{g}) $= \sigma_B(\textbf{c},\textbf{g}) - \sigma_I(\textbf{c},\textbf{g})$}
\ENDFOR
\STATE{diff\_arr $=$ Descending sorted diff values}
\STATE{$F^\textbf{c} = $ top10 values in diff\_arr}
\end{algorithmic}
\end{algorithm}

\subsection{The Visualization}
We propose two visualizations to present the result of GHSOM. To improve the current visualization of GHSOM, we provide \textit{Cluster Feature Map}, which can show the hierarchical structure of results at one sight. The color of it represents a feature that we want to observe, such as the value of a certain attribute.
\par
We want to visualize clustering results without dimension reduction and reveal the spatial relations between clusters. Therefore, we provide the second visualization, \textit{Cluster Distribution Map}. The color of it can be any feature we want to know, such as data types or the value of an attribute.
\subsubsection{Cluster feature map}
The result of GHSOM is a hierarchical structure that is hard to be analyzed. GHSOM tool presents the result of clustering with a table that contains the data in clusters~\cite{vienna_ghsom}. They show the presentation with an HTML file. If the cluster has not only one level, it will show the link of the next level HTML file. However, the presentation is hard to explore the result after L2 without click in the first sight. It only simply presents what data is in clusters without any feature. To improve that, we visualize the result of GHSOM with 
treemap structure~\cite{SquarifiedTreemap_2000} and name it \textit{Cluster Feature Map}. Cluster feature map is an efficient and compact display that efficiently shows the size of the final elements in the structure. The cluster feature map uses nested rectangles to show hierarchical data. We expected to directly see how all levels of the result are with a plot. On the cluster feature map, we can see the next-level clusters in clusters. The outermost rectangles are the L1 clusters and contain the following levels of clusters.
\par
\begin{figure}[htb]
    \centering
    \includegraphics[width=5in]{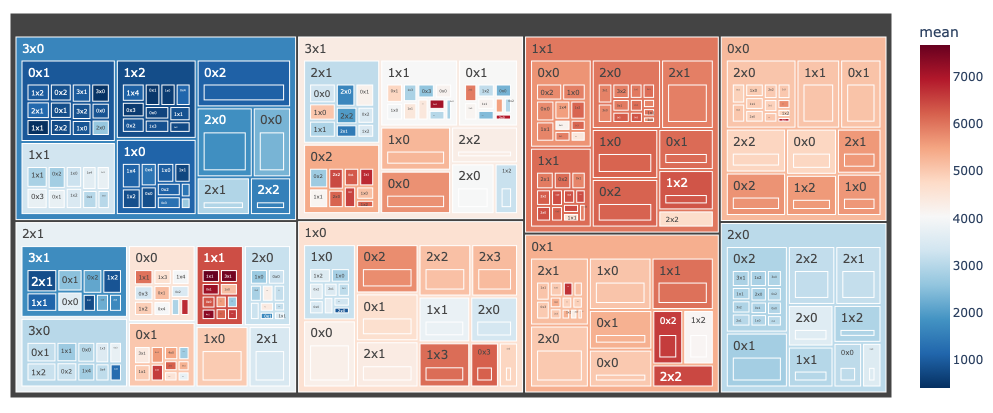}
    \caption{An example of cluster feature map. The cluster feature map of Wong dataset. The color represents the arithmetic mean of data in continuous color.}
    \label{fig:treemap}
\end{figure}

The cluster feature map visualization is shown as Fig.~\ref{fig:treemap}. A rectangle represents a cluster (in different levels). The size of a rectangle represents the number of data in that cluster. The bigger the rectangle is, the more data is in that cluster. The color of a rectangle represents the mean value of cells in that cluster. The (feature) value is defined as the bar on the right side. The feature that we would like to observe among clusters can be freely defined with continuous color. For instance, in Fig.~\ref{fig:treemap}, we define mean (the average value of genes of a cell) as the feature. We could observe that L1 3x0 (deep blue) and L1 1x1 (red) have significant difference on this feature. Within L1 3x1, the inner clusters L2 3x1-2x1 and L2 3x1-0x2 still have different colors. We implement the cluster feature map structure with python package \textit{Plotly}~\cite{plotly}. It is a dynamic cluster feature map structure. By clicking the box of the cluster on the cluster feature map, it will show the next level of the cluster.

\subsubsection{Cluster distribution map}
The position of a cluster is related to the relationship between the clusters. The closer positions mean that the clusters are more similar and relative. Although a cluster feature map can show hierarchical data with one plot, it automatically puts the rectangles as the larger ones in the upper left corner and the smaller ones in the lower right corner. Therefore, we can not know the actual positions of clusters in the GHSOM result through cluster feature map visualization.
We developed a cluster distribution map that can reflect the positions of leaf clusters. To map clusters, we have to get the coordinates of every cluster. 
\begin{figure}[htb]
    \centering
    \includegraphics[width=3in]{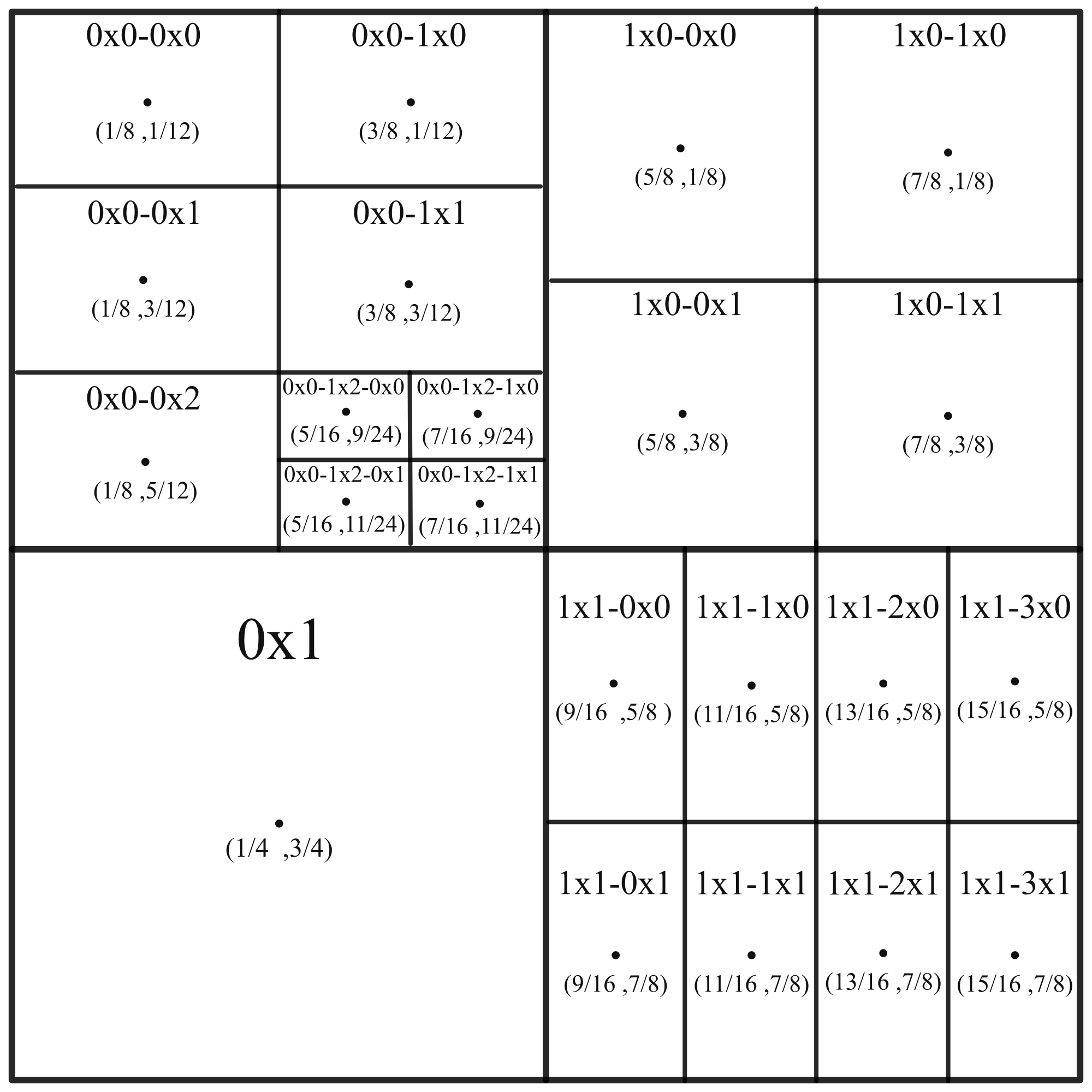}
    \caption{Coordinates of GHSOM result}
    \label{fig:coordinate}
\end{figure}
Because the structure of GHSOM result is hierarchical, we squeeze all levels into a flat level and calculate the two-dimensional coordinate for every leaf cluster, such as Fig.~\ref{fig:coordinate}. 
Eq\ref{equation:coor_wh} and Eq\ref{equation:point} show how we calculate the center coordinates of a leaf cluster of GHSOM clustering result. row$_i$(col$_i$) refers to the number of rows(columns) on level $i$ where the target $w_i$($h_i$) refers to the width(height) of square that presents the target cluster on $i$-th level. 
\begin{equation}
    w_i = w_{i-1} \times \frac{1}{\text{col}_i} \text{\qquad;\quad}
    h_i = h_{i-1} \times \frac{1}{\text{row}_i} \text{\quad}(w_0 =1, h_0 = 1)
    \label{equation:coor_wh}
\end{equation}
\par
After knowing the widths and heights of the squares on each level, we can calculate the center point of the target leaf cluster as Eq.~\ref{equation:point}. $l$ refers to the number of levels of the target cluster. $X_i$ refers to the $i$-th column at which the target cluster is, and $Y_i$ refers to the $i$-th row at which the target cluster is. We divide the width and height of the square on the last level by two to get the center point of the square.  Then, at the end of the equation, we add $w_l$(or $h_l$) $\times \frac{1}{2}$. 
\begin{equation}
    P_x = \sum_{i=1}^l (w_i \times X_i) + w_l \times \frac{1}{2} \text{\quad;\quad} P_y = \sum_{i=1}^l (h_i \times Y_i) + h_l \times \frac{1}{2}
    \label{equation:point}
\end{equation}
\par 
We use the coordinates of every leaf cluster to map them on a cluster distribution map. We take the cluster 0x0-1x1 in Fig.~\ref{fig:coordinate} as an example. We calculate the X coordinate $P_x$ first. By Eq.~\ref{equation:coor_wh},  $w_1$ equals $1 \times \frac{1}{2}$, and the result is $\frac{1}{2}$. $w_2$ equals $\frac{1}{2} \times \frac{1}{2}$, and the result is $\frac{1}{4}$. Then we sum $w_0 \times 0$ and $w_1 \times 1$, and the result is $0 + \frac{1}{4}$, which is $\frac{1}{4}$. And then add $w_2 \times \frac{1}{2}$ to the last step result $\frac{1}{4}$, which equals $\frac{1}{4} + \frac{1}{4} \times \frac{1}{2}$. Then we get $P_x$ that is $\frac{3}{8}$. We do the same math routine to get $P_y$ that is $\frac{3}{12}$. 

\par
\begin{figure}[htb]
    \centering
    \includegraphics[width=5in]{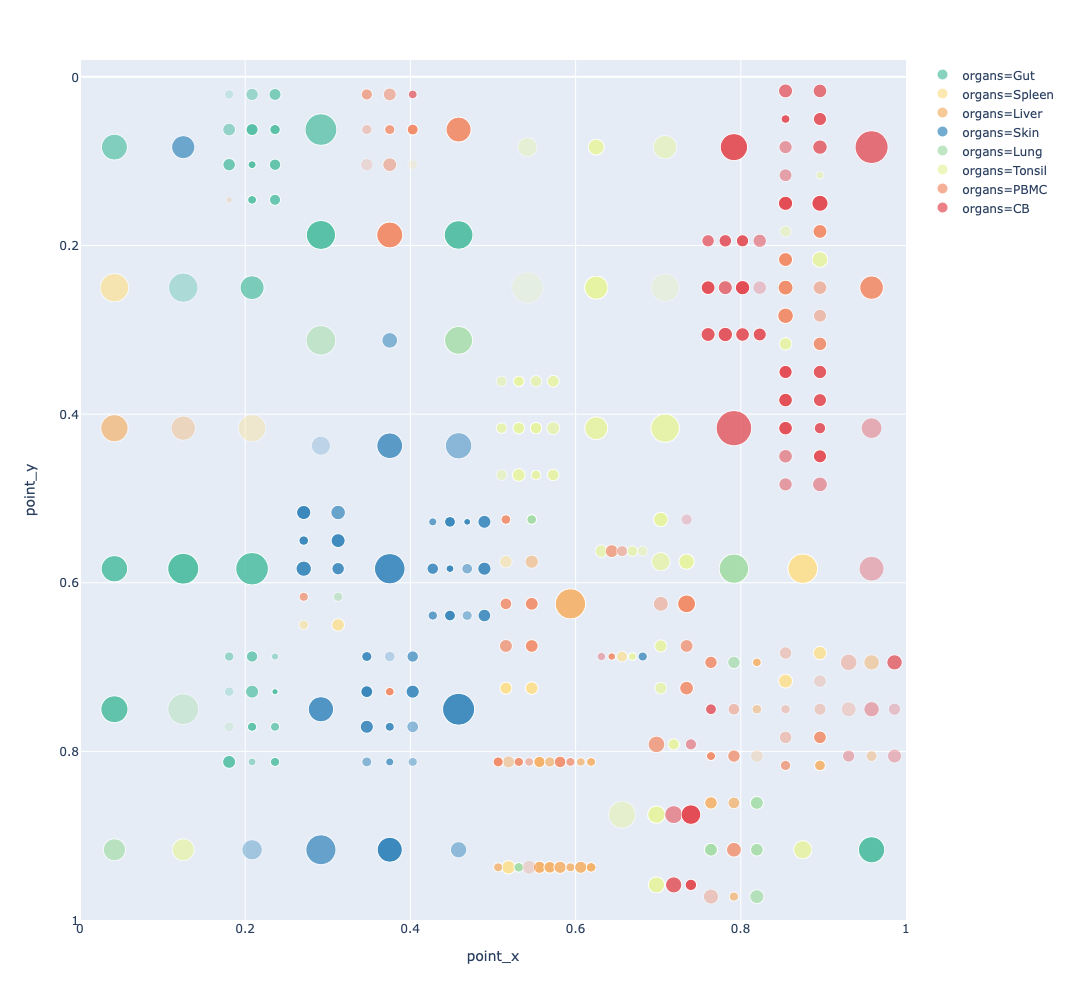}
    \caption{An example of cluster distribution map. The cluster distribution map of Wong dataset. The color represents the cell types of data.}
    \label{fig:bubblemap}
\end{figure}

The cluster feature map looks as Fig.~\ref{fig:bubblemap}. The size of bubbles represents the number of data in clusters. The color of a bubble can also represent features of data. It can show as continuous or discrete color. As a continuous color, it can represent any values that we would like to observe in each cluster; as discrete color, it can represent "cell type" or any labels. While it presents as discrete color, the opacity represents how pure the label is by counting the number of elements under this label divide number of all elements. 

\newpage
\section{GHSOM threshold selection} \label{section:threshold}
There are two parameters of GHSOM: $\tau_1$ and $\tau_2$ that specify thresholds on the variation of clusters between and within respectively. 
$\tau_1$ and $\tau_2$ respectively control the horizontal and hierarchical size of the resulting map. The smaller the parameter $\tau_1$ or $\tau_2$ is chosen, the larger the emerging SOM will be. There are no standard $\tau_1$ and $\tau_2$ to have the best clustering result. Therefore, we have to try different sets of $\tau_1$ and $\tau_2$ to figure out which of them has the best clustering result.
\par
To define a good clustering result, we need to evaluate the accuracy of the result. Because unsupervised clustering does not have truth labels to check the accuracy of the result, we use internal and external evaluations to evaluate the result \cite{comparisonFramework_2019}. Internal evaluation doesn't evaluate the clustering result based on input labels. We use is Calinski Harabasz Index (CH), which is a measure of how similar an object is to its own cluster compared to other clusters \cite{CH}. It evaluates the model when truth labels are not known where the validation of how well the clustering has been done is made using quantities and features inherent to the dataset.
\par

The CH index for $K$ number of clusters on a dataset $D$ = [ $d_1$ , $d_2$ , $d_3$ , … $d_N$ ] is defined as,
\begin{equation}
    \text{CH} = \left[ \frac{\sum_{k=1}^K n_k\left \| c_k-c \right \|^2}{K-1}\right] / \left[ \frac{\sum_{k=1}^K \sum_{i=1}^{n_k}\left \| d_i-c_k\right \|^2}{N-K}\right]
    \label{equation:CH}
\end{equation}
, where $n_k$ and $c_k$ are the number of points and centroid of the $k^{th}$ cluster respectively, $c$ is the global centroid, $N$ is the total number of data points. Higher value of CH index means the clusters are dense and well separated.

\begin{figure}[htb]
	Aa	[.*]	W

previousnext
0 of 0
Replace
Replace All

    \centering
    \includegraphics[width=3.5in]{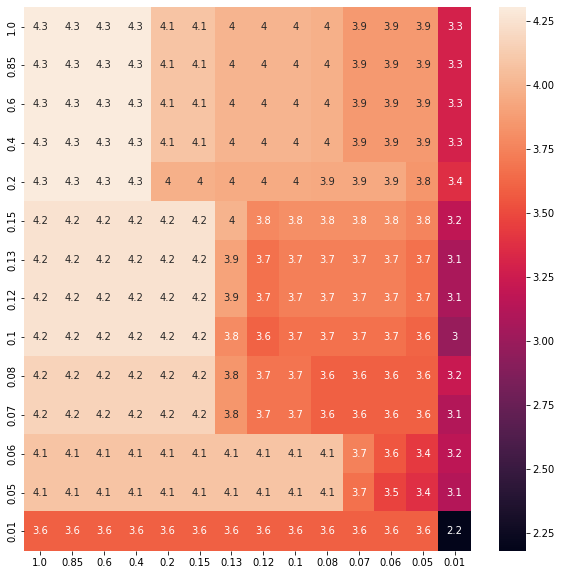}
    \caption{The CH value heatmap of different $\tau_1$ and $\tau_2$.}
    \label{fig:ch_heat}
\end{figure}
\par
Besides the internal evaluation that we just mentioned, we also evaluate the performance with external evaluation. External evaluation evaluates clustering result with the true labels. The external evaluation is Adjusted Rand Index (ARI), which measures the similarity between a clustering result and the true labels \cite{ARI}.

\begin{equation}
    \text{ARI($P^*$,$P$)} = \frac{\sum_{i,j} \binom{N_{ij}}{2} - \left[ \sum_i \binom{N_i}{2}\sum_{j} \binom{N_j}{2}\right] /\binom{N}{2}}{\frac{1}{2}\left[ \sum_i \binom{N_i}{2}+\sum_j \binom{N_j}{2} \right] - \left[ \sum_i \binom{N_i}{2}\sum_j \binom{N_j}{2} \right] / \binom{N}{2}}
    \label{equation:ARI}
\end{equation}
Where $N$ is the number of data points in a given data set and $N_{i,j}$ is the number of data points of the class label $C^*_j \in P^*$ assigned to cluster $C_i$ in partition $P$. $N_i$ is the number of data points in the cluster $C_i$ of partition $P$, and $N_j$ is the number of data points in class $C^*_j$. In general, an ARI value is between 0 and 1. The index value is equal to 1 only if a partition is completely identical to the intrinsic structure and close to 0 for a random partition.
\begin{figure}[htb]
    \centering
    \includegraphics[width=3.5in]{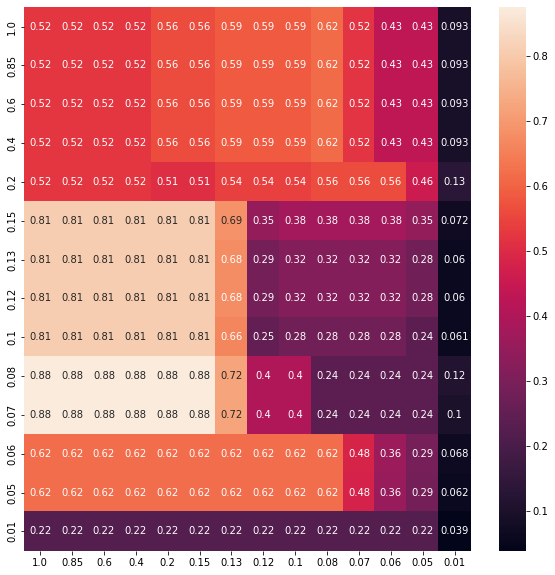}
    \caption{The ARI value heatmap of different $\tau_1$ and $\tau_2$.}
    \label{fig:ari_heat}
\end{figure}
\par
In this section, we used Samusik dataset for an example. Samusick dataset is a CyTOF dataset, and we also discuss the whole case study of it in the later chapter~\ref{section:samusik}. From Fig.~\ref{fig:ch_heat}, we noticed that the CH scores gradually decrease from higher $\tau_1$ and $\tau_2$ to lower $\tau_1$ and $\tau_2$. We also noticed that the scores drops while $\tau_2$ is lower than 0.15.
In Fig.~\ref{fig:ari_heat}, we found that the best ARI score appears while $\tau_1$ is 0.08 or 0.07 and $\tau_2$ is between 1.0 and 0.15, and then the score starts to drop when $\tau_2$ is lower than 0.15. Therefore, we considered both CH and ARI performance and found that the best clustering result appears while $\tau_1$ equals 0.08 and $\tau_2$ equals 0.15.
\par
To decide the best $\tau_1$ and $\tau_2$, this threshold selection process is included in the clustering process. We try different sets of $\tau_1$ and $\tau_2$ and choose the pair with the best clustering performance through the CH and the ARI evaluations.
 
\section{Case study}\label{section:casestudy}

\subsection{CyTOF-Wong dataset}
By using GHSOM, we expect that we can analyze and visualize our results without dimension reduction techniques. There are many dimension reduction methods, such as UMAP and tSNE. In this case, we would like to compare our methodology with UMAP as dimension reduction. We did the experiment on the Mass-cytometry (CyTOF) dataset —— Wong dataset which is one of the datasets that \emph{\textbf{Dimensionality reduction for visualizing single-cell data using UMAP}}~\cite{2019DimensionalityReduction} uses.
Mass-cytometry (CyTOF) is a technique based on inductively coupled plasma mass spectrometry and time of flight mass spectrometry used for the determination of the properties of cells (cytometry). It is also a variation of flow cytometry in which antibodies are labeled with heavy metal ion tags.
Wong dataset contains 327,457 single cells in eight cell types including CB, PBMC, Liver, Spleen, Tonsil, Lung, Gut, and Skin. There are 39 proteins as analyzed parameters. We cluster single-cell data with 39 proteins as attributes.
\begin{figure}[htb]
    \centering
    \includegraphics[width=5in]{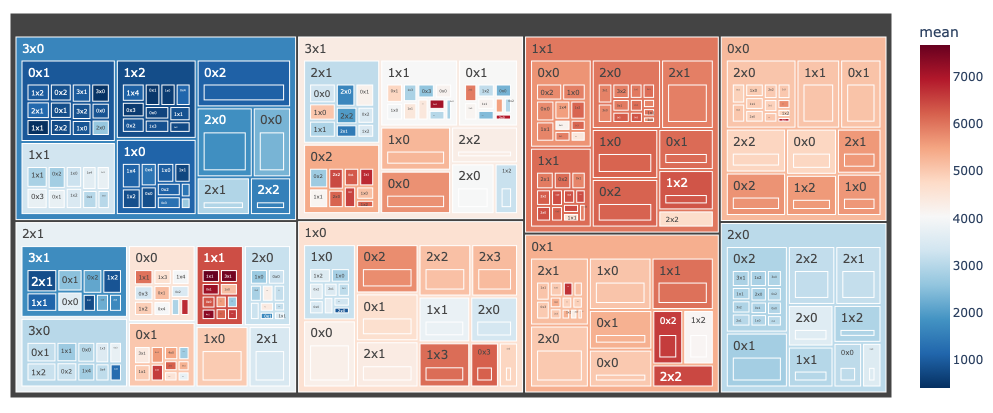}
    \caption{The cluster feature map with color in the mean value of attributes on CyTOF-Wong dataset.}
    \label{fig:wong_treemap}
\end{figure}
\subsubsection{GHSOM clustering}
The parameter setting of this case is $\tau_1$=0.1 and $\tau_2$=0.01.
The GHSOM clustering result is visualized with a cluster feature map as Fig.~\ref{fig:wong_treemap}. The clustering result is four levels in depth and has eight clusters on the L1 result. The feature here is the mean value of data. The color represents the average cell mean across attributes. We can observe which cluster has a high value or low value through the cluster feature map and quickly get the highest or lowest clusters, which are more special. Cluster 2x1-1x1 has the highest value, and cluster 3x0 has leaf clusters that have relatively low values.
\subsubsection{Gene expression}
\begin{figure}[htb]
    \centering
    \subfigure[CD103]{
        \begin{minipage}[t]{0.3\linewidth}
            \includegraphics[width=1.8in]{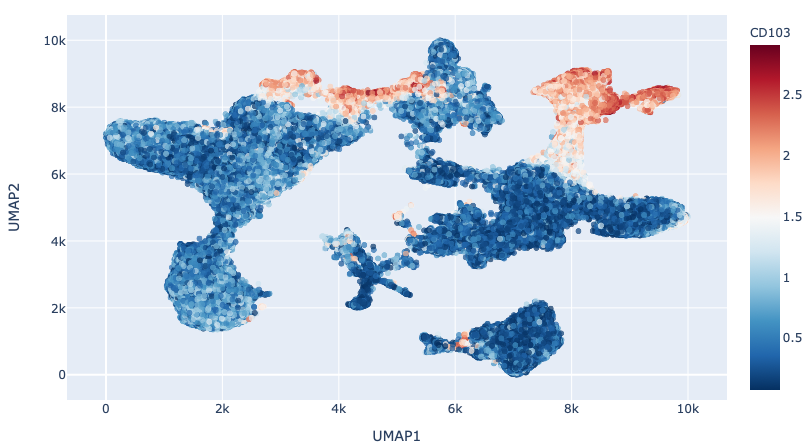}\\
            \vspace{0.02cm}         
            \includegraphics[width=1.8in]{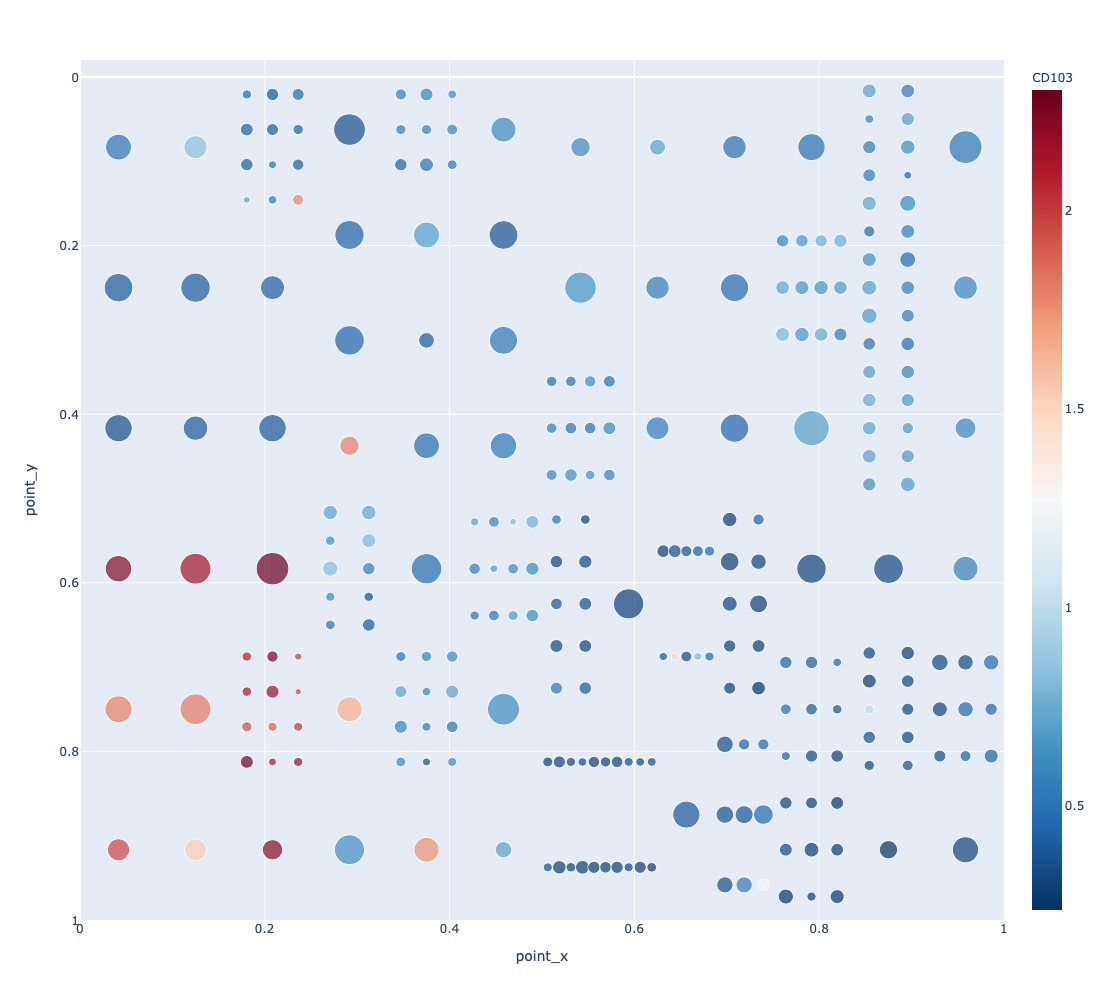}
        \end{minipage}%
    }%
    \subfigure[CD45RO]{
        \begin{minipage}[t]{0.3\linewidth}
            \includegraphics[width=1.8in]{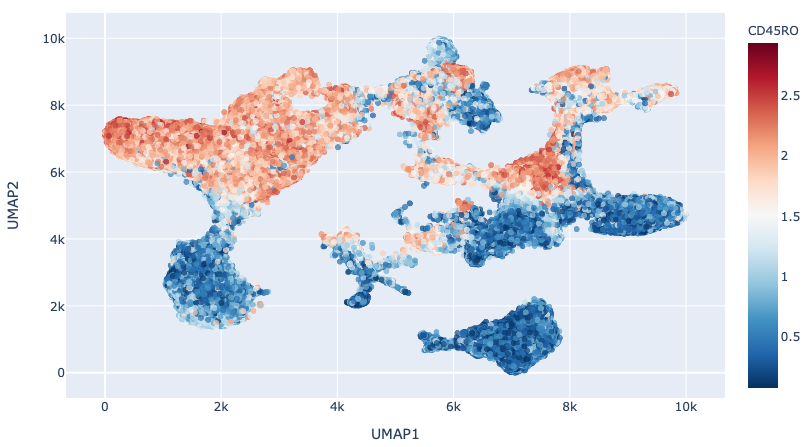}\\
            \vspace{0.02cm}         
            \includegraphics[width=1.8in]{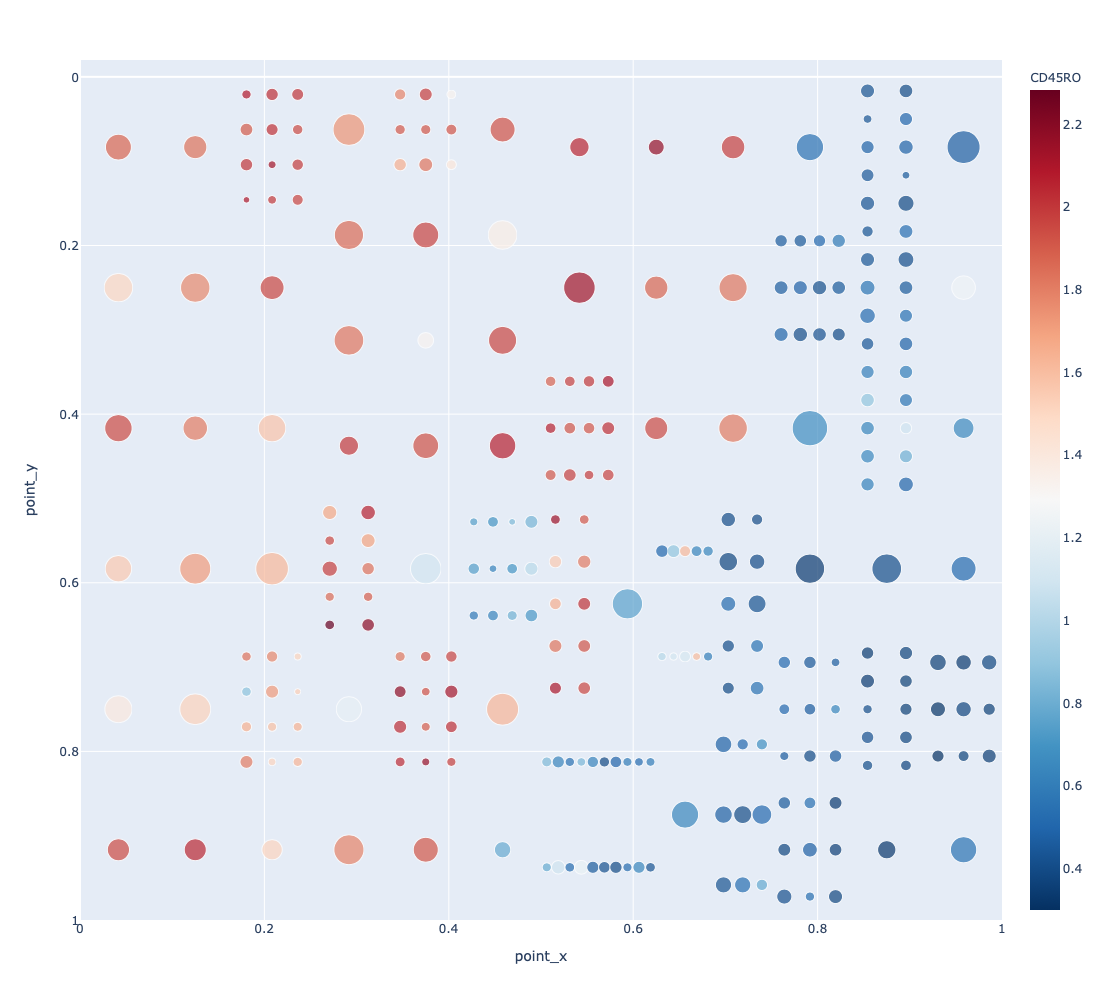}
        \end{minipage}%
    }%
    \subfigure[CCR7]{
        \begin{minipage}[t]{0.3\linewidth}
            \includegraphics[width=1.8in]{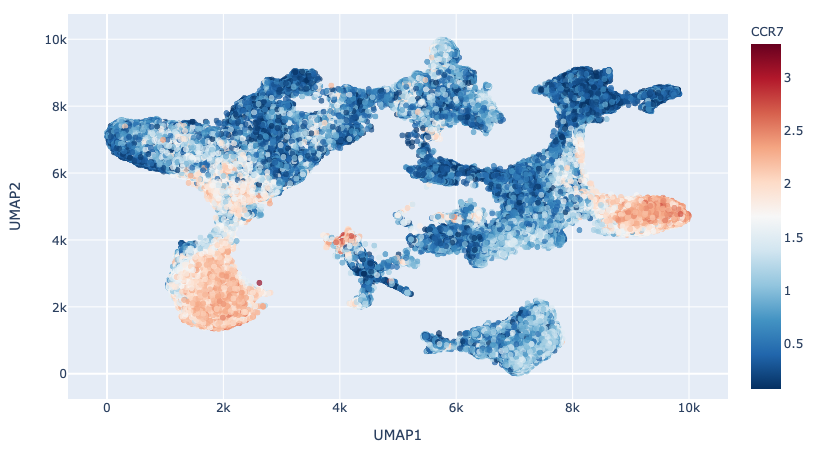}\\
            \vspace{0.02cm}         
            \includegraphics[width=1.8in]{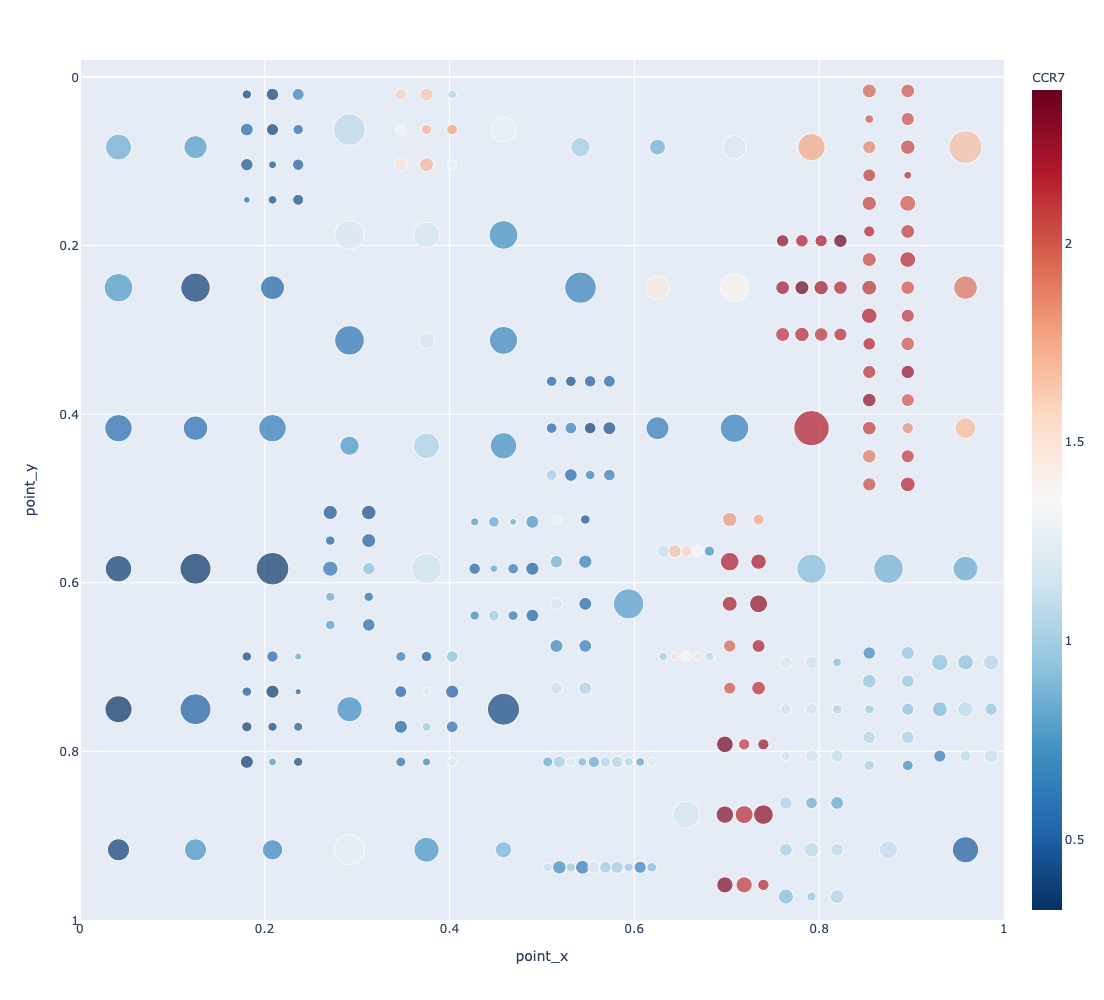}
        \end{minipage}%
    }
    \caption{Cell expression with UMAP and cluster distribution maps on attributes CD103, CD45RO and CCR7.}
    \label{fig:3gene-expression}
\end{figure}

In the relative work, they had seen the expression levels of events for the resident-memory T cell marker CD103, the memory T cell marker CD45RO, and the naive T cell marker CCR7 on UMAP projection and observed that UAMP was able to recap the differentiation stage of T cells within each major cluster. 
We projected those three genes on cluster distribution maps (Fig.~\ref{fig:3gene-expression}) where the upper graphs are gene expressions shown by UMAP, and our cluster distribution map shows the lower ones. Clusters, with red color corresponding to higher genes expressions values, are close to each other. We confirmed that clusters are distinguished by gene expressions. 

\subsubsection{Significant Attributes Identification}
We picked five relative special leaf clusters from the cluster feature map. 3x0-0x1-1x1, is the cluster with the lowest mean value; 2x1-1x1-1x1, is the cluster with the highest mean value. We randomly picked three clusters with ordinary mean values, which are cluster 1x0-1x0-2x0, cluster 1x1-1x2, and cluster 2x0-0x1.
We identified the significant attributes of these five leaf clusters with the Significant Attributes Identification method (Table.~\ref{table:sf-wongleaf}).
\par
We noticed that certain attributes are identified as significant in different clusters, CD45RO ,CD45RA are identified in four out of five leaf clusters that we picked, and CD8 is identified in all of them. They might be the significant attributes to the most clusters, and that shows they are important and impact the clustering result.
\linespread{1.2}
\begin{table}[htb]
\centering
\begin{tabular}{|c|p{330pt}|} 
\hline 
leaf cluster & Significant attributes \\
\hline 
3x0-0x1-1x1 & CD4, CD62L, CCR7, CD27, CD45RO, CD8, ICOS, CD38, CD127, CD45RA\\
\hline
2x1-1x1-1x1 & CD4, CD8, CD45RO, CD45RA, CD62L, CLA, CD127, IntegrinB7, CD103, CXCR5\\
\hline
1x0-1x0-2x0 & CD4, CCR4, CD8, CD69, CD45RA, CCR5, CD25, CD127, ICOS, CLA\\
\hline
1x1-1x2 & CLA, CD103, IntegrinB7, CD69, CD45RO, CD62L, CD95, CD8, CD27, CD56\\
\hline
2x0-0x1 & CXCR5, PD-1, CD4, ICOS, CD45RO, CD69, CD27, CD8, CD45RA, CD95\\
\hline
\end{tabular}
\caption{Significant attributes of leaf clusters in CyTOF-Wong dataset.}
\label{table:sf-wongleaf}
\end{table}

\subsubsection{Cluster feature map}
\begin{figure}[htb]
    \centering
    \subfigure[leaf 1x0-1x0-2x0] {
        \includegraphics[width=0.3\textwidth]{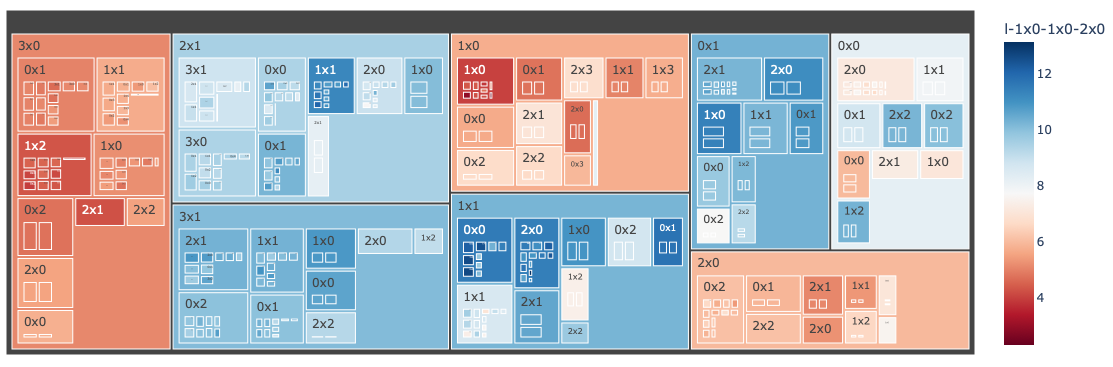}
        \label{fig:w-l1x0-1x0-2x0}
    }
    \subfigure[leaf 1x1-1x2] {
        \includegraphics[width=0.3\textwidth]{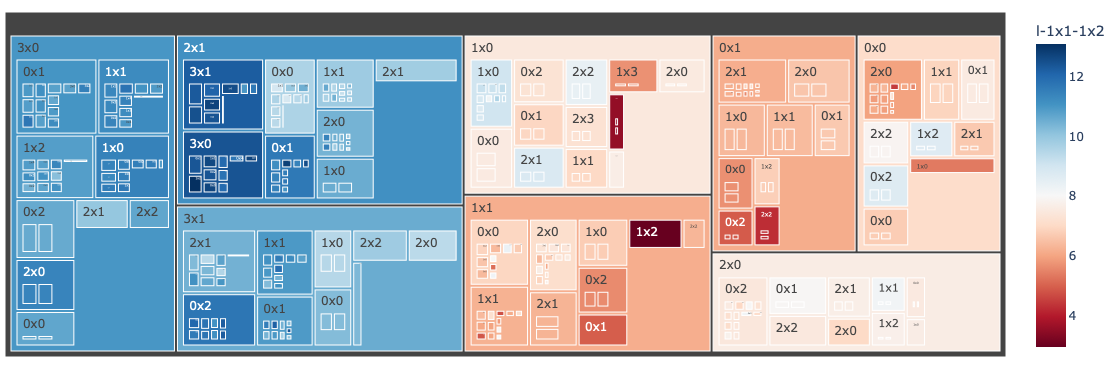}
        \label{fig:w-l1x1-1x2}
    }
    \subfigure[leaf 2x0-0x1] {
        \includegraphics[width=0.3\textwidth]{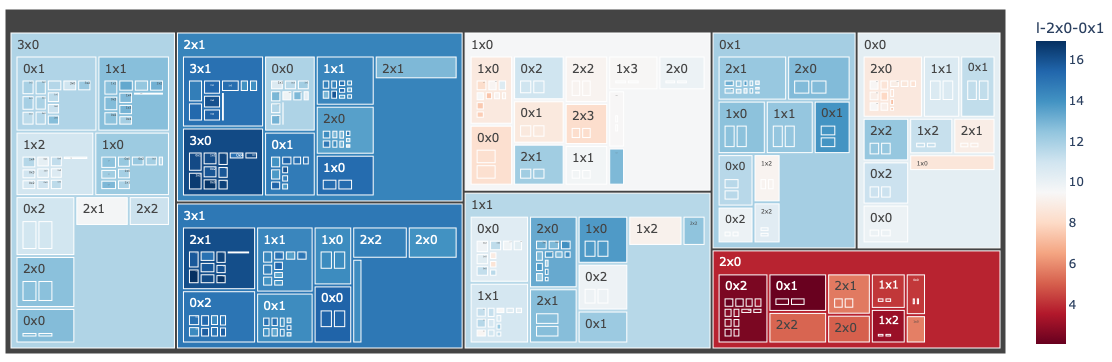}
        \label{fig:w-l2x0-0x1}
    }
    \subfigure[leaf 2x1-1x1-1x1] {
        \includegraphics[width=0.3\textwidth]{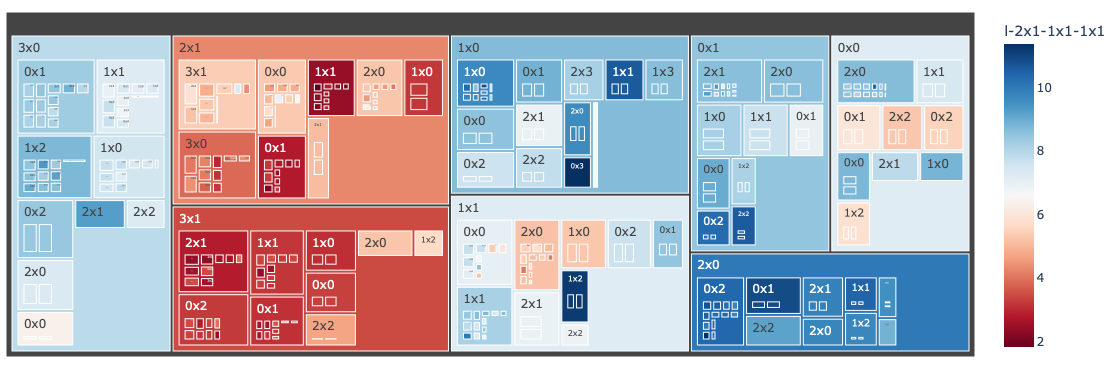}
        \label{fig:w-l2x1-1x1-1x1}
    }
    \subfigure[leaf 3x0-0x1-1x1] {
        \includegraphics[width=0.3\textwidth]{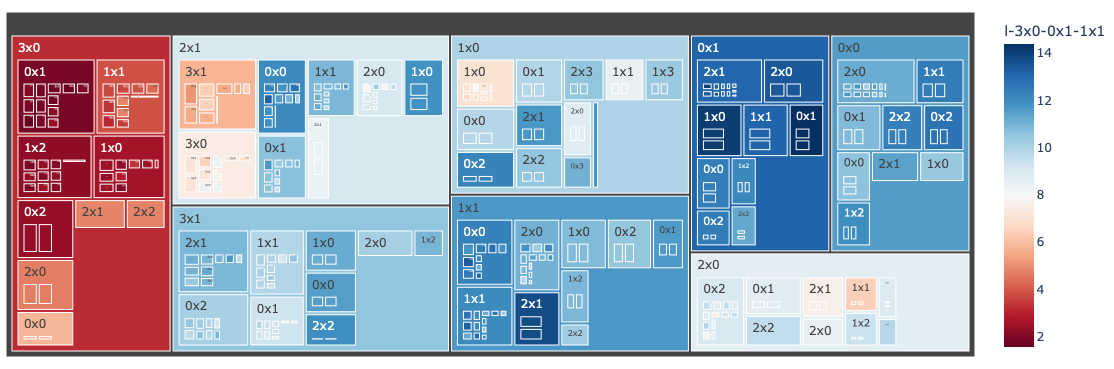}
        \label{fig:w-l3x0-0x1-1x1}
    }
    \caption{The cluster feature maps with color in value difference for significant attributes of target clusters on CyTOF-Wong dataset.}
    \label{fig:sf-wong-dis}
\end{figure}

We calculated the data value difference of identified significant attributes to check the selection results. The result is showed with cluster feature maps (Fig.~\ref{fig:sf-wong-dis}).
The continuous color in plots represents the value difference of identified significant attributes between the target cluster and other clusters. For example, the identified significant attributes of cluster 3x0-0x1-1x1 are [CD4, CD62L, CCR7, CD27, CD45RO, CD8, ICOS, CD38, CD127, CD45RA], then we calculate the value difference of these attributes between the target cluster data and other clusters data. The cluster feature map shows the difference with continuous color. Blue represents the higher difference, and red represents the lower difference. Therefore, we expected the target clusters should be the reddest rectangles in cluster feature maps. 
\par
In Fig.~\ref{fig:w-l2x0-0x1} and Fig.~\ref{fig:w-l3x0-0x1-1x1}, clusters that are near the target clusters or in the same L1 cluster have similar value and are presented red. The significant attributes of these two clusters have unique performance in the target clusters and the upper-level cluster.
\subsubsection{Cluster distribution map: spatial relations of clusters}
\begin{figure}[htb]
    \centering
    \includegraphics[width=3.1in]{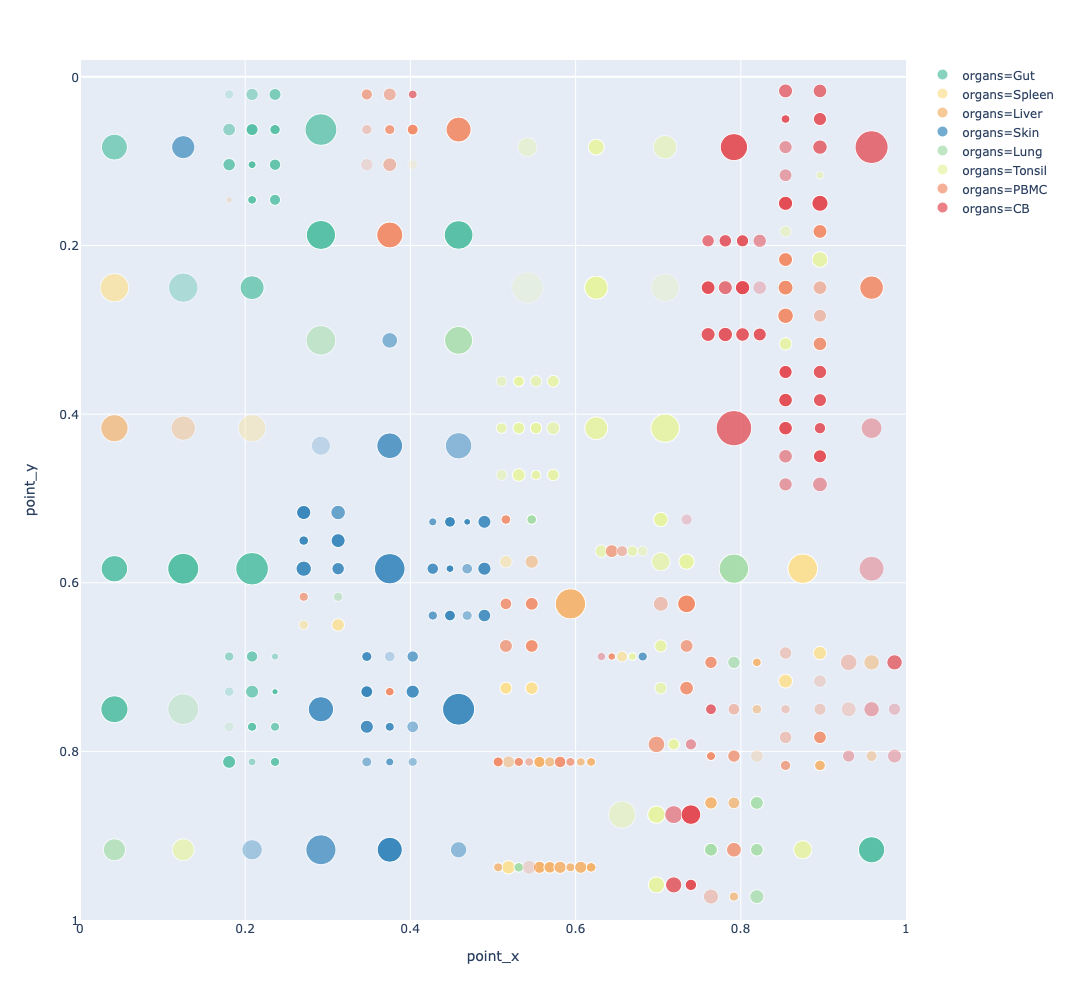}
    \caption{The cluster distribution map with color in cell types on CyTOF-Wong dataset.}
    \label{fig:wong_bubble}
\end{figure}

The colors of the cluster distribution map represent the sample type of cells (Fig.~\ref{fig:wong_bubble}). The color saturation shows that how pure the cell type is in the cluster. The more opaque, the more cells of the label are in the cluster. Through the cluster distribution map, we confirm that GHSOM clustering result can distinguish cell type. The clusters that have the same cell type label are near to each other. The related study revealed that the relation of different sample types such as sample development trend~\cite{2019DimensionalityReduction}. We can also find the same trend through cluster distribution map as that CB (Red) and PBMC (Orange) develop to Liver and Spleen, then develop to Tonsil, Skin, Gut, and Lung.

\newpage
\subsection{CyTOF-Samusik dataset}\label{section:samusik}
Samusik dataset contains 86,864 single cells and has 39 marker proteins as analyzed parameters. Cell population labels are available for 24 manually gated populations. We cluster single-cell data with 39 proteins as attributes.
\subsubsection{GHSOM Clustering}
\begin{figure}[htb]
    \centering
    \includegraphics[width=5in]{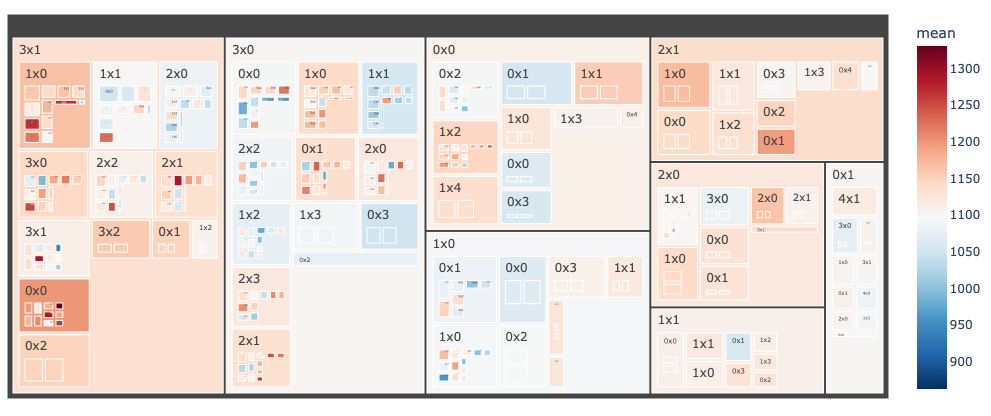}
    \caption{The cluster feature map with color in the mean value of attributes on CyTOF-Samusik dataset.}
    \label{fig:samusik_treemap}
\end{figure}

The parameter setting of this case is $\tau_1$=0.1 and $\tau_2$=0.01.
The GHSOM clustering result is visualized with a cluster feature map as Fig.~\ref{fig:samusik_treemap}. The clustering result is four levels in depth and has eight clusters on the L1 result. The feature color represents the average cell mean across attributes. We can observe which cluster has a high or low value through the cluster feature map and quickly get the highest or lowest clusters. Cluster 3x1-0x0-1x4 has the highest value, and cluster 3x1-3x1-0x0 has the lowest value.

\subsubsection{Significant Attributes Identification}
We picked five leaf clusters. Cluster 3x1-3x1-0x0 is the cluster with the lowest mean value; cluster 3x1-0x0-1x4 is the cluster with the highest mean value. And randomly picked three clusters with common mean values: cluster 2x1-0x1, cluster 0x0-0x1, and cluster 1x0-1x0-1x0. 
We identified the significant attributes of these five leaf clusters with the Significant Attributes Identification method. The significant attributes of clusters are shown in Table~\ref{table:sf-samusikleaf}. 
We found some attributes that were identified as significant in clusters. MHCII and Sca1 are identified in four of the five leaf clusters we picked. Ly6C, CD11b, CD43, and B220 are all identified.

\begin{table}[htb]
\centering
\resizebox{\textwidth}{!}{
\begin{tabular}{|c|p{330pt}|} 
\hline 
Leaf cluster & Significant attributes \\
\hline 
3x1-3x1-0x0 & Ly6C, CD11b, B220, CD44, CD34, IgM, CD16\_32, MHCII, Sca1, CD43\\
\hline
3x1-0x0-1x4 & Ly6C, CD11b, CD64, cKit, Sca1, B220, CD43, CD44, MHCII, IgM\\
\hline
2x1-0x1 & SiglecF, Ly6C, F480, CD34, CD16\_32, CD43, B220, Sca1, CD11b, IgM\\
\hline
0x0-0x1 & Ly6C, B220, MHCII, CD11b, IgM, IgD, CD43, Sca1, CD44, cKit\\
\hline
1x0-1x0-1x0 & Ly6C, CD11b, IgM, B220, CD16\_32, CD43, MHCII, cKit, CD64, SiglecF\\
\hline
\end{tabular}
}
\caption{Significant attributes of leaf clusters in CyTOF-Samusik dataset.}
\label{table:sf-samusikleaf}
\end{table}

\begin{figure}[htb]
    \centering
    \subfigure[leaf 0x0-2x0] {
        \includegraphics[width=0.3\textwidth]{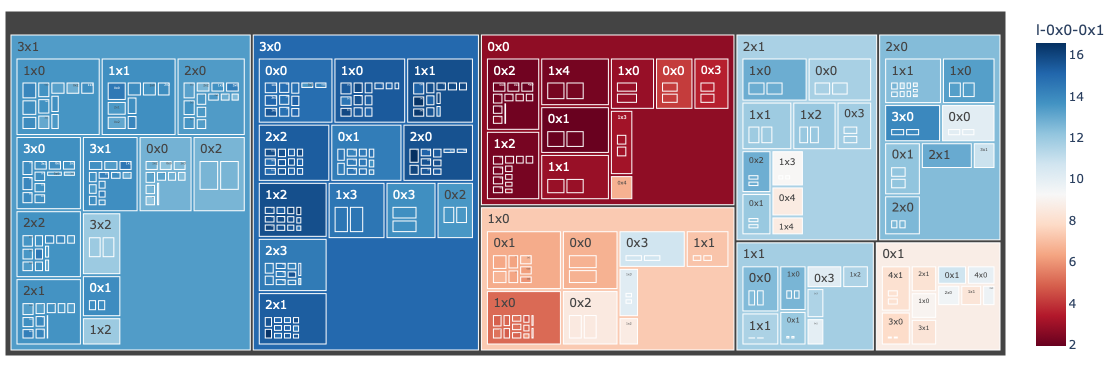}
        \label{fig:s-l0x0-2x0}
    }
    \subfigure[leaf 1x0-1x0-1x0] {
        \includegraphics[width=0.3\textwidth]{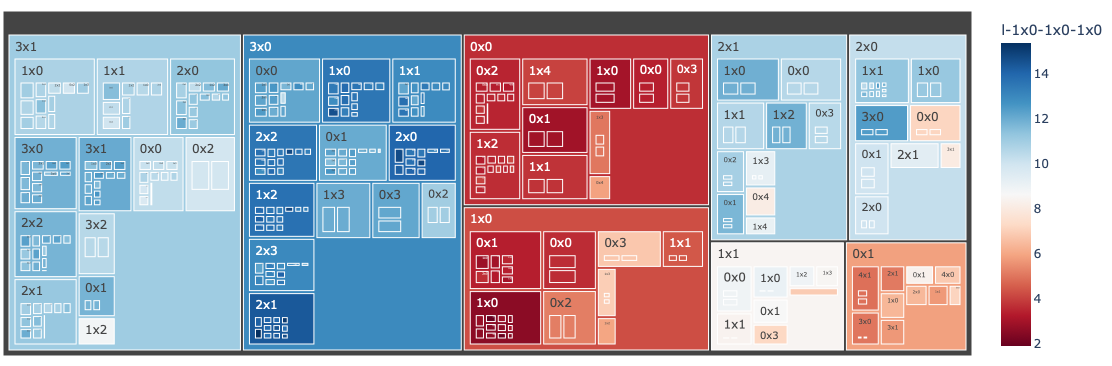}
        \label{fig:s-l1x0-1x0-1x0}
    }
    \subfigure[leaf 2x1-0x1] {
        \includegraphics[width=0.3\textwidth]{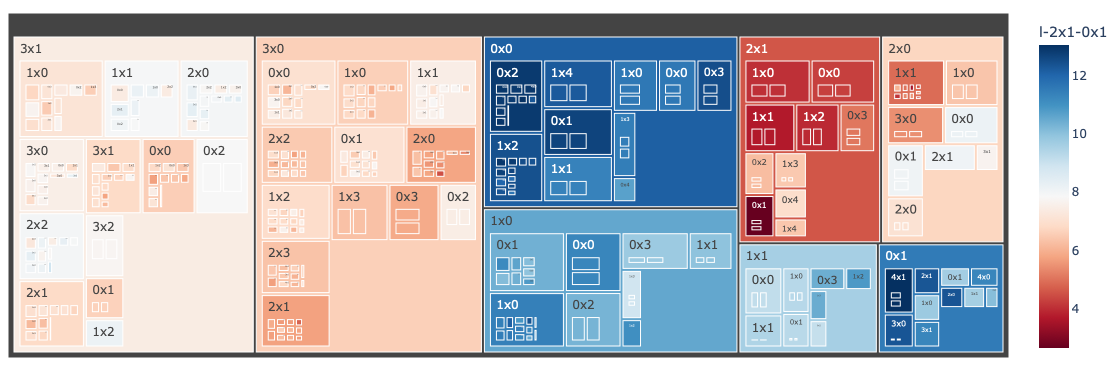}
        \label{fig:s-l2x1-0x1}
    }
    \subfigure[leaf 3x1-0x0-1x4] {
        \includegraphics[width=0.3\textwidth]{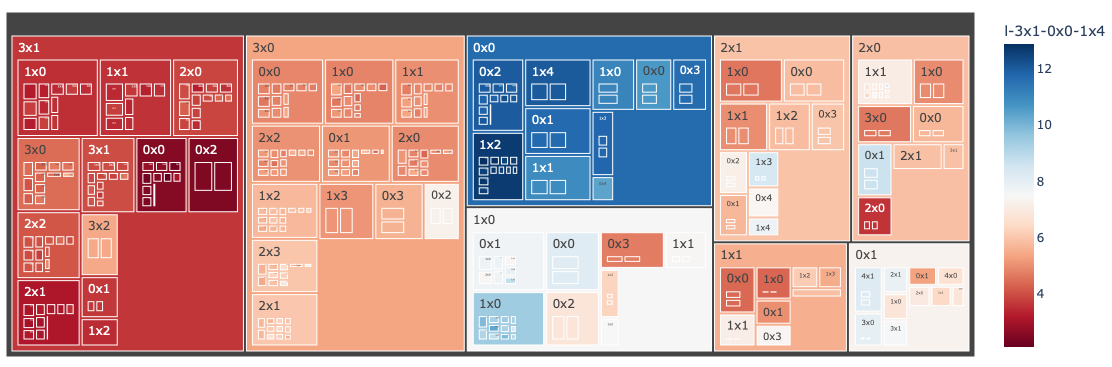}
        \label{fig:s-l3x1-0x0-1x4}
    }
    \subfigure[leaf 3x1-3x1-0x0] {
        \includegraphics[width=0.3\textwidth]{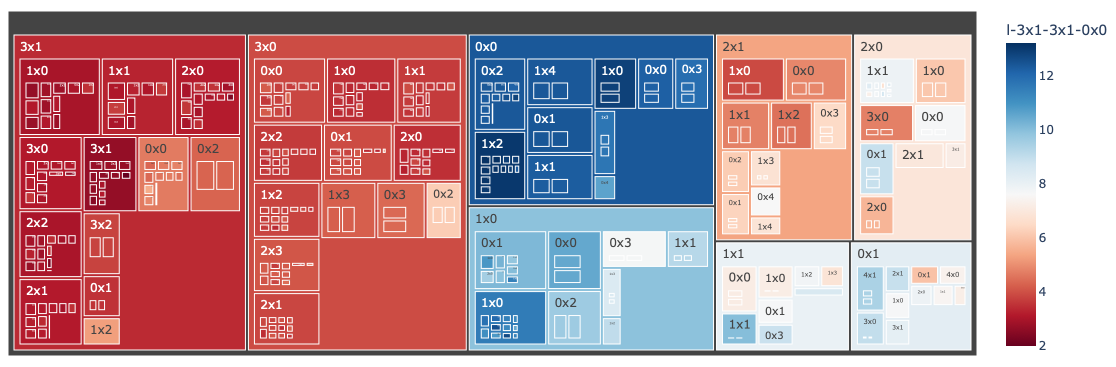}
        \label{fig:s-l3x1-3x1-0x0}
    }
    \caption{The cluster feature maps with color in value difference for significant attributes of target clusters on CyTOF-Samusik dataset.}
    \label{fig:sf-samusik-dis}
\end{figure}
\subsubsection{Cluster feature map}
After identifying the significant attributes, we showed how the result is with the cluster feature map. We calculated the difference between significant attributes in the target cluster and other clusters to check the identified results. The cluster feature maps is shown in Fig.~\ref{fig:sf-samusik-dis}. 
The continuous color in plots represents the same feature as the last case, which is the value difference of identified significant attributes between the target cluster and other clusters. Blue represents the higher difference, and red represents the lower difference. Therefore, we expected the target clusters should be the reddest rectangles in cluster feature maps. 
\par
In this case, the five leaf clusters we picked do not have special performance on their significant attributes. But we noticed that Fig.~\ref{fig:s-l0x0-2x0} and Fig.~\ref{fig:s-l1x0-1x0-1x0} usually have opposite color of Fig.~\ref{fig:s-l2x1-0x1}, Fig.~\ref{fig:s-l3x1-0x0-1x4} and Fig.~\ref{fig:s-l3x1-3x1-0x0}. We can infer that clusters have clearly different expressions of significant attributes.

\begin{figure}[htb]
    \centering
    \includegraphics[height=4in]{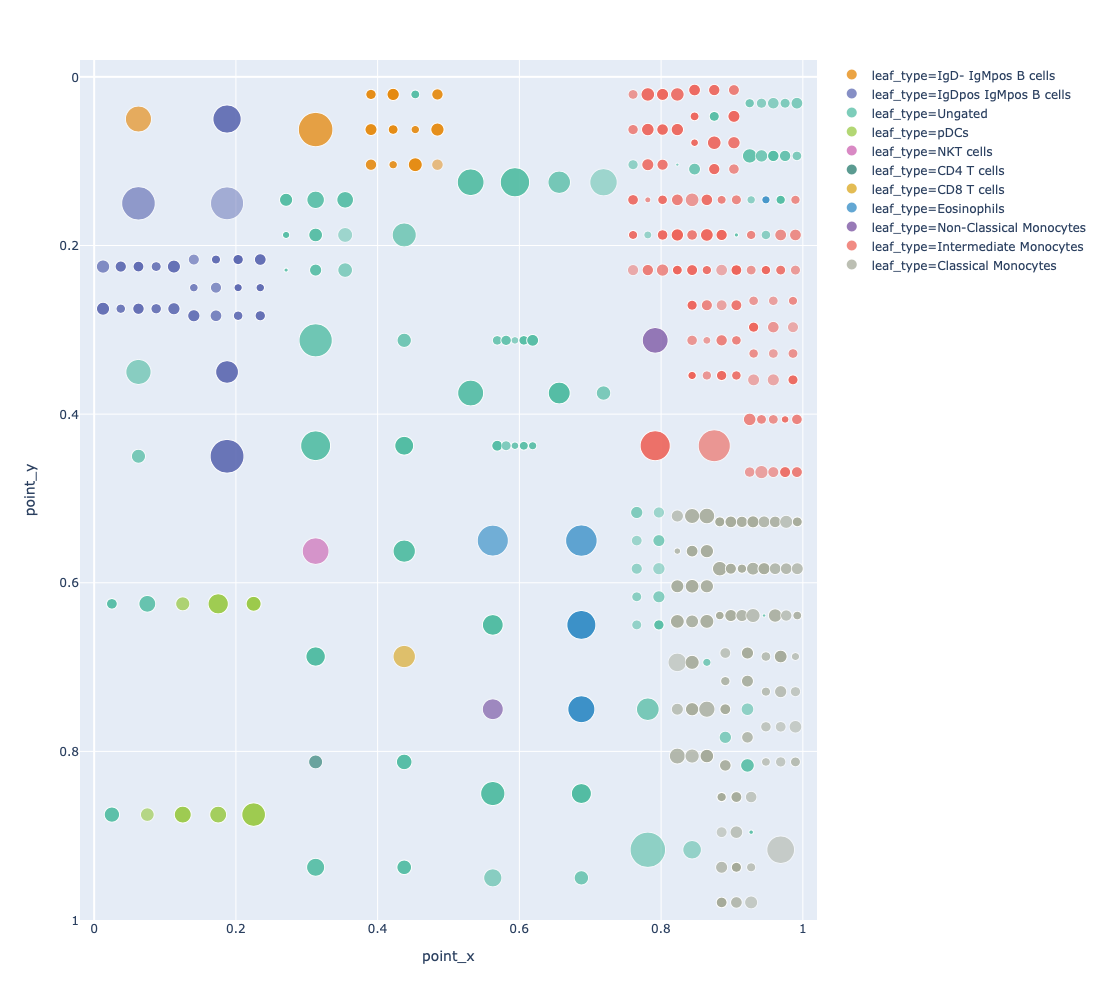}
    \caption{The cluster distribution map with color in cell types CyTOF-Samusik dataset.}
    \label{fig:samusik_bubble}
\end{figure}
\subsubsection{Cluster distribution map: spatial relations of clusters}
We show the spatial relations between clusters with the cluster distribution map. The map's colors (Fig.~\ref{fig:samusik_bubble}) represent the sample type of cells. The color saturation shows how pure the cell type is in the cluster. The more opaque, the more label cells are in the cluster. There are many ungated cells (green) in this dataset. Thus we are hard to analyze the relationship between clusters. Nevertheless, we can still tell that clusters of other cell types gather together.

\subsubsection{Evaluation Comparison}
To evaluate GHSOM clustering performance, we compared evaluation scores to other clustering methods, such as ACCENSE, PhenoGraph, X-shift, kmeans, flowMeans, FlowSOM and DEPECHE. We refer to \textit{A comparison framework and guideline of clustering methods for mass cytometry data}\cite{comparisonFramework_2019}, which did a comparison of different clustering methods on some mass cytometry datasets, we did internal and external evaluations to GHSOM clustering result. The internal and external evaluations we did in this section are introduced in Sec.~\ref{section:threshold}. The internal evaluation is CH and the external one is ARI.
\par
Table~\ref{table:samusik-comparison} shows the CH and ARI scores of clustering methods in this case. As we mentioned in Sec.~\ref{section:threshold}, we chose the parameter $\tau_1$ and $\tau_2$ by trying different pairs of them. Here we chose $\tau_1$ is 0.08 and $\tau_2$ is 0.15. We can tell that the performance of GHSOM is above average. In the external evaluation, GHOSM is the third place of all. In the internal evaluation, GHSOM is the best. GHSOM performs great in the external evaluation, showing that GHSOM's result reveals data types at a certain level. GHSOM performs outstandingly in the internal evaluation, showing that GHSOM clusters data clearly with attribute values.
\begin{table}[htb]
\centering
\begin{tabular}{|c|c|c|} 
\hline 
Clustering method & CH & ARI \\
\hline 
ACCENSE & 3.2952 & 0.5574\\
\hline
PhenoGraph & 3.7380 & 0.9250\\
\hline
X-shift & 3.2660 & 0.8781\\
\hline
K-means & 3.7087 & 0.4655\\
\hline
flowMeans & 3.4029 & 0.9206\\
\hline
FlowSOM & 3.6941 & 0.8561\\
\hline
DEPECHE & 4.1028 & 0.8298\\
\hline
GHSOM & 4.2 & 0.88\\
\hline
\end{tabular}
\caption{Internal and External evaluation of different clustering methods. The scores of other clustering methods are referred from~\cite{comparisonFramework_2019}.}
\label{table:samusik-comparison}
\end{table}
\subsection{scRNA-Seq data-PBMCs dataset}
We experimented on 3k Peripheral Blood Mononuclear Cells (PBMCs) dataset. PBMCs consist of lymphocytes (T cells, B cells, NK cells) and monocytes. PBMCs could refer to any peripheral blood cell having a round nucleus. PBMCs may be susceptible to pathogenic infections and viral infections. PBMCs are frequently used in immunology, infectious disease, hematological malignancies, vaccine development, transplant immunology, and high-throughput screening. PBMCs have also been thought to be an important route of vaccination.
3k PBMCs dataset is a single-cell gene expression with 2,700 single cells detected and 13714 genes. 
\par 
We perform preprocessing including data normalization, Highly Variable Genes (HVGs) identification, and data scaling. Data normalization is implemented with \textit{LogNormalize}. Then we identify 2,000 HVGs to filter more relative genes and remove the genes with less information using the Feature Selection method~\cite{STUART20191888}, which is used in Seurat. The feature selection method learns the mean-variance relationship from the data, computes standardized feature variance, and then selects the highest standardized variance as "highly variable." For data scaling, we scale the expression data to make the mean across cells be 0 and the variance across cells be 1. We implemented above process through R package, Seurat~\cite{seurat_2015}.

\subsubsection{GHSOM clustering}
\begin{figure}[htb]
    \centering
    \includegraphics[width=5in]{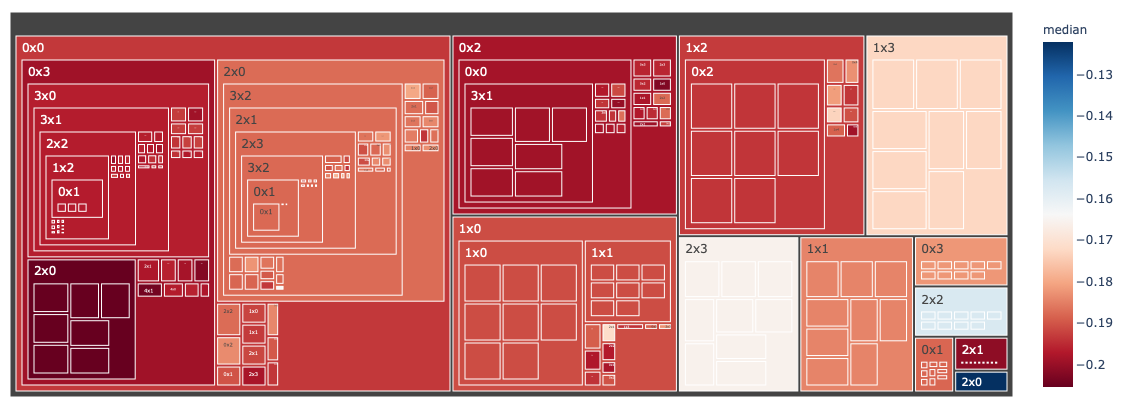}
    \caption{The cluster feature map with color in the median value of attributes on scRNA-seq-PBMCs dataset.}
    \label{fig:fg_map}
\end{figure}

The parameter setting of this case is $\tau_1$=0.1 and $\tau_2$=0.1.
The result of the clustering PBMCs dataset is ten levels deep and has twelve clusters on the L1 result (Fig.~\ref{fig:fg_map}). The color represents the average value of cells in the same cluster. Here the value is defined as the cell median across genes (The color bar in Fig.~\ref{fig:fg_map}). We can observe how clusters perform through the cluster feature map and quickly know the highest or lowest clusters. Cluster 0x0-0x3-2x0 has the highest value, and cluster 2x0 has the lowest value.

\begin{figure}[htb]
    \centering
    \subfigure[seurat]{
        \includegraphics[width=1.7in]{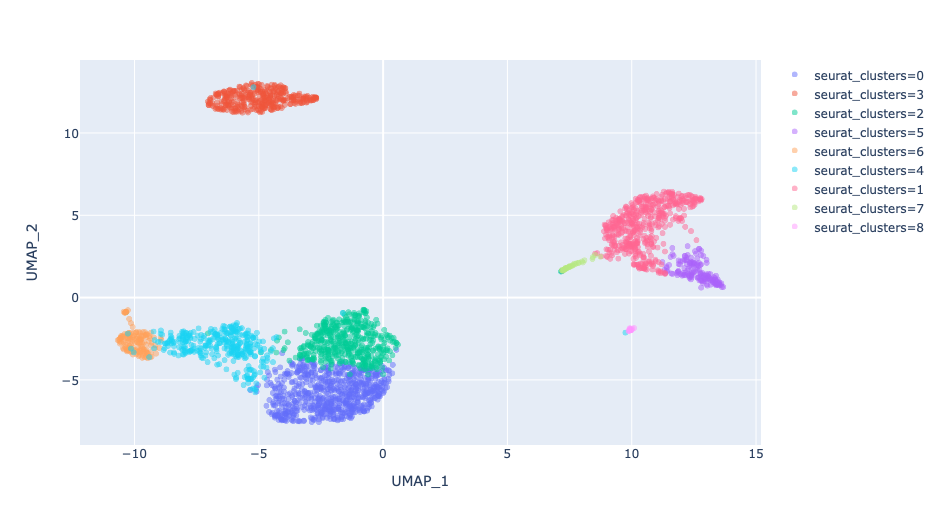}
    }    
    \subfigure[GHSOM L1-clusters]{
        \includegraphics[width=1.7in]{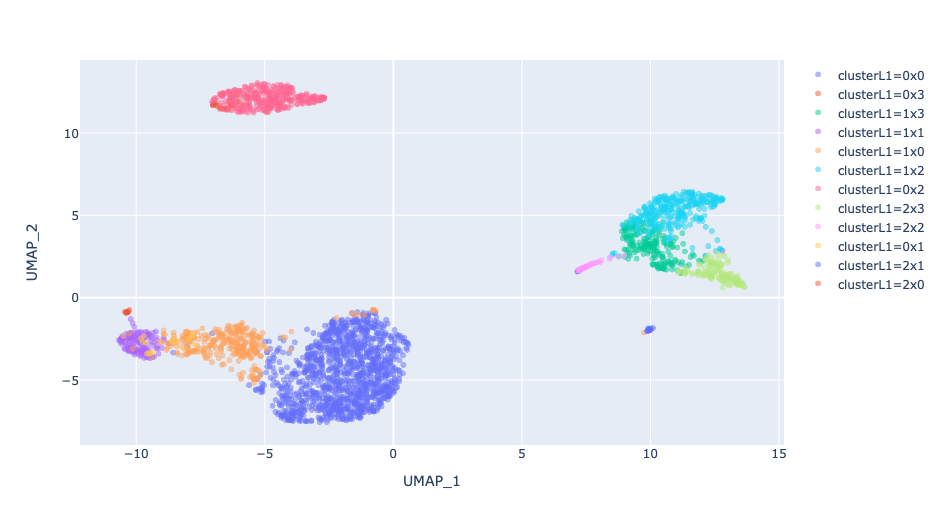}
    }
    \subfigure[GHSOM L2-clusters in L1 0x0]{
        \includegraphics[width=1.7in]{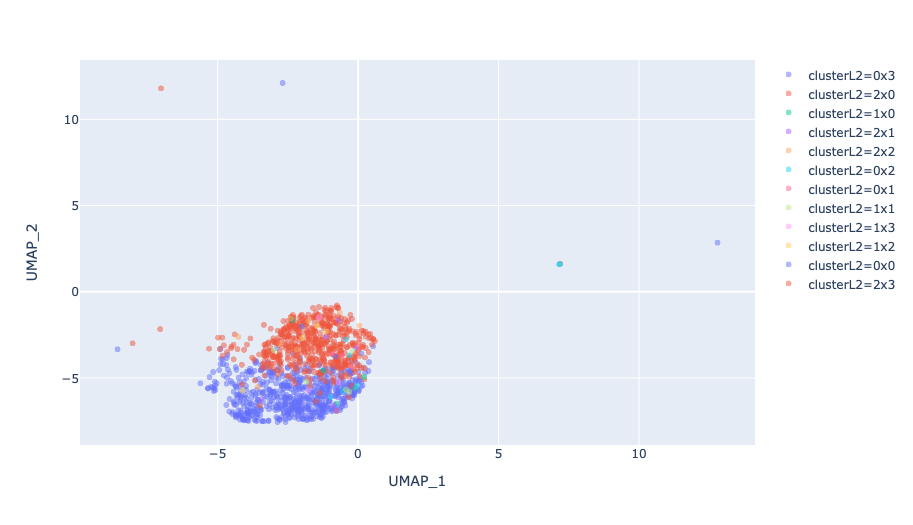}
    }
    \caption{The UMAP of Seurat and GHSOM clusters on scRNA-seq-PBMCs dataset.}
    \label{fig:umap}
\end{figure}

Fig.~\ref{fig:umap} contains the clustering results of Seurat and GHSOM's Level1 (L1) on UMAP. Of course, there is some difference between the two clustering results. The obvious one is that the data in the cluster (L1 0x0, purple Fig.~\ref{fig:umap}b) of the GHSOM result was clustered into two clusters, cluster 2(green) and cluster 0(purple) in Seurat's result (Fig.~\ref{fig:umap}a). Moreover, if we explore deeper into the cluster, L1 0x0, we would see that the clusters of L2 (Fig.~\ref{fig:umap}c) are generally similar to the result of Seurat. Thus, we could observe that GHSOM has a similar big-picture result as Seurat's.

\subsubsection{Significant Attributes Identification}
We picked five leaf clusters that might be interesting to be taken a look. Cluster 2x0 is the cluster with the lowest mean value; cluster 2x2 is the cluster with the second-lowest mean value; cluster 0x0-0x3-2x0 is the cluster with the highest mean value. Then, we randomly picked two clusters with common mean values: cluster 1x2-1x3 and cluster 0x2-0x0-3x1.
\par
We identified the significant attributes of these five leaf clusters with the Significant Attributes Identification method. The significant attributes of clusters are shown in Table.~\ref{table:sf-pbmcleaf}. 
In this case, we identified various significant attributes. The significant attributes that we identified are most different in clusters. Nevertheless, we found some common significant attributes in different clusters, such as CD74 and HLA-DRB1.
\linespread{1.2}
\begin{table}[htb]
\centering
\resizebox{\textwidth}{!}{
\begin{tabular}{|c|p{330pt}|} 
\hline 
Leaf cluster & Significant attributes\\
\hline 
2x0 & KIAA0101, TYMS, ZWINT, RRM2, BIRC5, TK1, PCNA, GINS2,STMN1, MALAT1\\
\hline 
2x2 & FCER1A, CST3, HLA-DPA1, HLA-DRA, HLA-DPB1, HLA-DRB1, CD74, HLA-DRB5, HLA-DQA1, HLA-DMA\\
\hline
0x2-0x0-3x1 & HLA-DRA, CD74, CD79A, HLA-DPB1, HLA-DRB1, CD79B, HLA-DPA1, KRT1, HLA-DQA1, RP11-70P17.1\\
\hline
1x2-1x3 & S100A9, S100A8, JUND, FCGRT, TYROBP, GPX1, LYZ, CST3, FTL, H2AFY\\
\hline
0x0-0x3-2x0 & CCR10, FCRLA, CPNE5, RCL1, RP11-706O15.1, qAC113189.5, ADAL, TTN-AS1, MYOM2, ZFAND4\\
\hline
\end{tabular}
}
\caption{Significant attributes of leaf clusters in scRNA-seq-PBMCs dataset.}
\label{table:sf-pbmcleaf}
\end{table}
\begin{figure}[htb]
    \centering
    \subfigure[leaf 2x0] {
        \includegraphics[width=0.3\textwidth]{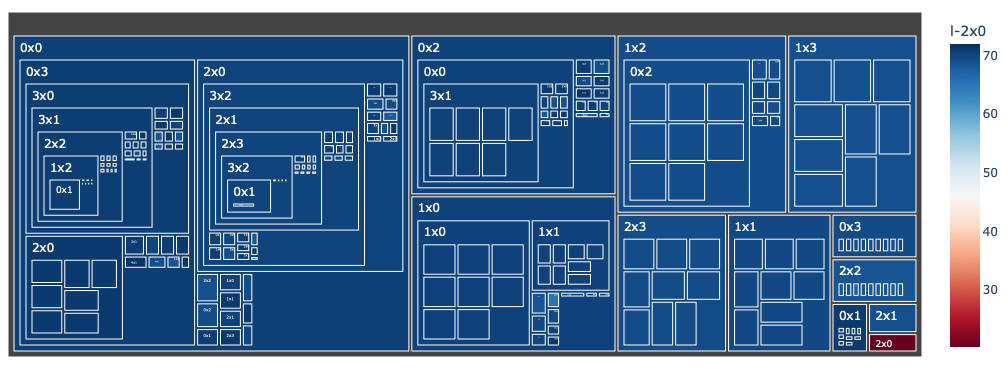}
        \label{fig:p-l2x0}
    }
    \subfigure[leaf 2x2] {
        \includegraphics[width=0.3\textwidth]{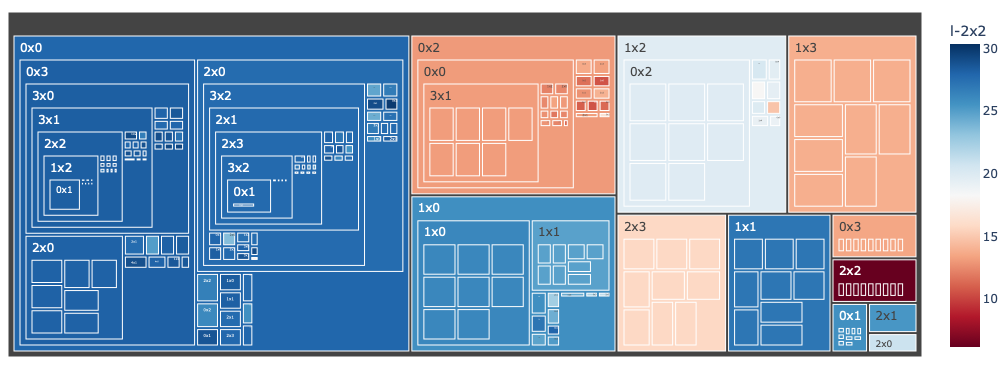}
        \label{fig:p-l2x2}
    }
    \subfigure[leaf 0x2-0x0-3x1] {
        \includegraphics[width=0.3\textwidth]{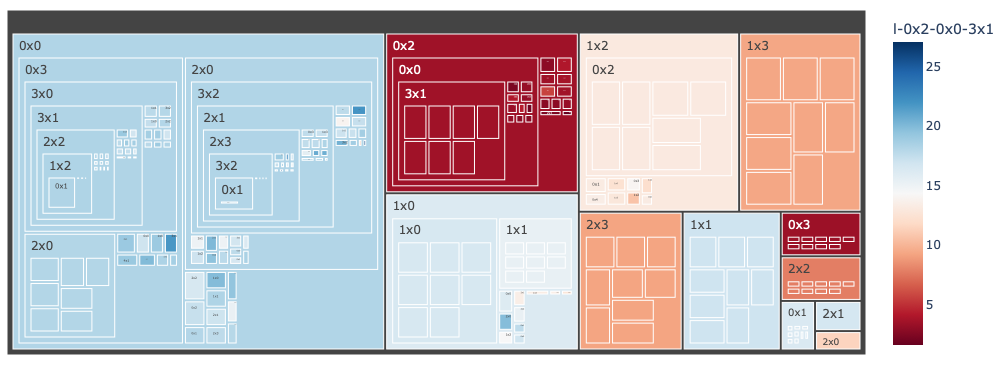}
        \label{fig:p-l0x2-0x0-3x1}
    }
    \subfigure[leaf 1x2-1x3] {
        \includegraphics[width=0.3\textwidth]{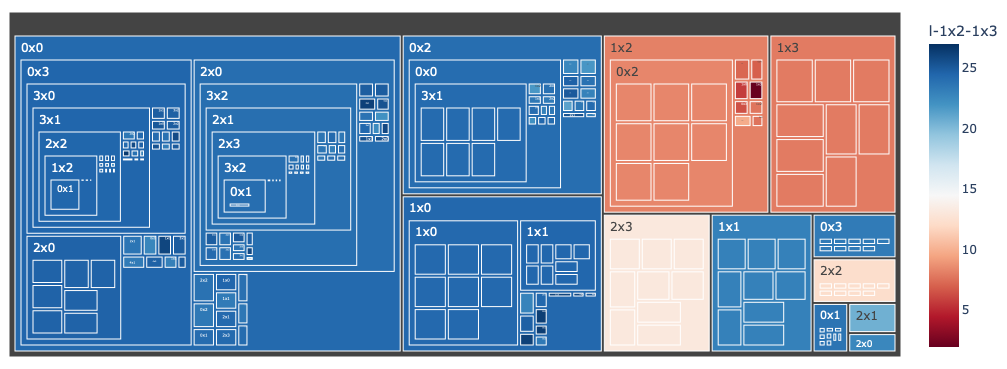}
        \label{fig:p-l1x2-1x3}
    }
    \subfigure[leaf 0x0-0x3-2x0] {
        \includegraphics[width=0.3\textwidth]{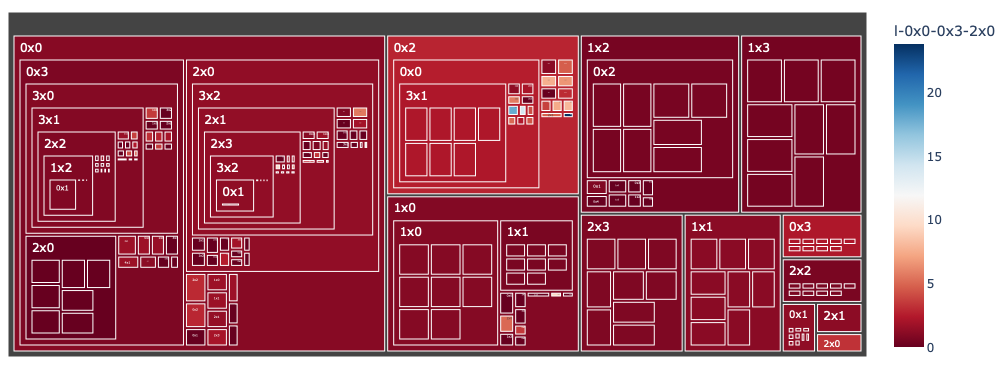}
        \label{fig:p-l0x0-0x3-2x0}
    }
    \caption{The cluster feature map with color in value difference for significant attributes of target clusters on scRNA-seq-PBMCs dataset.}
    \label{fig:sf-pbmc-dis}
\end{figure}
\subsubsection{Cluster feature map}
We calculated the difference between significant attributes in the target cluster and other clusters to check the selection results. The result is showed with cluster feature maps (Fig.~\ref{fig:sf-pbmc-dis}). The continuous color in plots still represents the value difference of identified significant attributes between the target cluster and other clusters. 
\par
In Fig.~\ref{fig:p-l2x0}, we saw that cluster 2x0 is the unique red one among other blue clusters. Therefore, we could infer that the significant attributes in cluster 2x0 have different expressions from other clusters. Unlike cluster 2x0, cluster 0x0-0x3-2x0 almost has all red clusters. Therefore, we could infer that the significant attributes in cluster 0x0-0x3-2x0 have a similar expression in most clusters. We also noticed that some features would be identified in more than one cluster. Cluster 2x2 and cluster 0x2-0x0-3x1 have 6 significant attributes in common. Through the cluster feature map of them (Fig.~\ref{fig:p-l2x2} and Fig.~\ref{fig:p-l0x2-0x0-3x1}), we can observe that the color distributions of two clusters are similar.
\par
\begin{figure}[htb]
    \centering
    \subfigure[MS4A1] {
        \includegraphics[width=1.33in]{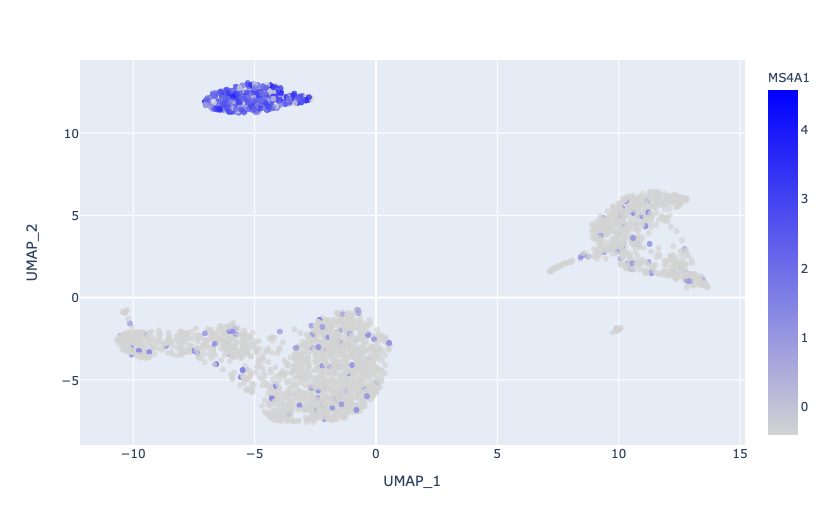}
        \label{fig:e-MS4A1}
    }
    \subfigure[GNLY] {
        \includegraphics[width=1.33in]{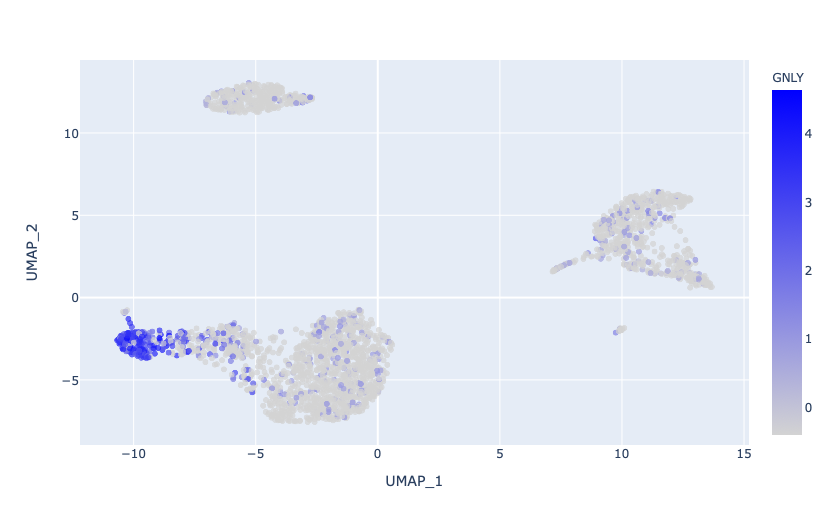}
        \label{fig:e-GNLY}
    }
    \subfigure[CD3E] {
        \includegraphics[width=1.33in]{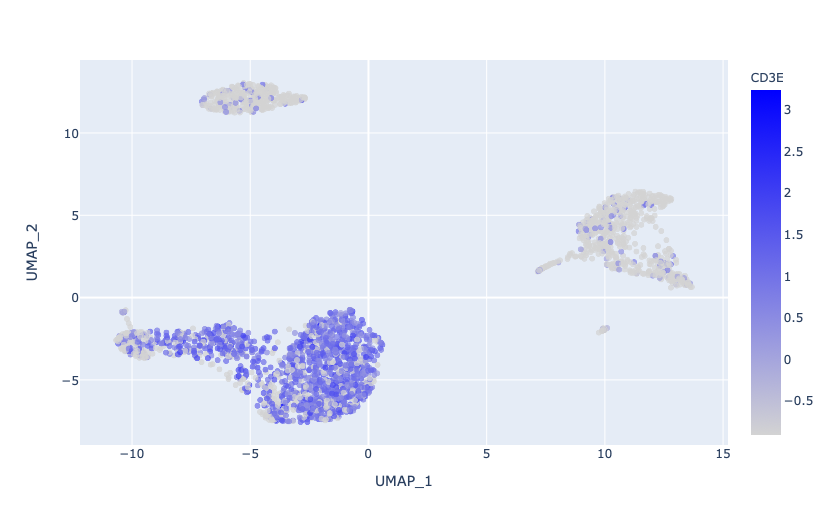}
        \label{fig:e-CD3E}
    }
    \subfigure[CD14] {
        \includegraphics[width=1.33in]{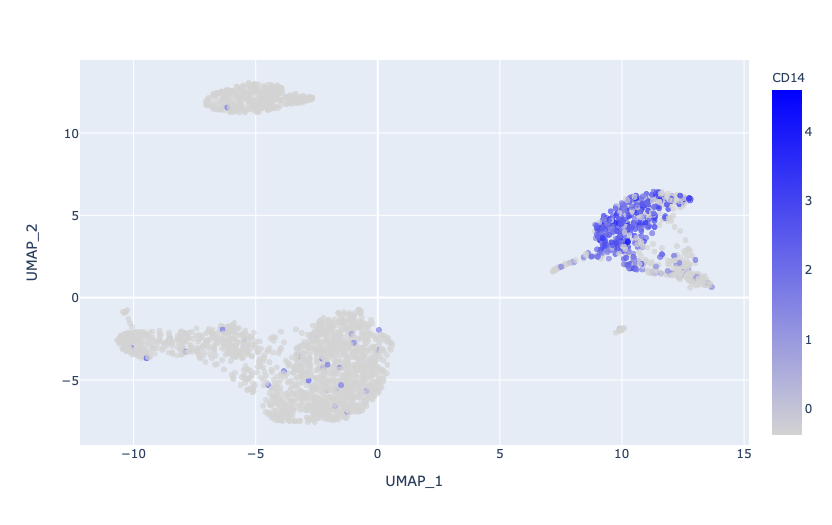}
        \label{fig:e-CD14}
    }
    \subfigure[FCER1A] {
        \includegraphics[width=1.33in]{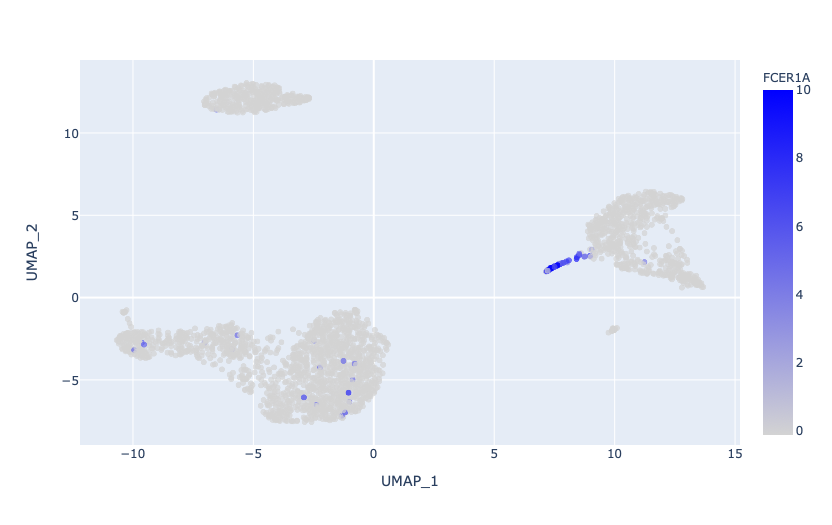}
        \label{fig:e-FCER1A}
    }
    \subfigure[FCGR3A] {
        \includegraphics[width=1.33in]{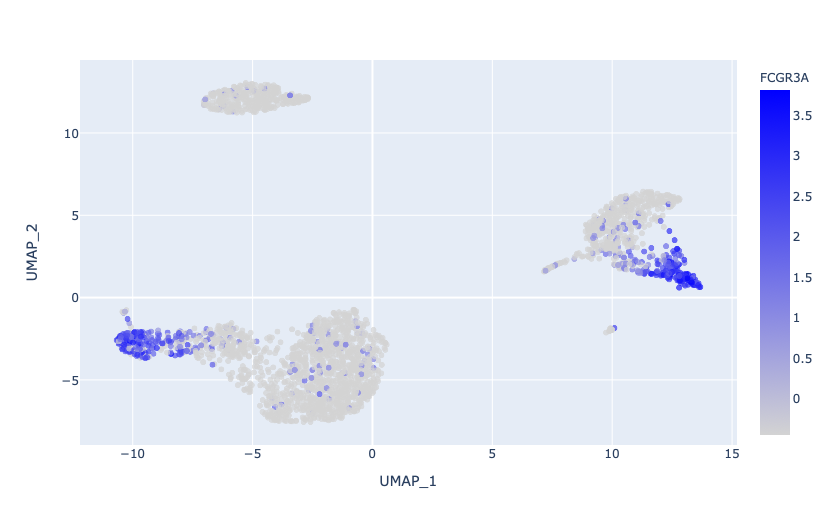}
        \label{fig:e-FCGR3A}
    }
    \subfigure[LYZ] {
        \includegraphics[width=1.33in]{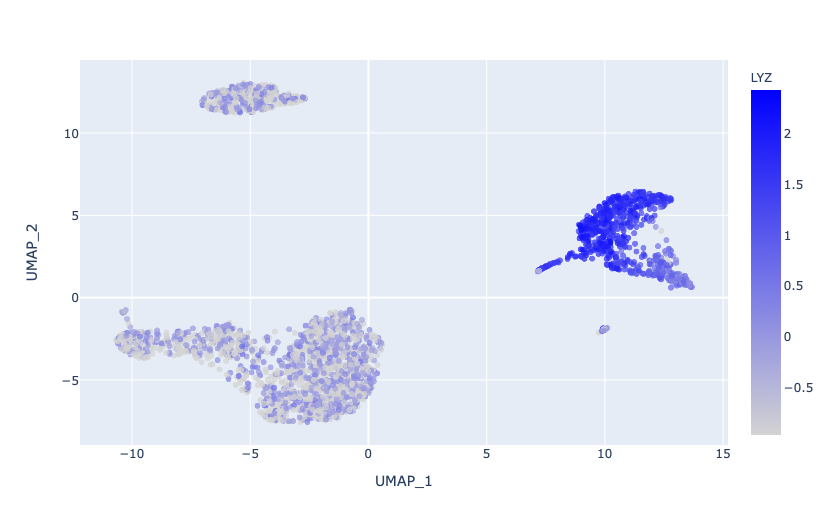}
        \label{fig:e-LYZ}
    }
    \subfigure[PPBP] {
        \includegraphics[width=1.33in]{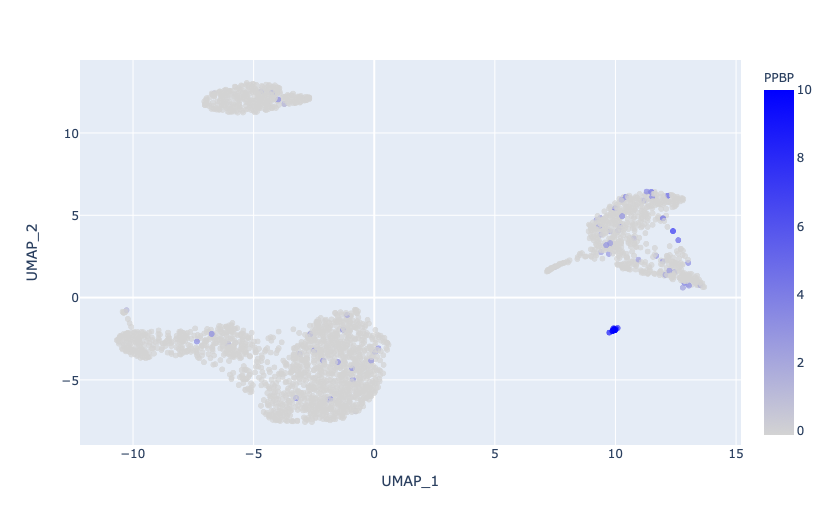}
        \label{fig:e-PPBP}
    }
    \subfigure[CD8A] {
        \includegraphics[width=1.33in]{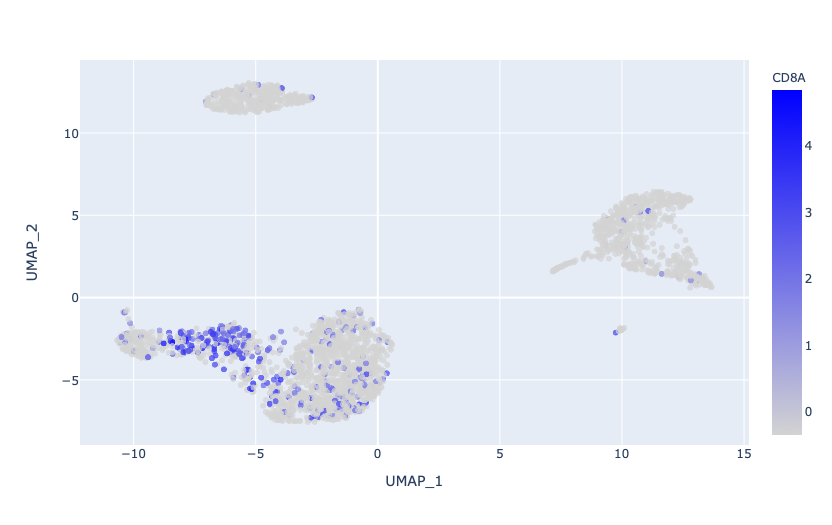}
        \label{fig:e-CD8A}
    }
    \caption{The expression map of the given genes in scRNA-seq-PBMCs dataset.}
    \label{fig:9-expression}
\end{figure}

In Seurat experiment, they showed the gene expression UMAP of nine marker genes that are identified by Seurats FindMarkers() function as example. To compare with them and also make sure that our visualization can reveal information as UMAP, we show the UMAP displays of nine marker genes (Fig.~\ref{fig:9-expression}). Compare the expression of Gene CD14(Fig.~\ref{fig:e-CD14}) and Gene LYZ(Fig.~\ref{fig:e-LYZ}), we observed that CD14 has a high expression on half of the group of cells on the right part, and LYZ has a high expression on the almost whole part of the group of cells on the right part. This situation is hard to be observed with the Seurat clustering result (Fig.~\ref{fig:umap}) that is not hierarchical, but the GHSOM clustering can achieve it.
\par
\begin{figure}[h]
    \centering
    \subfigure[MS4A1] {
        \includegraphics[width=1.33in]{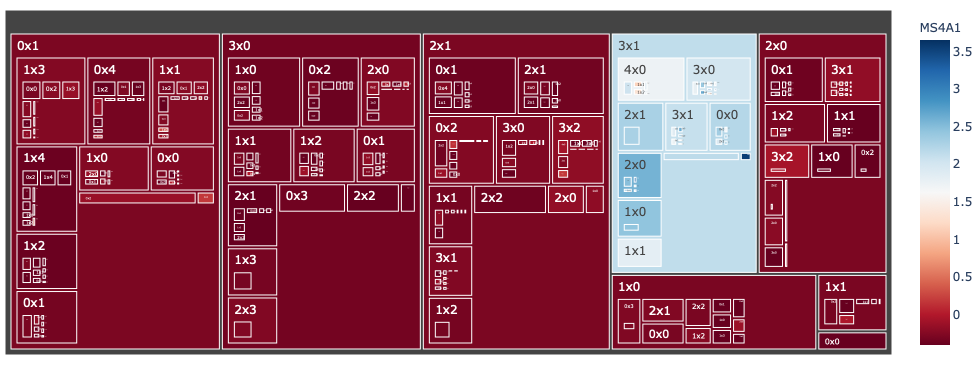}
        \label{fig:MS4A1}
    }
    \subfigure[GNLY] {
        \includegraphics[width=1.33in]{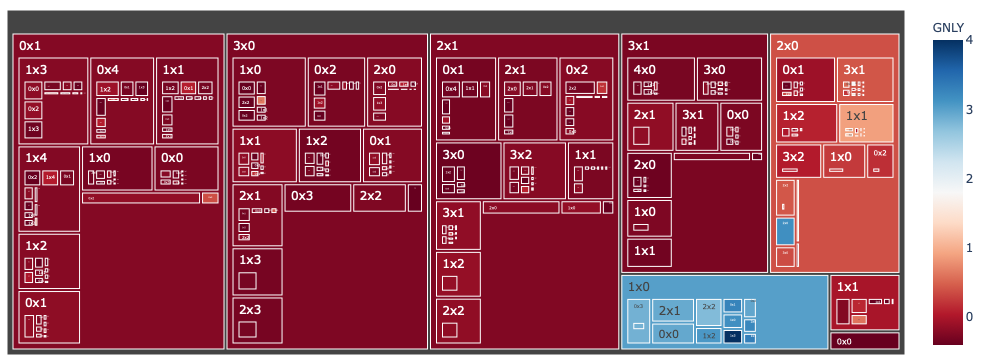}
        \label{fig:GNLY}
    }
    \subfigure[CD3E] {
        \includegraphics[width=1.33in]{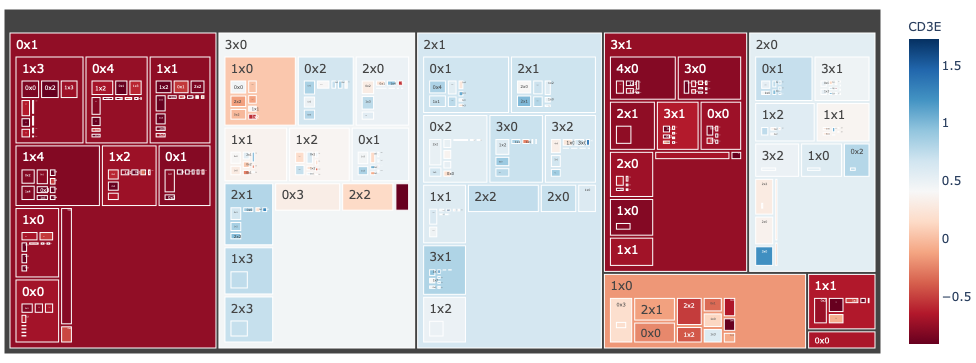}
        \label{fig:CD3E}
    }
    \subfigure[CD14] {
        \includegraphics[width=1.33in]{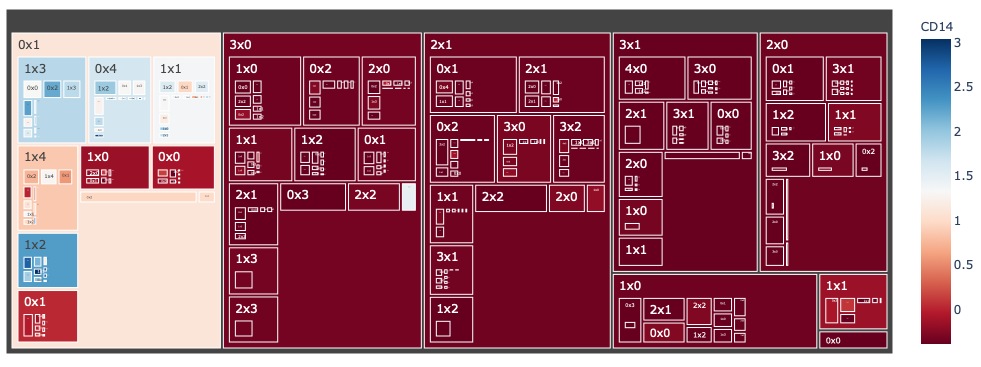}
        \label{fig:CD14}
    }
    \subfigure[FCER1A] {
        \includegraphics[width=1.33in]{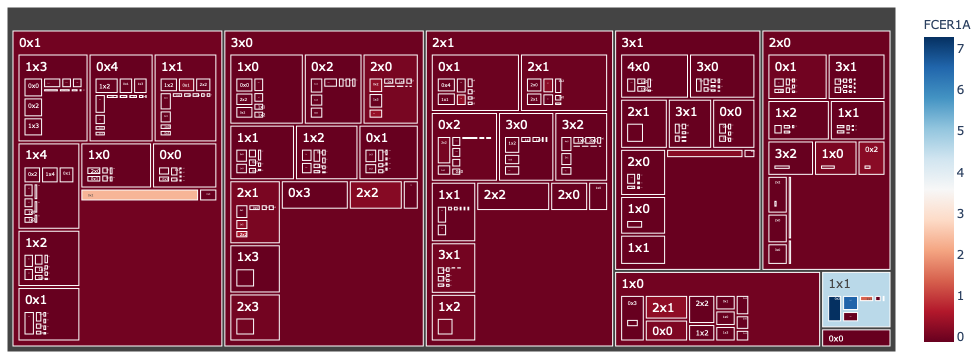}
        \label{fig:FCER1A}
    }
    \subfigure[FCGR3A] {
        \includegraphics[width=1.33in]{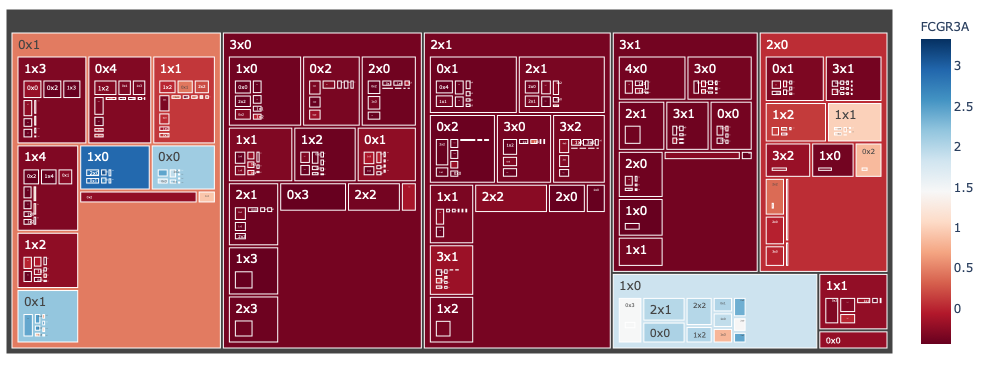}
        \label{fig:FCGR3A}
    }
    \subfigure[LYZ] {
        \includegraphics[width=1.33in]{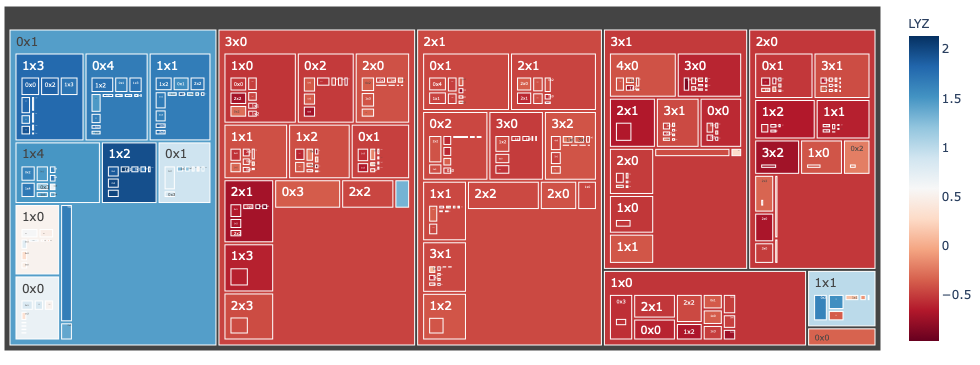}
        \label{fig:LYZ}
    }
    \subfigure[PPBP] {
        \includegraphics[width=1.33in]{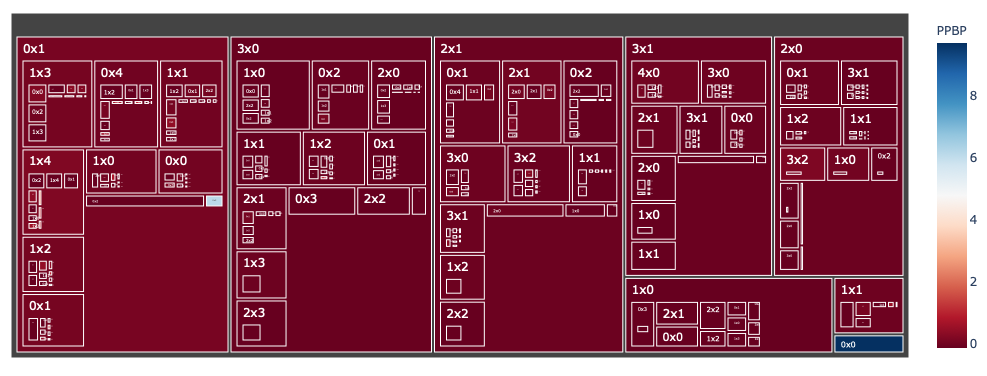}
        \label{fig:PPBP}
    }
    \subfigure[CD8A] {
        \includegraphics[width=1.33in]{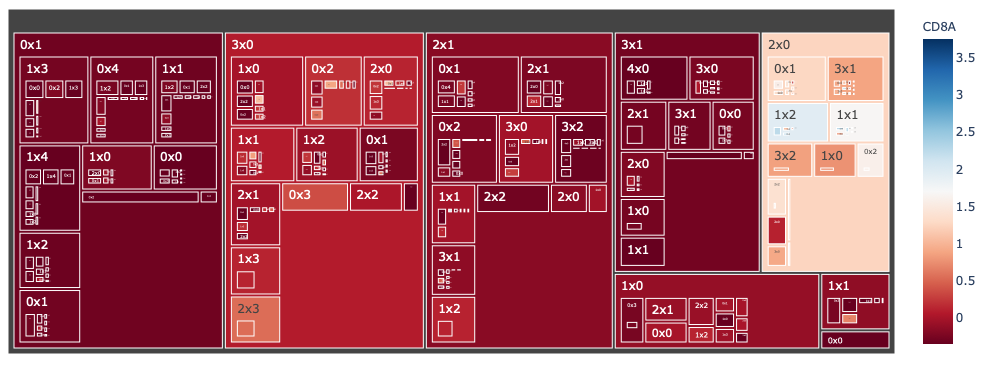}
        \label{fig:CD8A}
    }
    \caption{The cluster feature maps with color in the given genes expression value in scRNA-seq-PBMCs dataset.}
    \label{fig:GHSOM_9}
\end{figure}

The cluster feature map visualization (Figure~\ref{fig:GHSOM_9}) reveals the cluster with high expression of the given genes. In Fig.~\ref{fig:GHSOM_9}, the color represents the expression value of the marker gene. "Blue" represents the top values, and "Red" represents the bottom values. We confirmed that the blue boxes in the cluster feature maps are the same part of cells in blue color in Fig.~\ref{fig:9-expression}. With the cluster feature map, we can even know the high expression cluster of a gene without dimensional reduction methods such as UMAP or tSNE. The cluster feature map can reveal the same thing with UMAP or tSNE results without making a dimensional reduction to data.


\clearpage
\subsubsection{Cluster distribution map: spatial relations of clusters}
\begin{figure}[htb]
    \centering
    \includegraphics[width=3in]{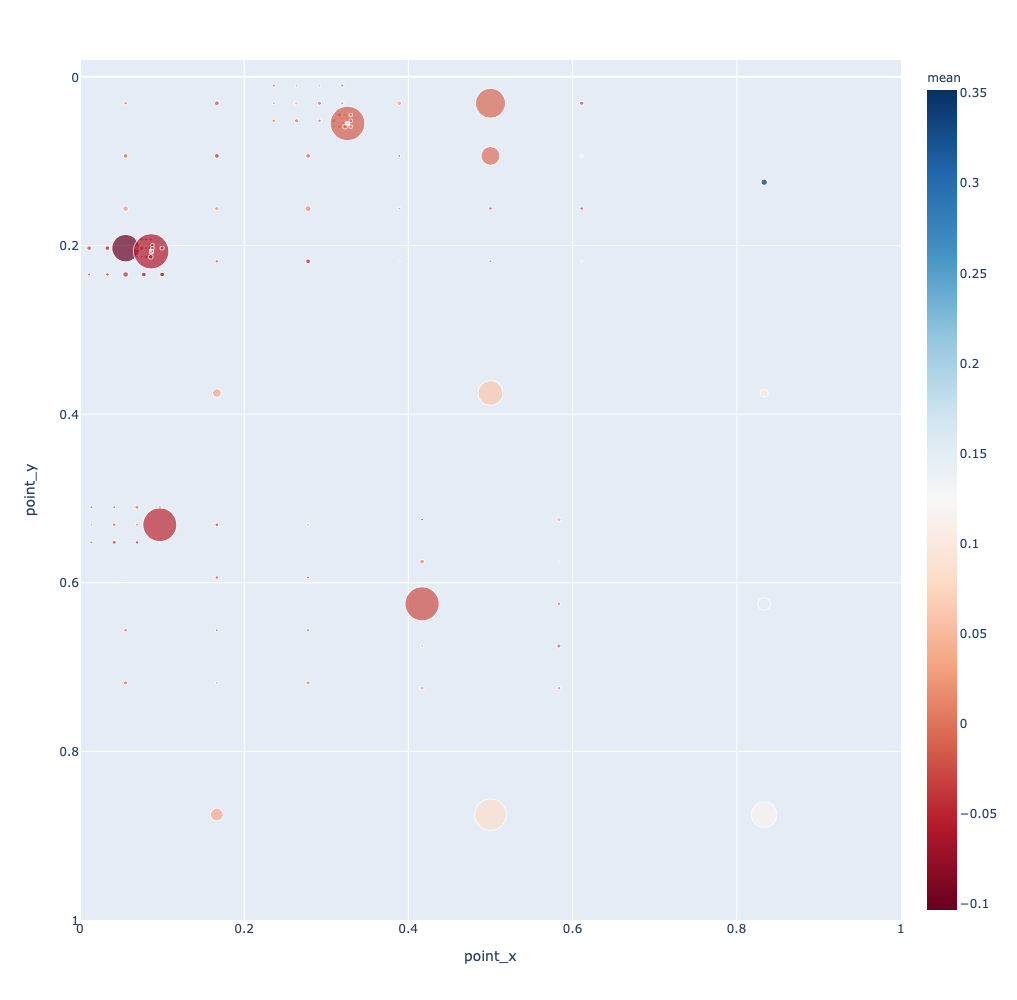}
    \caption{The cluster distribution map with color in the mean value of attributes on scRNA-seq-PBMCs dataset..}
    \label{fig:fg_bubble}
\end{figure}

The cluster distribution map of PBMCs clustering result is shown as Fig.~\ref{fig:fg_bubble}. In this case, we observe the different feature from last cases. Instead of data types, the color represents the average expression value of clusters. We can view the overall result with it and know how the values distribute in leaf clusters.

\subsection{CRISPR data} \label{section:crispr}
The Cancer Dependency Map (DepMap) is an ongoing project to uncover gene dependency in hundreds of cancer cell lines. The DepMap portal aims to empower the research community to make discoveries related to cancer vulnerabilities by providing access to key cancer dependencies analytical and visualization tools. 
They use genome-scale RNAi\cite{depmap_2017} and CRISPR-Cas9\cite{CRISPR_2019} genetic perturbation reagents to silence or knockout individual genes and identify those genes that affect cell survival. By linking these dependencies to the genetic or molecular features of the tumors, Achilles is providing the foundation for the “DepMap”. Achilles systematically identifies and catalogs gene essentiality across hundreds of genomically characterized cancer cell lines.
\par
In this case, we use the same dataset as shinyDepMap\cite{shinyDepMap_2021}. It is the combination of CRISPR and RNAi datasets from Depmap Portal. The Broad Institute normalized the CRISPR and shRNA efficacy data. ShinyDepMap computed the combined perturbation score with the data of 15,847 genes in 423 cell lines, which were examined with both CRISPR and shRNA. The perturbation score is as scaled as the efficacy of the original data. The lower value presents that this cancer cell is more dependent on this gene.
\begin{figure}[htb]
    \centering
    \includegraphics[width=3in]{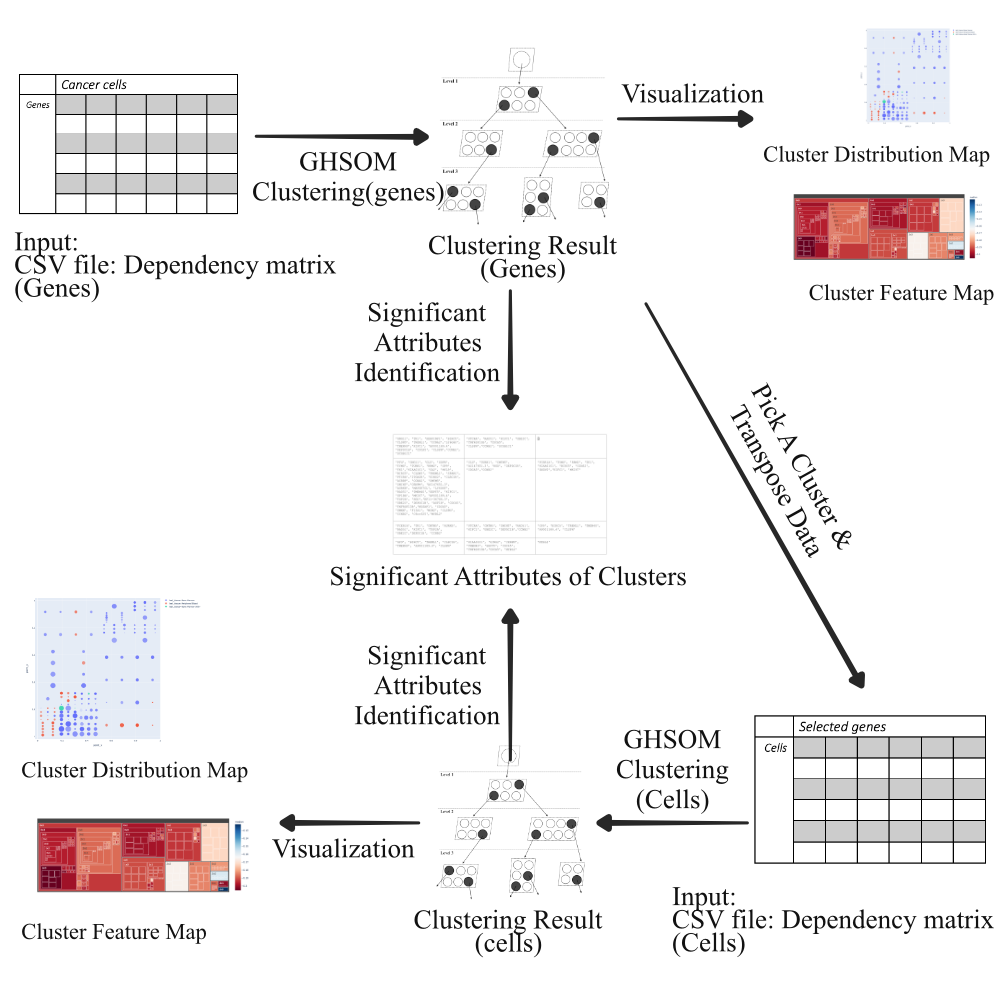}
    \caption{The workflow of CRISPR case.}
    \label{fig:crispr workflow}
\end{figure}

\subsubsection{Workflow} \label{sec:crispr-workflow}
The workflow of this case is similar to the cases before. We do the same workflow on cluster genes. There is an extra step: we transpose the data and do the experiment routine to cluster "cells." Therefore, we do twice the clustering routine in this case, one for clustering genes and another for clustering cells.

\paragraph{Cluster genes}The workflow is as Fig.~\ref{fig:crispr workflow}. 
Cluster gene data with all cancer cells as attributes. We will have the result of genes with a similar dependency value for every cancer cell in one cluster.
\paragraph{Pick a cluster of genes}
Choose a specific cluster. If we have the target gene that we are interested in or would like to research, we can directly choose the genes cluster. In this case, we choose the cluster(L1 2x0, L2 2x0) with the lowest efficacy value representing highly relative.
\paragraph{Cluster cells}
Transpose of the data. Cluster cancer cell data with the genes we picked out at the last step as attributes. We will have the result of cancer cells with similar dependency values of these genes in one cluster.
\paragraph{Efficacy and selectivity System}
To observe our result, we referred to shinyDepMap\cite{shinyDepMap_2021} that we introduced in Section \ref{section:Related work}. shinyDepMap is a tool to browse the Depmap portal. 
In both CRISPR and shRNA efficacy data, the algorithms consider “off-target”. So the normalized CRISPR and shRNA data should give similar scores. They combined CRISPR and shRNA screens datasets into a unified “perturbation score”. This score reports two degrees:
\begin{itemize}
\item[1.] which loss of the gene reduces cell growth in sensitive lines(“efficacy”)
\item[2.] Which its essentiality varies across lines(“selectivity”)
\end{itemize}

\subsubsection{Cluster genes with all cells} \label{sec:crispr-all}
\paragraph{GHSOM clustering}

\begin{figure}[htb]
    \centering
    \includegraphics[width=4.5in]{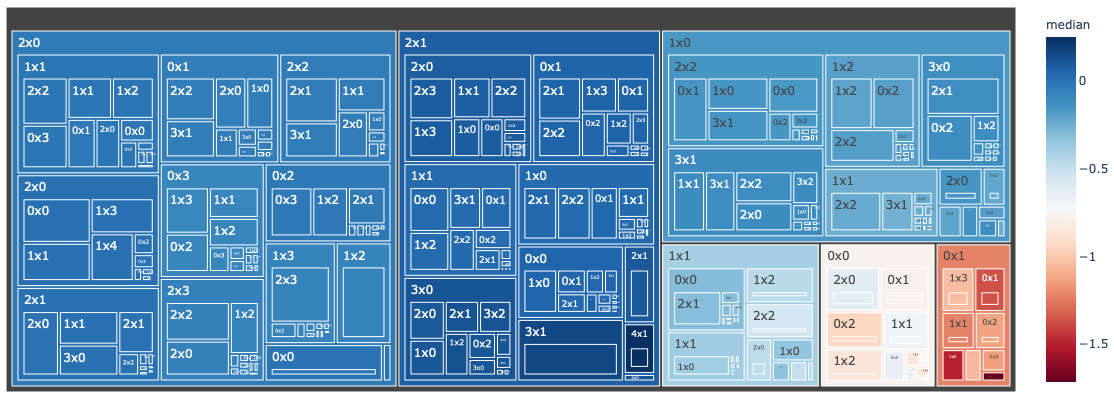}
    \caption{The cluster feature map with color in the median value of attributes on CRISPR dataset (Clustering genes).}
    \label{fig:shiny-all-tree}
\end{figure}
\begin{figure}[htb]
    \centering
    \subfigure[L1 clusters] {
        \includegraphics[width=1.8in]{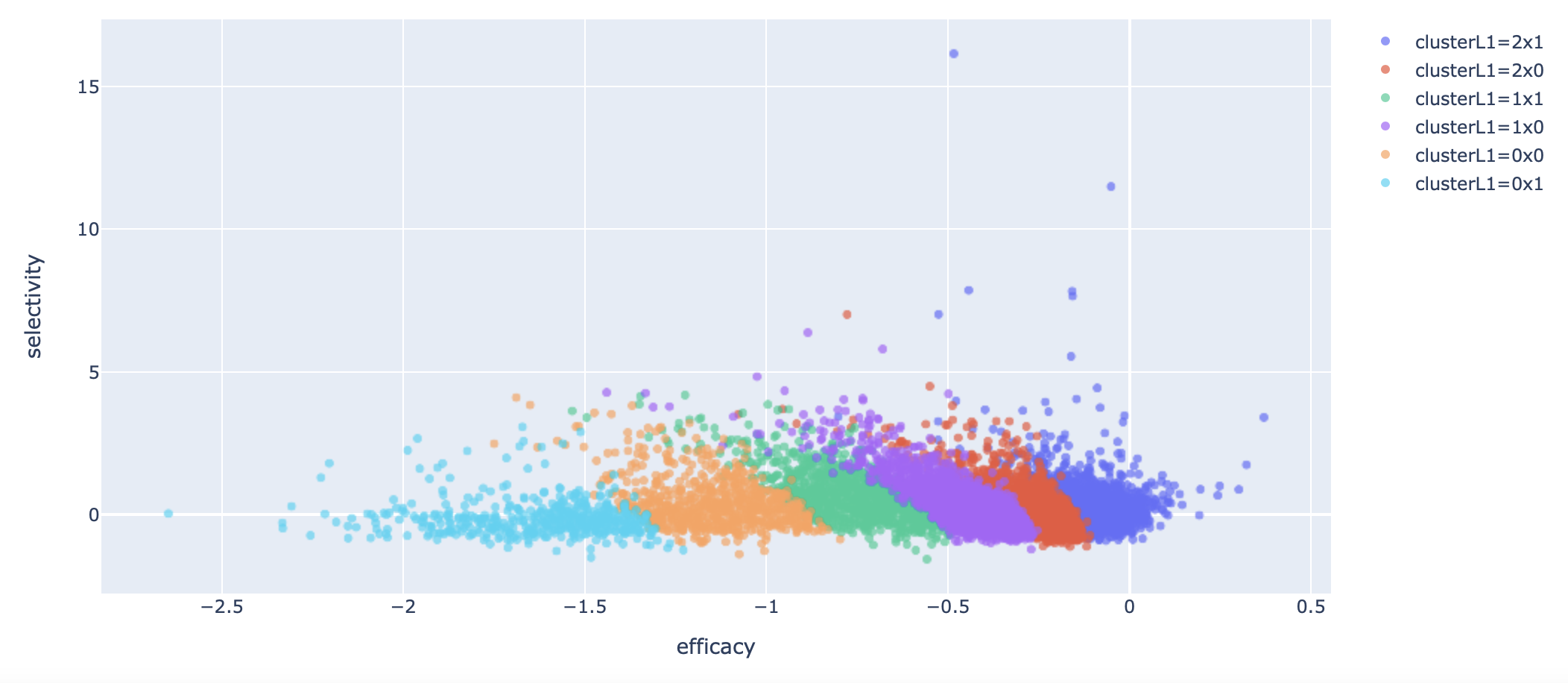}
        \label{fig:all_L1}
    }
    \subfigure[leaf clusters] {
        \includegraphics[width=1.8in]{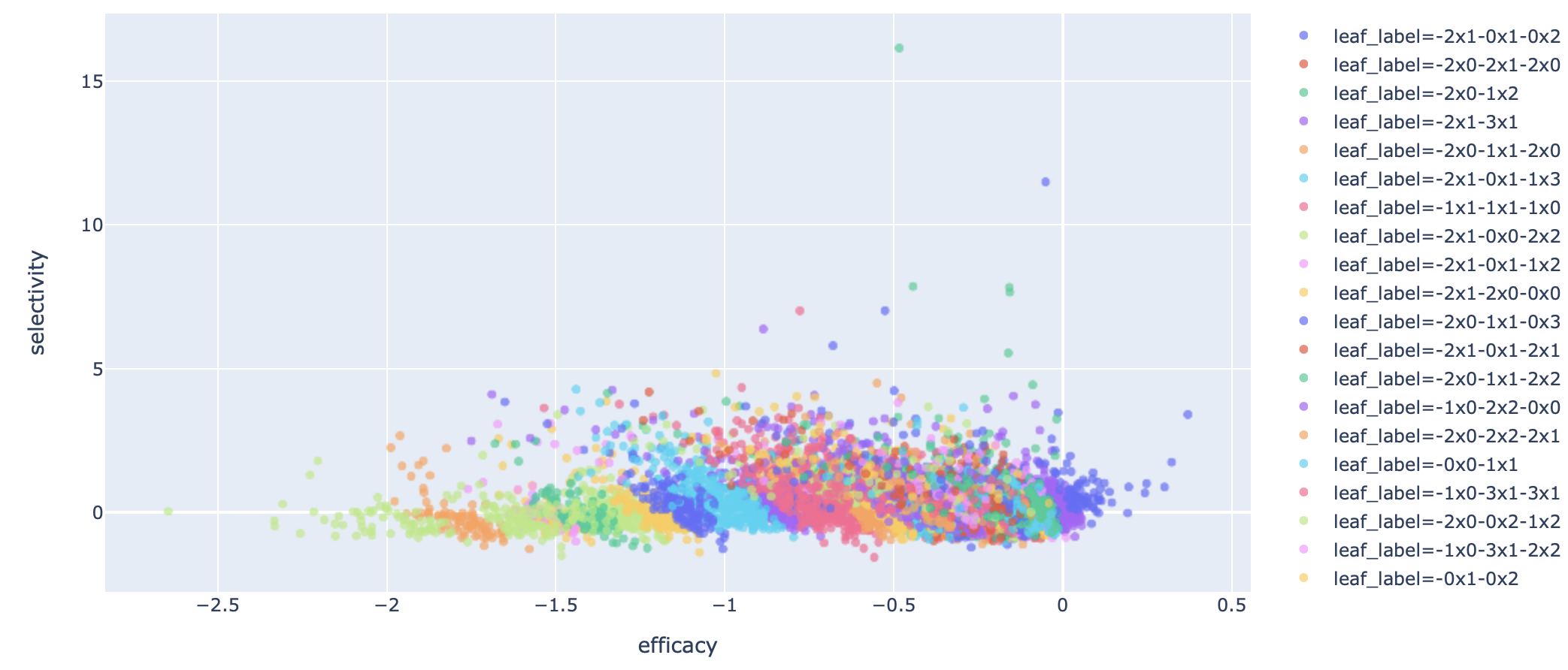}
        \label{fig:all_leaf}
    }
    \subfigure[L2 clusters in L1 0x1] {
        \includegraphics[width=1.8in]{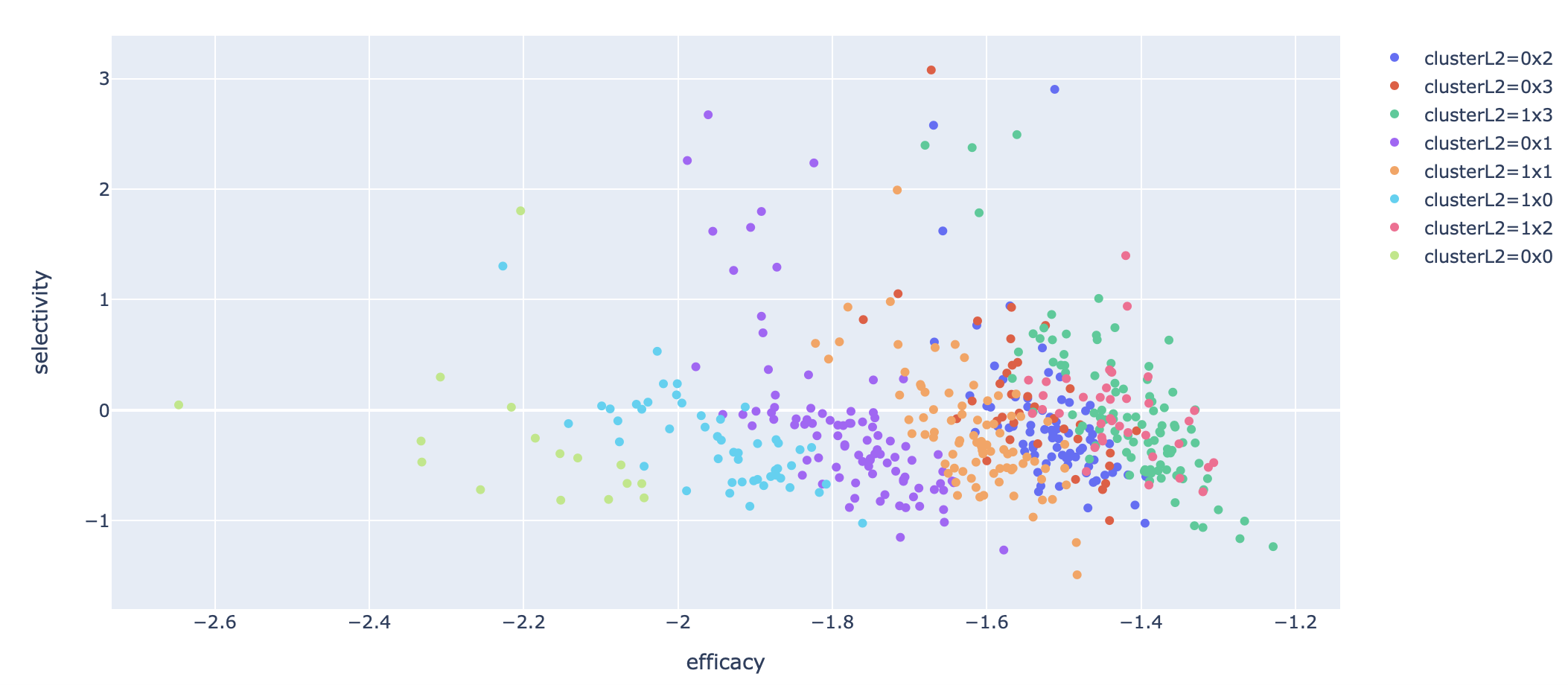}
        \label{fig:all-low-L2}
    }
    \caption{Data projection on Efficacy and Selectivity \textbf{a.} Clustering result with L1 clusters  \textbf{b.} Clustering result with leaf clusters  \textbf{c.} The L2 clusters of the cluster on L1 with the lowest efficacy values(L1 0x1)}.
    \label{fig:all-eff}
\end{figure}
The parameter setting of this case is $tau_1$=0.1, $tau_2$=0.01.
The GHSOM clustering result is visualized with a cluster feature map as Fig.~\ref{fig:shiny-all-tree}. The clustering result is three levels in depth and has six clusters on the L1 result. The color represents the average cell median across attributes. We can observe which cluster has a high or low value through the cluster feature map and quickly get the highest or lowest clusters. Cluster 0x1 has the highest value, and most of the other clusters have the low value.
\par
The smaller the dependency data is, the more dependent relation is between a gene and a cancer cell. So we know that the genes in cluster "0x1" on L1 are highly dependent on cancer cells. Cluster "L1 0x1, L2 0x0" is the cluster with the lowest value. Since genes in it have high dependencies on all the cancer cells, we might infer that they could be the target genes for many cancers.
\par
We cluster genes with all cells as attributes to observe clustering results generally. The result of clustering CRISPR data is three levels deep. As we mentioned in Section \ref{section:crispr}, we referred shinyDepMap\cite{depmap_2017} and mapped our clustering result on the same "Efficacy and selectivity System." The result is shown in Fig.~\ref{fig:all-eff}. On the plot with "Efficacy" and "Selectivity"(Fig.~\ref{fig:all_L1}), our clusters are quite told apart. In other words, GHSOM clusters quite well on CRISPR data. 

\paragraph{Significant Attributes Identification}
We picked five leaf clusters that might be interesting to be researched. Cluster 0x1-0x0 is the cluster with the lowest mean value; cluster 2x1-4x1 is the cluster with the highest mean value. We randomly picked three clusters with common mean values, which are cluster 1x0-3x1-2x0, cluster 2x0-0x0, and cluster 0x0-0x2.
\par
We identified the significant attributes of these five leaf clusters with the Significant Attributes Identification method. The significant attributes of clusters are shown in Table.~\ref{table:sf-shinyleaf}. We found that some attributes are identified as significant in different clusters, such as ACH-000383, ACH-000414, and ACH-000235.
\clearpage
\linespread{1.2}
\begin{table}[htb]
\centering
\begin{tabular}{|c|p{330pt}|} 
\hline 
Leaf cluster & Significant attributes\\
\hline 
2x1-4x1 & ACH-000383, ACH-000783, ACH-000804, ACH-000725, ACH-000910, ACH-000608, ACH-000953, ACH-000414, ACH-000785, ACH-000890\\
\hline
0x1-0x0 & ACH-000115, ACH-000557, ACH-000571, ACH-000294, ACH-000146, ACH-000608, ACH-000005, ACH-000472, ACH-000362, ACH-000724\\
\hline 
1x0-3x1-2x0 & ACH-000383, ACH-000414, ACH-000235, ACH-000259, ACH-000403, ACH-000788, ACH-000624, ACH-000943, ACH-000517, ACH-000146\\
\hline
2x0-0x0 & ACH-000235, ACH-000753, ACH-000274, ACH-000437, ACH-000259, ACH-000383, ACH-000250, ACH-000403, ACH-000517, ACH-000657\\
\hline
0x0-0x2 & ACH-000784, ACH-000917, ACH-000291, ACH-000518, ACH-000943, ACH-000601, ACH-000232, ACH-001048, ACH-000657, ACH-000095\\
\hline
\end{tabular}
\caption{Significant attributes of leaf clusters on CRISPR dataset (Clustering genes).}
\label{table:sf-shinyleaf}
\end{table}

\begin{figure}[htb]
    \centering
    \subfigure[leaf 2x1-4x1] {
        \includegraphics[width=0.3\textwidth]{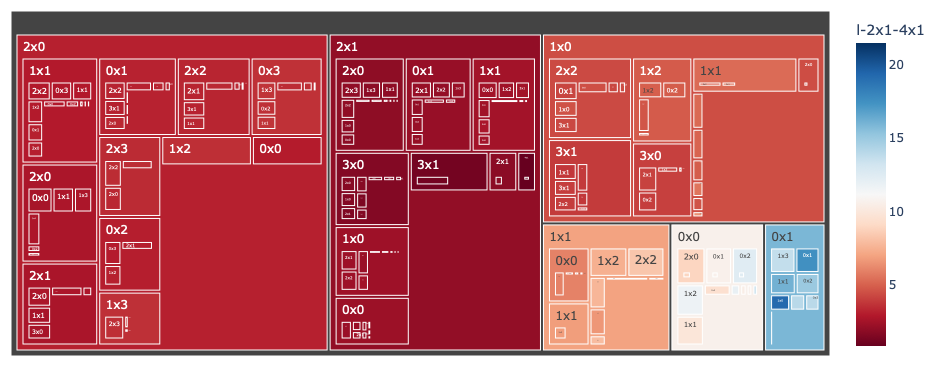}
        \label{fig:a-l2x1-4x1}
    }
    \subfigure[leaf 0x1-0x0] {
        \includegraphics[width=0.3\textwidth]{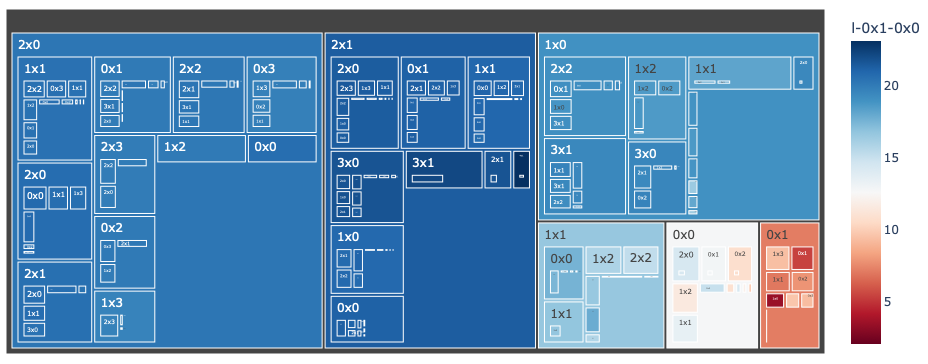}
        \label{fig:a-l0x1-0x0}
    }
    \subfigure[leaf 1x0-3x1-2x0] {
        \includegraphics[width=0.3\textwidth]{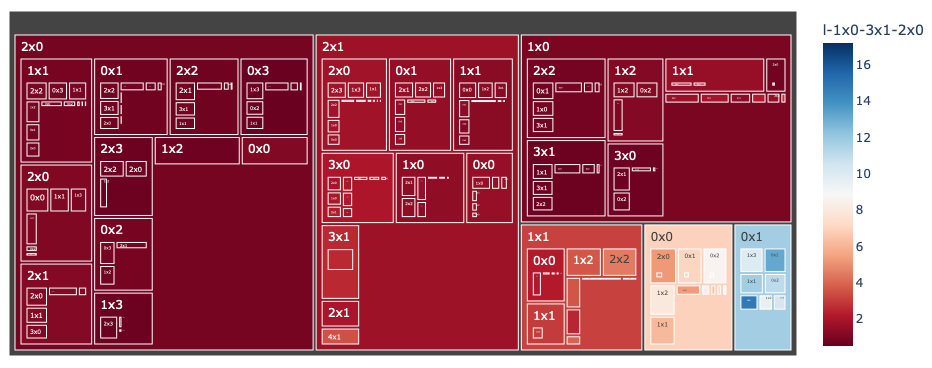}
        \label{fig:a-l1x0-3x1-2x0}
    }
    \subfigure[leaf 2x0-0x0] {
        \includegraphics[width=0.3\textwidth]{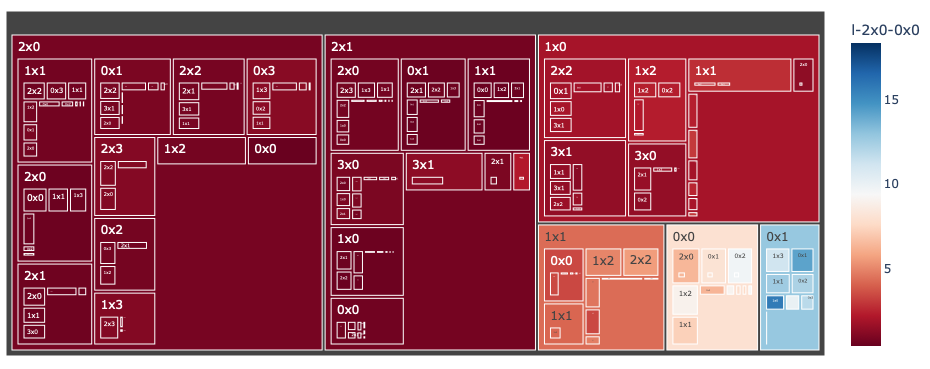}
        \label{fig:a-l2x0-0x0}
    }
    \subfigure[leaf 0x0-0x2] {
        \includegraphics[width=0.3\textwidth]{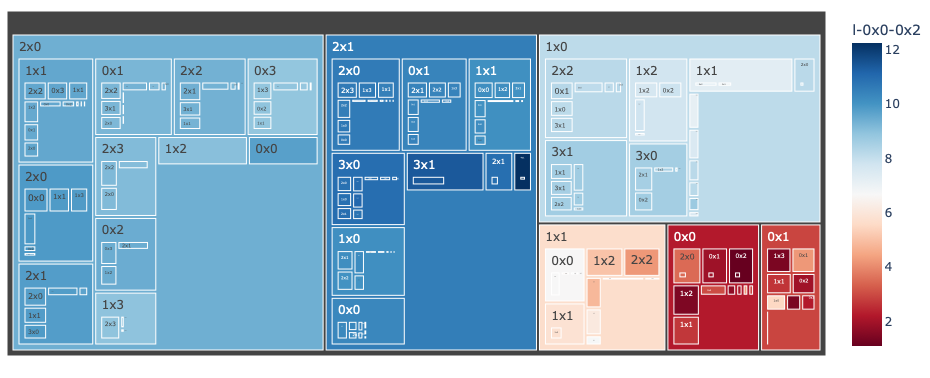}
        \label{fig:a-l0x0-0x2}
    }
    \caption{The cluster feature map with color in value difference for significant attributes of target clusters on CRISPR dataset (Clustering genes).}
    \label{fig:sf-crisprall-dis}
\end{figure}

\paragraph{Cluster feature map}
We calculated the difference between significant attributes in the target cluster and other clusters to check the identified results. The result is showed with cluster feature maps(Fig.~\ref{fig:sf-crisprall-dis}). 
The continuous color in plots represents the difference between significant attributes in the target cluster and other clusters. Blue represents the larger difference, and red represents the smaller difference. Therefore, we expected the target clusters should be the reddest rectangles in these cluster feature maps. 
\par
Through Fig.~\ref{fig:a-l0x0-0x2} and Fig.~\ref{fig:a-l0x1-0x0}, we can know that the significant attributes of these two clusters have dissimilar expression with most of clusters.

\begin{figure}[htb]
    \centering
    \includegraphics[width=3.5in]{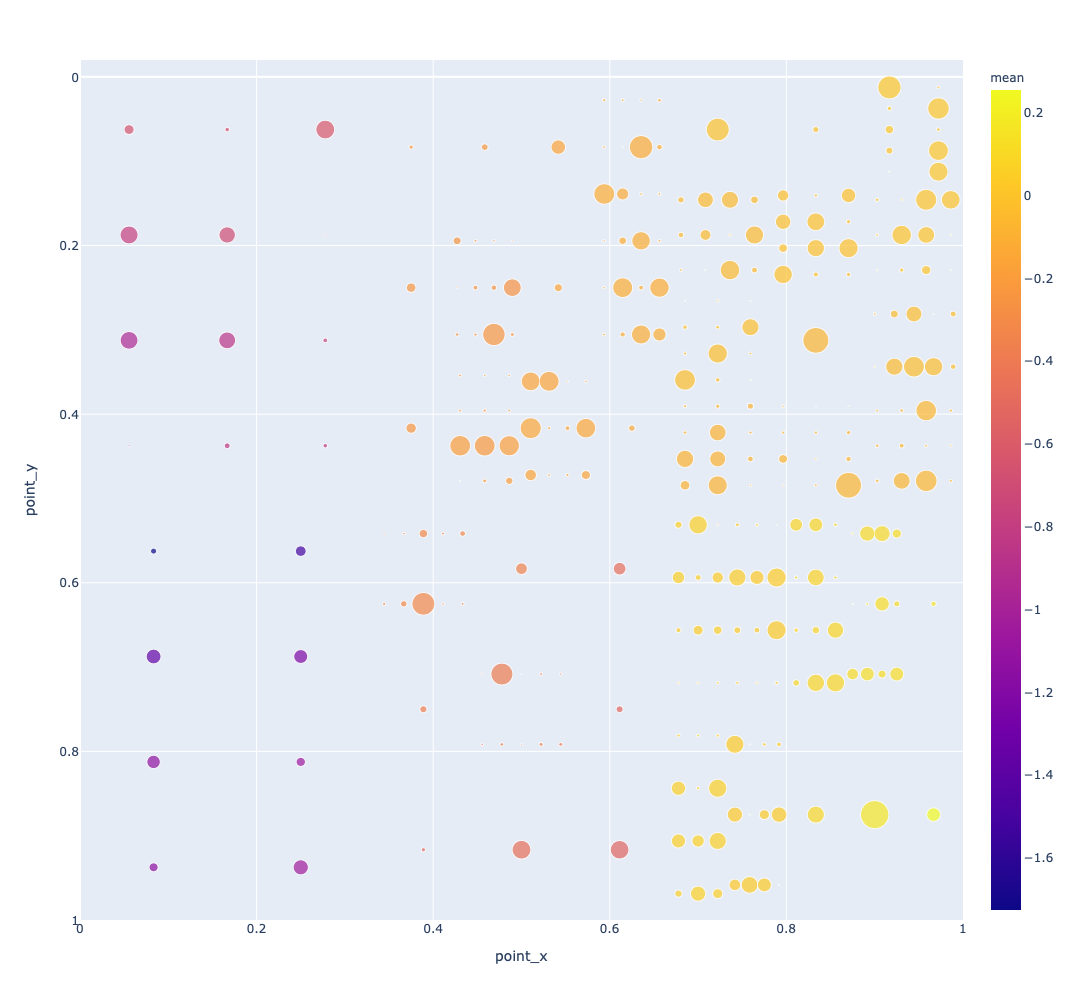}
    \caption{The cluster distribution map with color in the mean value of attributes on CRISPR dataset (Clustering genes).}
    \label{fig:shiny-all-bubble}
\end{figure}

\paragraph{Cluster distribution map: spatial relations of clusters}
Through the cluster distribution map (Fig.~\ref{fig:shiny-all-bubble}), we know the relative positions of every cluster. The colors of the cluster distribution map(Fig.~\ref{fig:shiny-all-bubble}) represent the average expression value of clusters. More yellow represents a higher value, and more blue represents a lower value. The lower the expression value is, the more dependent that gene is on cells. We noticed that the lower values clusters are on the left-downer part of the cluster distribution map. We knew that those clusters have more dependent genes on cells.
\subsubsection{Cluster cells with the lowest value genes}
\paragraph{GHSOM clustering}
Follow the workflow that we introduced in Sec.~ \ref{sec:crispr-workflow}. After Sec.~\ref{sec:crispr-all}, we choose the cluster with the lowest value, which represents the most dependent on cells. Then use the genes of this cluster((L1 0x1, L2 0x0)) as the attributes to cluster all cancer cells. The purpose is to figure out which cancer is highly dependent on these picked genes. The genes in it are ["EEF2 (1938)", "HSPE1 (3336)", "KIF11 (3832)", "KPNB1 (3837)", "POLR2L (5441)", "PRPF19 (27339)", "PSMA3 (5684)", "PSMA6 (5687)", "PSMB3 (5691)", "PSMD7 (5713)", "RAN (5901)", "RPL37 (6167)", "RUVBL1 (8607)", "SNRNP200 (23020)", "SNRPD1 (6632)", "UBA1 (7317)"].
\par
The parameter setting of this case is $tau_1$=0.1, $tau_2$=0.05.
The GHSOM clustering result is visualized with a cluster feature map as Fig.~\ref{fig:shiny-all-tree}. The clustering result is two levels in depth and has 12 clusters on the L1 result. The color represents the average cell median across attributes. We can observe which cluster has a high or low value through the cluster feature map and quickly get the highest or lowest clusters. Cluster 0x1 has the highest value, and cluster 3x2-1x3 has the lowest value.
\begin{figure}[htb]
    \centering
    \includegraphics[width=4.5in]{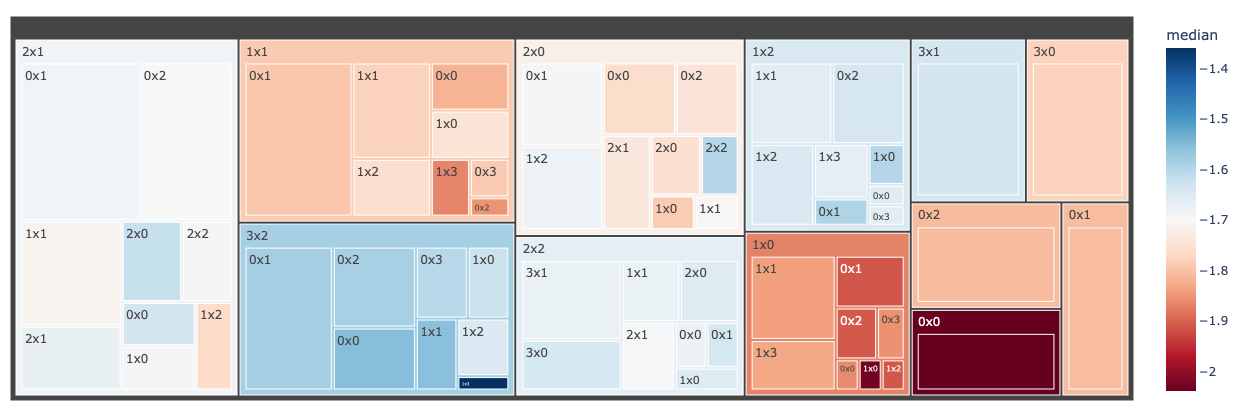}
    \caption{The cluster feature map with color in the median value of attributes on CRISPR dataset (Clustering cells).}
    \label{fig:highest-cells-tree}
\end{figure}
The color of the cluster feature map(Fig.~\ref{fig:highest-cells-tree}) represents the median value of cells across picked genes in clusters. Through the cluster feature map, we know that cluster "L10x0" on L1 has an overall low value, and Cluster "L1 0x0, L2 0x0" is the cluster with the lowest value. Since cluster "L1 0x0, L2 0x0" has only 2 cells, we take the upper level of it to observe the result. The cells in cluster "L1 0x0" are 'ACH-000005', 'ACH-000115', 'ACH-000146', 'ACH-000334', 'ACH-000348', 'ACH-000459', 'ACH-000557', 'ACH-000571', 'ACH-000699', 'ACH-000724', 'ACH-000763', 'ACH-000778', 'ACH-000838',  'ACH-000876'.
\paragraph{Significant Attributes Identification}
We picked five leaf clusters that might be interesting to be researched. Cluster 0x0 is the cluster with the lowest mean value; cluster 1x0-1x0 is the cluster with the second-lowest mean value; cluster 3x2-1x3 is the cluster with the highest mean value. We randomly picked three clusters with ordinary mean values, which are cluster 0x0-0x0-0x2, cluster 1x1-1x3, and cluster 2x1-2x0.
Because there are only sixteen attributes in this clustering case, we identify only five attributes. The significant attributes of clusters are shown in Table.~\ref{table:sf-cellleaf}. Similar to other cases, we found the common significant attributes in different clusters, such as EEF2(1938), HSPE1(3336), and KIF11(3832).
\begin{table}[hbt]
\centering
\resizebox{\textwidth}{!}{
\begin{tabular}{|c|p{330pt}|} 
\hline 
Leaf cluster & Significant attributes \\
\hline 
0x0 & EEF2(1938), HSPE1(3336), KIF11(3832), KPNB1(3837), POLR2L(5441)\\
\hline 
3x2-1x3 & RAN(5901), RPL37(6167), HSPE1(3336), SNRNP200(23020), POLR2L(5441)\\
\hline
1x0-1x0 & EEF2(1938), HSPE1(3336), KIF11(3832), KPNB1(3837), POLR2L(5441)\\
\hline
1x1-1x3 & RPL37(6167), RAN(5901), PSMA6(5687), EEF2(1938), HSPE1(3336)\\
\hline
2x1-2x0 & RAN(5901), KIF11(3832), PSMB3(5691), RPL37(6167), PSMA3(5684)\\
\hline
\end{tabular}
}
\caption{Significant attributes of leaf clusters on CRISPR datasets (Clustering cells).}
\label{table:sf-cellleaf}
\end{table}

\begin{figure}[htb]
    \centering
    \subfigure[leaf 0x0] {
        \includegraphics[width=0.3\textwidth]{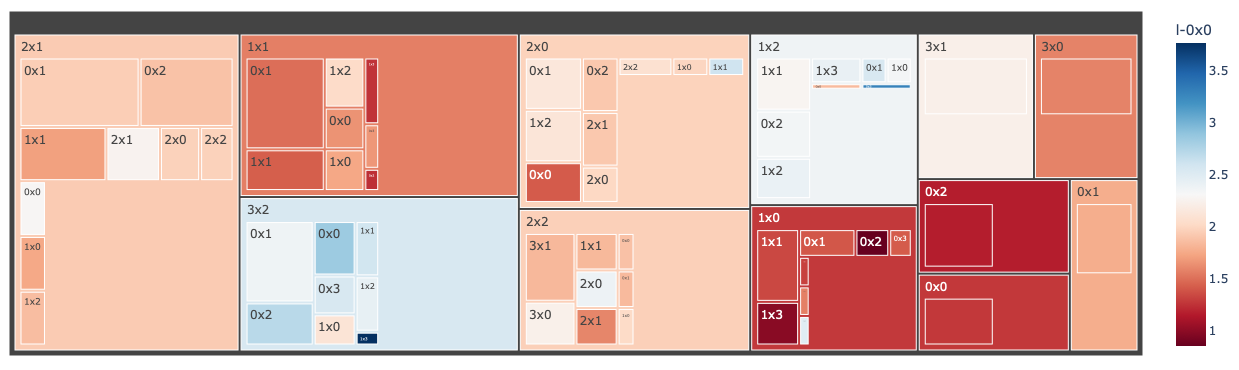}
        \label{fig:c-l0x0}
    }
    \subfigure[leaf 1x0-1x0] {
        \includegraphics[width=0.3\textwidth]{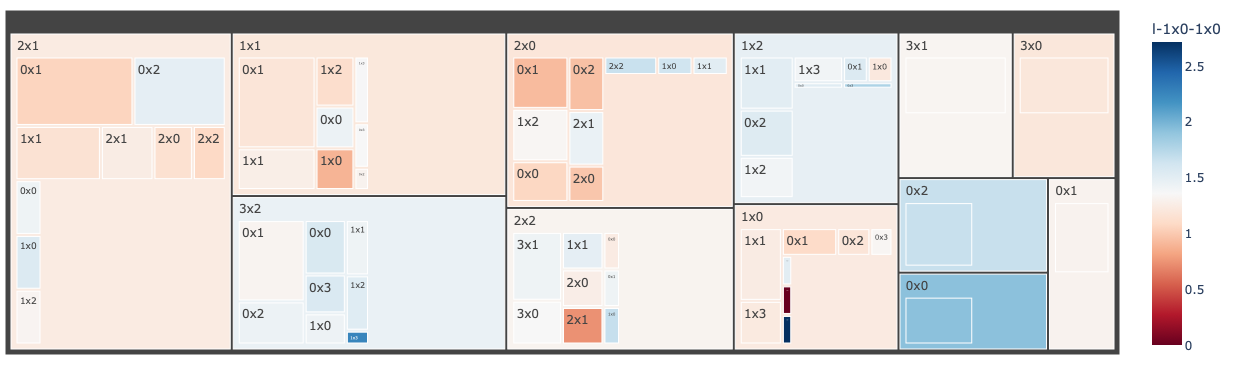}
        \label{fig:c-1x0-1x0}
    }
    \subfigure[leaf 1x1-1x3] {
        \includegraphics[width=0.3\textwidth]{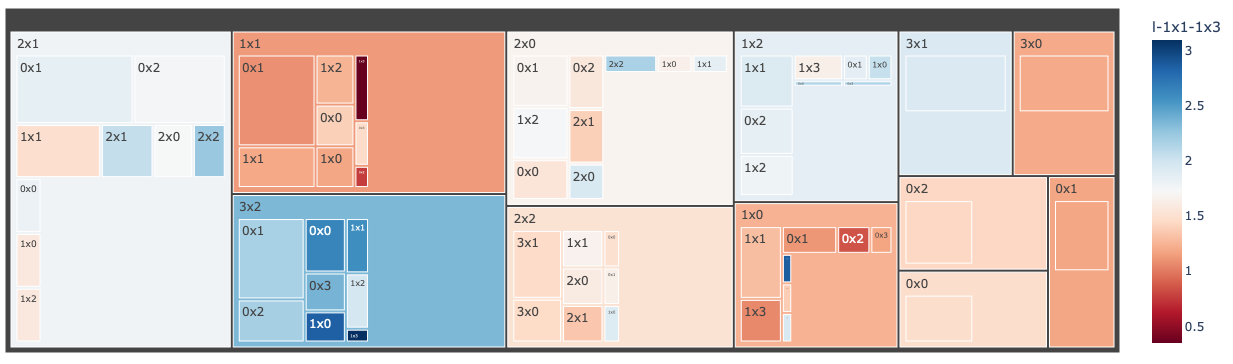}
        \label{fig:c-l1x1-1x3}
    }
    \subfigure[leaf 2x1-2x0] {
        \includegraphics[width=0.3\textwidth]{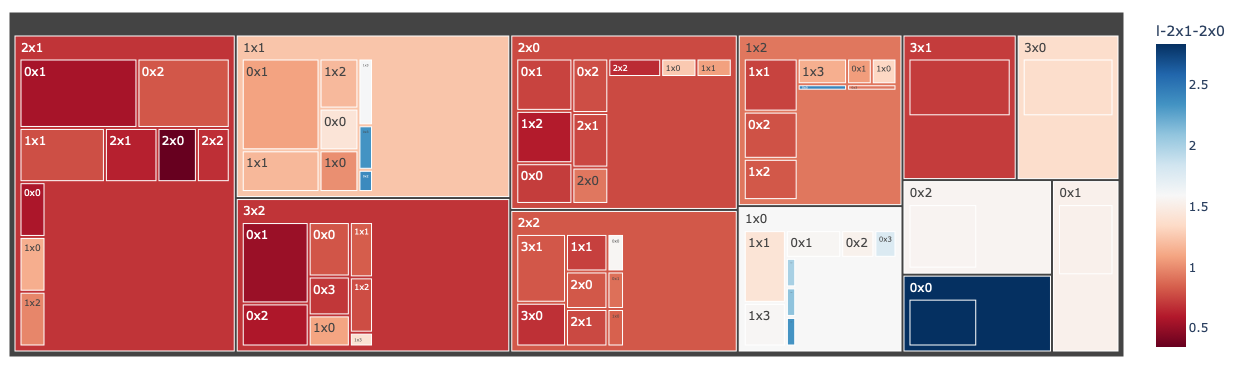}
        \label{fig:c-l2x1-2x0}
    }
     \subfigure[leaf 3x2-1x3] {
        \includegraphics[width=0.3\textwidth]{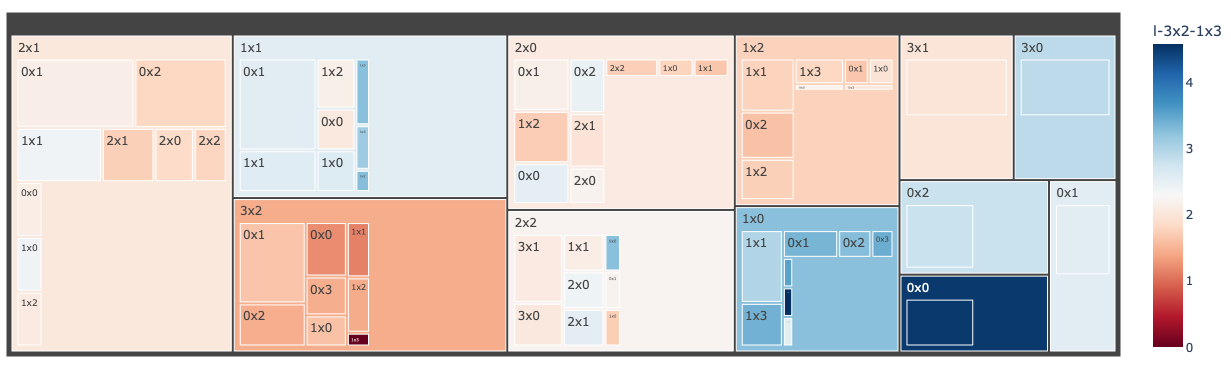}
        \label{fig:c-l3x2-1x3}
    }
    \caption{The cluster feature map with color in value difference for significant attributes of target clusters on CRISPR dataset (Clustering cells).}
    \label{fig:sf-crisprcell-dis}
\end{figure}

\paragraph{Cluster feature map}
We calculated the difference between significant attributes in the target cluster and other clusters to check the identified results. The result is showed with the cluster feature maps(Fig.~\ref{fig:sf-crisprcell-dis}). 
\par
Fig.~\ref{fig:c-l2x1-2x0} is the cluster feature map that has most obvious feature. Most clusters are red, so they have little difference from the significant attributes of Cluster 2x1-2x0. However, only one cluster in blue has a big difference in these attributes, which also happened to Cluster 3x2-1x3's cluster feature map.

\paragraph{Cluster distribution map: spatial relations of clusters}
\begin{figure}[htb]
    \centering
    \subfigure[Color in mean value]{
        \includegraphics[width=2.5in]{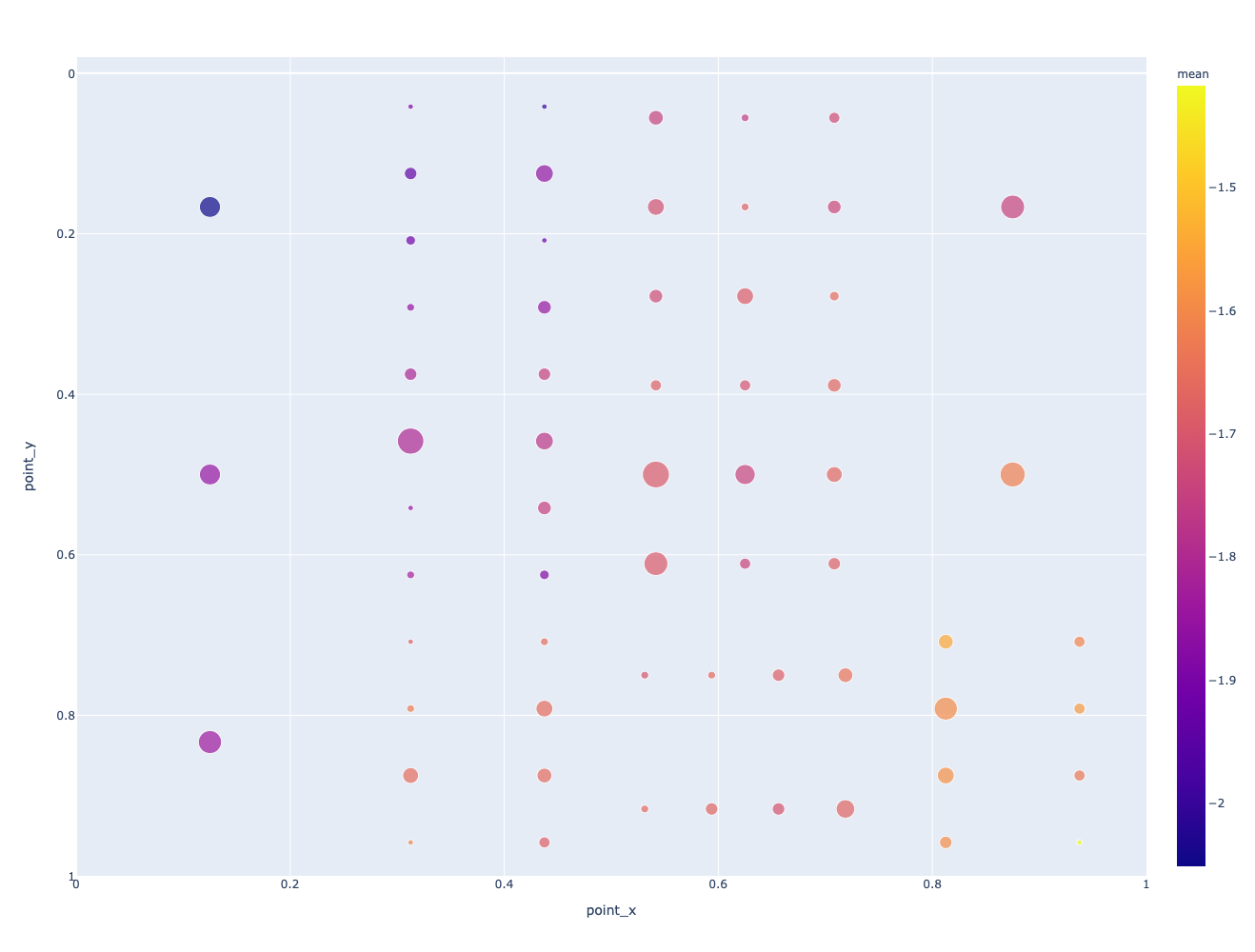}
        \label{fig:cell-bubble-mean}
    }
    \subfigure[Color in disease type]{
        \includegraphics[width=2.5in]{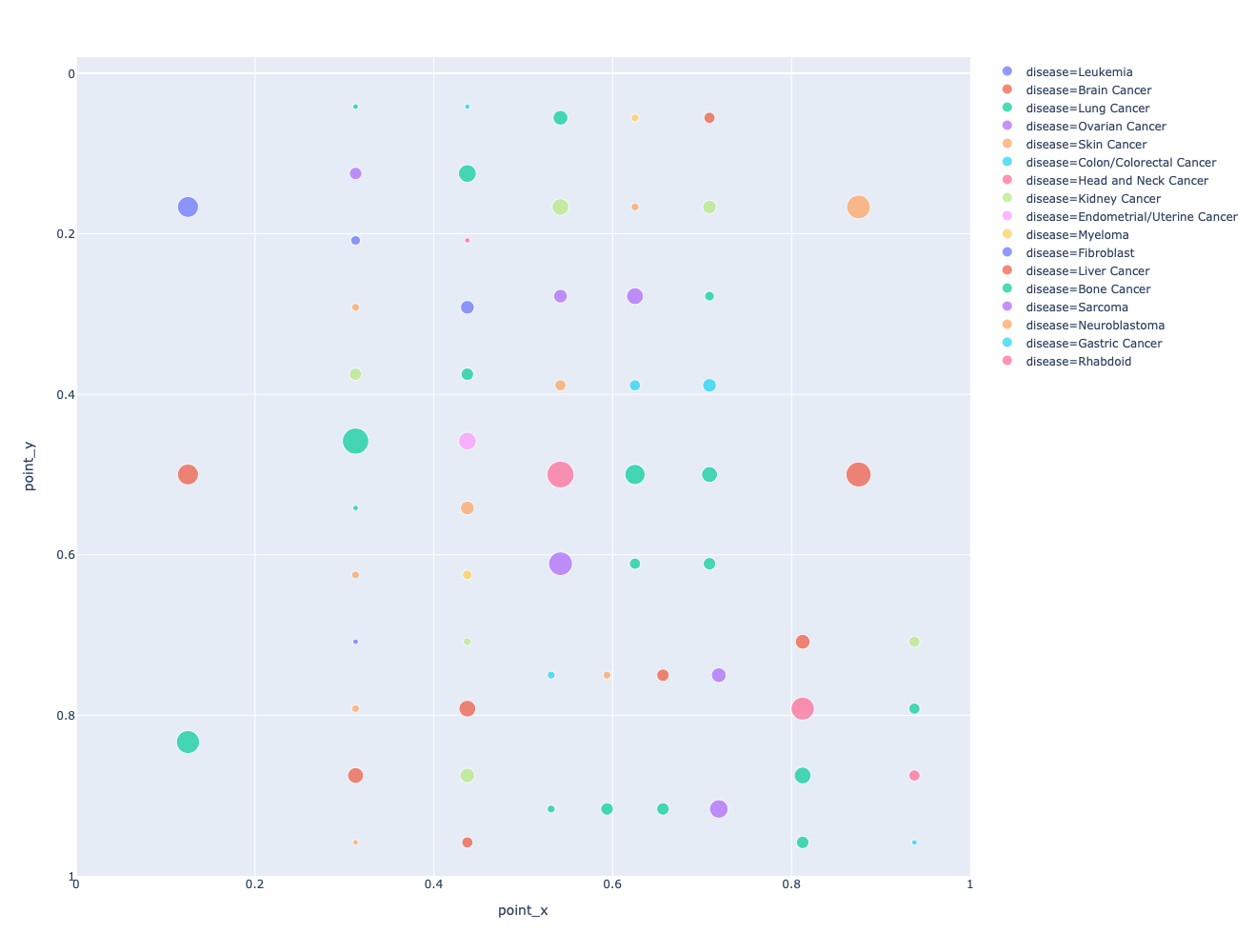}
        \label{fig:cell-bubble-disease}
    }
    
    \caption{The cluster distribution map with color in the mean value of attributes and cell types on CRISPR dataset (Clustering cells)}
    \label{fig:highest-cells-bubble}
\end{figure}
We show two cluster distribution maps with different features. Fig.~\ref{fig:cell-bubble-mean} is presented with average expression value, and Fig.~\ref{fig:cell-bubble-disease} is presented with cell diseases. The colors of the disease one represent diseases of cells. The color saturation shows how pure the disease is in the cluster. The more opaque, the more label cells are in the cluster.
\par
These two cluster distribution maps show that the clusters have lower values belonging to Leukemia, Colon cancer, and Brain cancer. Cluster 0x0 has the lowest value, so we especially list the disease distribution of cluster 0x0 in Table.~\ref{table:disease-0x0}. We might refer that these cancers are highly dependent on the picked genes.
\begin{table}[htb]
\centering
\begin{tabular}{|c|l|} 
\hline 
L2 cluster & Diseases\\
\hline 
0x0 & Prostate Cancer, Brain Cancer\\
\hline 
1x0 & Leukemia, Myeloma, Breast Cancer\\
\hline
2x0 & Skin Cancer, Kidney Cancer\\
\hline
0x1 & Bladder Cancer\\
\hline
1x1 & Lymphoma, Breast Cancer, Head and Neck Cancer\\
\hline
2x1 & Liekemia, Myeloma\\
\hline
\end{tabular}
\caption{The cancer statistics in cluster "L1 0x0"}
\label{table:disease-0x0}
\end{table}

\subsubsection{Cluster genes with lung cells}
\paragraph{GHSOM clustering}
In this case, we would like to cluster data by specific disease, and try to find out target gene that the disease is dependent on. We clustered the genes with only lung cancer cells as attributes. 
\par
\begin{figure}[htb]
    \centering
    \includegraphics[width=5in]{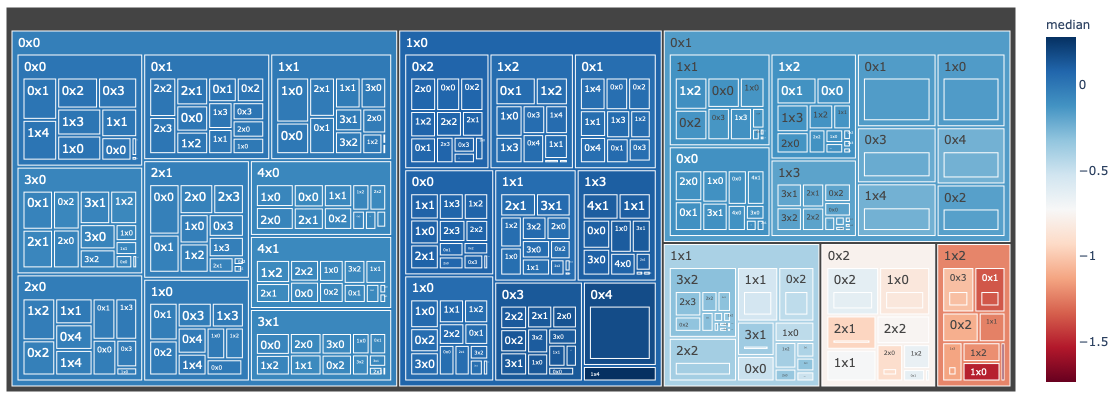}
    \caption{The cluster feature map with color in the median value of attributes on CRISPR dataset(Clustering genes with lung cells).}
    \label{fig:lung-tree}
\end{figure}

In the cluster feature map(Fig.~\ref{fig:lung-tree}), the color represents the median value of genes across only lung cancer cells in clusters. Through the cluster feature map, we know that cluster "1x2" on L1 has an overall low value, and Cluster "L1 1x2, L2 0x0" is the cluster with the lowest value. The genes in it are ["EEF2 (1938)", "HSPE1 (3336)", "KIF11 (3832)", "KPNB1 (3837)", "POLR2L (5441)", "PRPF19 (27339)", "PSMA3 (5684)", "PSMB3 (5691)", "PSMD7 (5713)", "RAN (5901)", "SNRNP200 (23020)", "SNRPD1 (6632)", "UBA1 (7317)"]. These genes have the lowest value, meaning they highly depend on lung cancer cells. We might infer that these genes could be considered the target genes for lung cancer.
\linespread{1.2}
\begin{table}[htb]
\centering
\begin{tabular}{|c|p{330pt}|} 
\hline 
Leaf cluster & Significant attributes\\
\hline 
1x2-0x0 & ACH-000869, ACH-000414, ACH-000681, ACH-000867, ACH-000311, ACH-000290, ACH-000785, ACH-000841, ACH-000264, ACH-000781\\
\hline 
1x0-1x4 & ACH-000785, ACH-000414, ACH-000264, ACH-000390, ACH-000840, ACH-000337, ACH-000030, ACH-000395, ACH-000251, ACH-000528\\
\hline
0x0-0x0-0x2 & ACH-000414, ACH-000867, ACH-000781, ACH-000639, ACH-000916, ACH-000585, ACH-000030, ACH-000510, ACH-000901, ACH-000785\\
\hline
0x2-2x1 & ACH-000867, ACH-000767, ACH-000481, ACH-000290, ACH-000448, ACH-000414, ACH-000858, ACH-000921, ACH-000866, ACH-000869\\
\hline
1x1-0x2 & ACH-000414, ACH-000900, ACH-000681, ACH-000853, ACH-000769, ACH-000861, ACH-000912, ACH-000638, ACH-000851, ACH-000916\\
\hline
\end{tabular}
\caption{Significant attributes of leaf clusters on CRISPR dataset (Clustering genes with lung cells).}
\label{table:sf-lungleaf}
\end{table}
\paragraph{Significant Attributes Identification}
We picked five leaf clusters that might be interesting to be researched. Cluster 1x0-1x4 is the cluster with the highest mean value; cluster 1x2-0x0 is the cluster with the lowest mean value. We randomly picked three clusters with ordinary mean values, which are cluster 0x0-0x0-0x2, cluster 0x2-2x1, and cluster 1x1-0x2.
We identified the significant attributes of these five leaf clusters with the Significant Attributes Identification method. The significant attributed genes of clusters are shown in Table.~\ref{table:sf-lungleaf}.

\begin{figure}[hp]
    \centering
    \subfigure[leaf 1x0-1x4] {
        \includegraphics[width=0.3\textwidth]{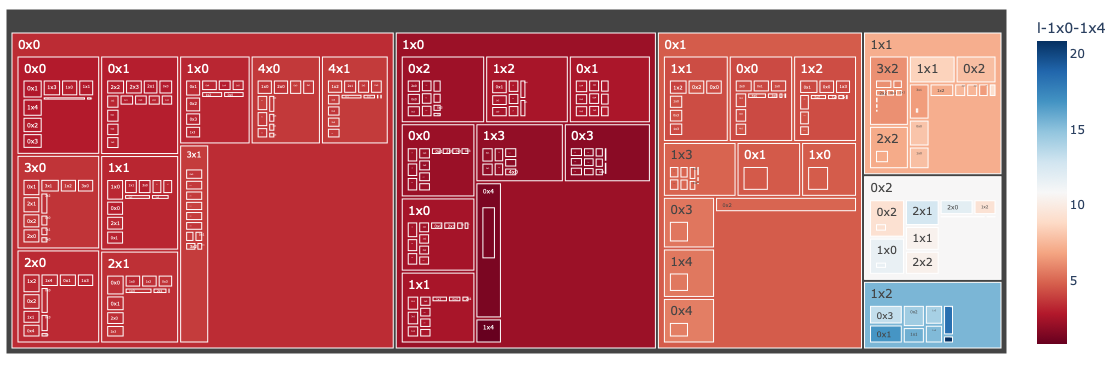}
        \label{fig:l-l1x0-1x4}
    }
    \subfigure[leaf 1x2-0x0] {
        \includegraphics[width=0.3\textwidth]{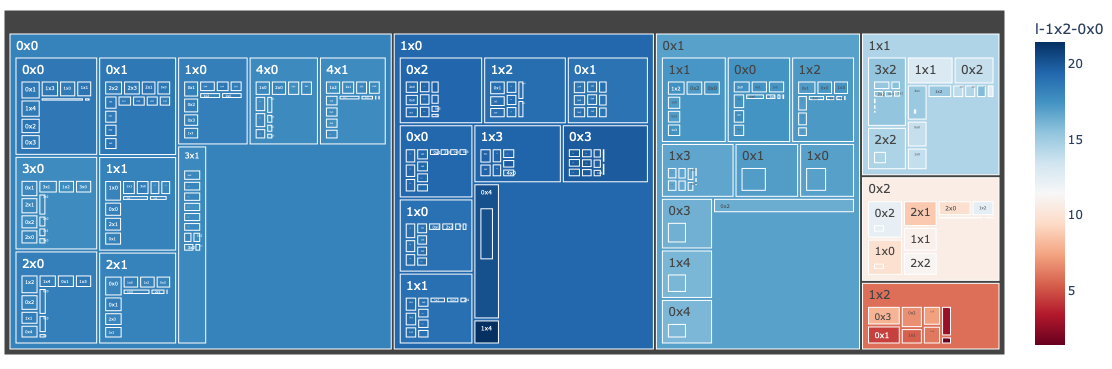}
        \label{fig:l-l1x2-0x0}
    }
    \subfigure[leaf 0x0-0x0-0x2] {
        \includegraphics[width=0.3\textwidth]{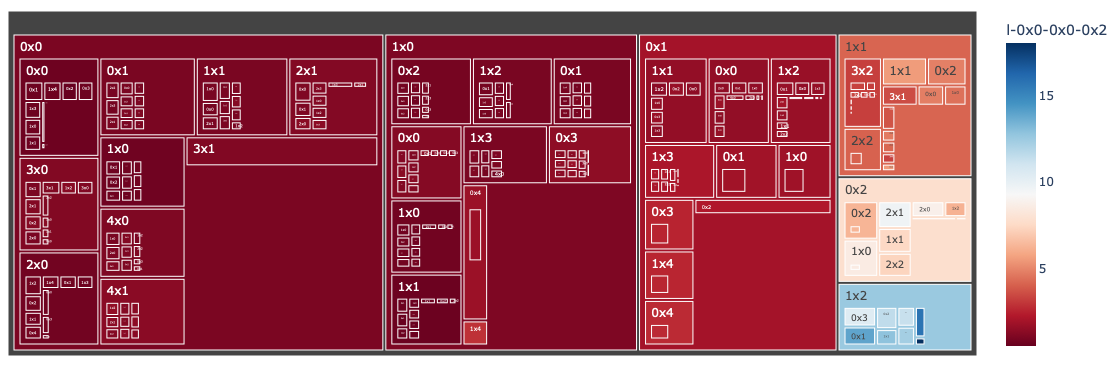}
        \label{fig:l-l0x0-0x0-0x2}
    }
    \subfigure[leaf 0x2-2x1] {
        \includegraphics[width=0.3\textwidth]{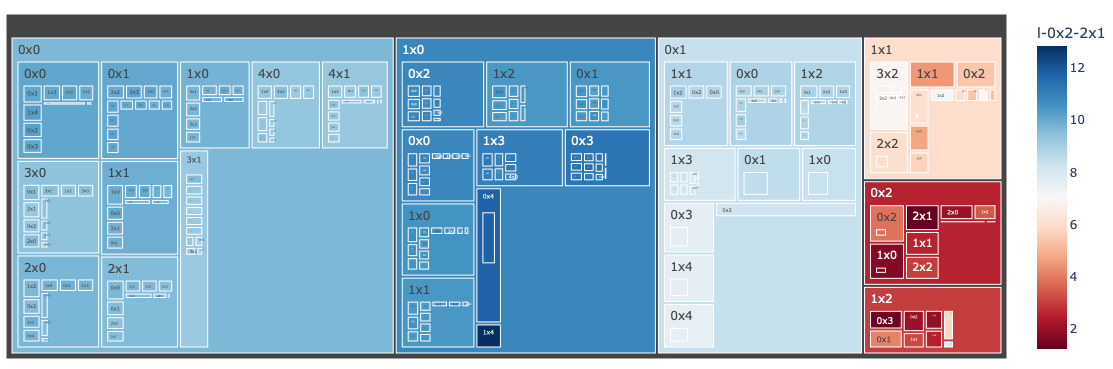}
        \label{fig:l-l0x2-2x1}
    }
    \subfigure[leaf 1x1-0x2] {
        \includegraphics[width=0.3\textwidth]{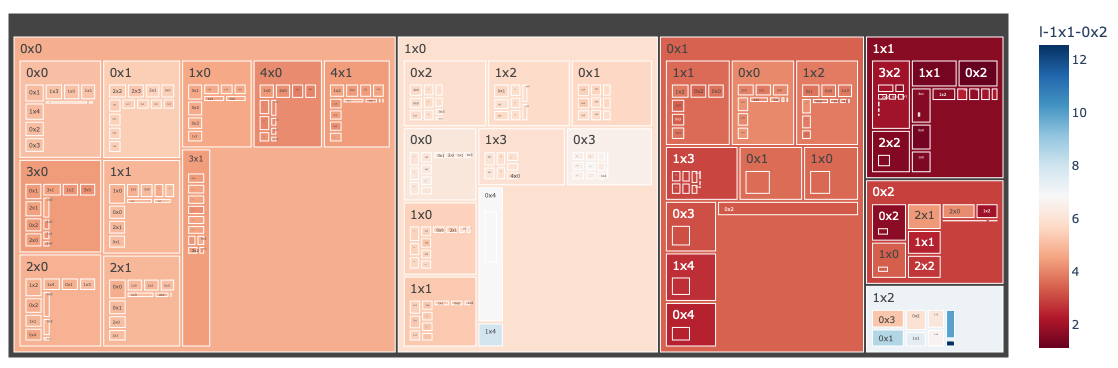}
        \label{fig:l-l1x1-0x2}
    }
    
    \caption{The cluster feature map with color in value difference for significant attributes of target clusters on CRISPR dataset (Clustering genes with lung cells).}
    \label{fig:sf-crisprlung-dis}
\end{figure}

\paragraph{Cluster feature map}
We calculated the difference between significant attributes in the target cluster and other clusters to check the identified results. The result is showed with cluster feature maps(Fig.~\ref{fig:sf-crisprlung-dis}). 
\par
We noticed that the cluster feature maps have two types of performance. Fig.~\ref{fig:l-l1x0-1x4}, Fig.~\ref{fig:l-l0x0-0x0-0x2}, and Fig.~\ref{fig:l-l1x1-0x2} are have similar distribution and most red in the graph. Thus, most of clusters' value has little difference to their significant attributes. In the other hand, Fig.~\ref{fig:l-l1x2-0x0} and Fig.~\ref{fig:l-l0x2-2x1} have most blue in the graphs. Most of clusters have big difference to their significant attributes.
\paragraph{Cluster distribution map: spatial relations of clusters}
\begin{figure}[htb]
    \centering
    \includegraphics[width=3.5in]{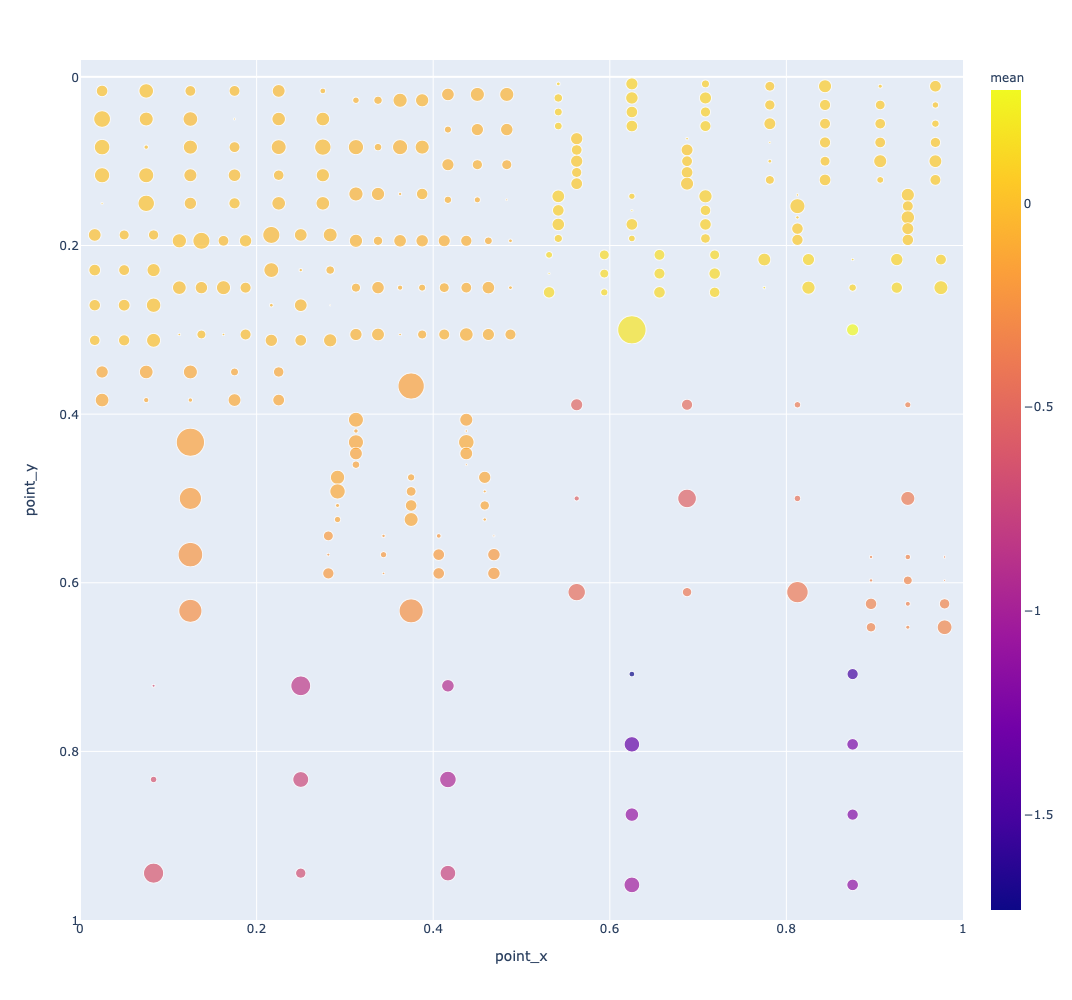}
    \caption{The cluster distribution map with color in the mean value of attributes on CRISPR dataset (Clustering genes with lung cells)}
    \label{fig:shiny-lung-bubble}
\end{figure}

\par
Through the cluster distribution map (Fig.~\ref{fig:shiny-lung-bubble}), we knew the relative positions of every cluster. The colors of cluster distribution map(Fig.~\ref{fig:shiny-lung-bubble}) represent average expression value of clusters. The lower the expression value is, the more dependent on genes. We can see that the clusters that have lower values are on the left-downer part of the cluster distribution map. The low values cluster is at the bottom of the cluster distribution map. We can directly focus on the genes of those clusters to research lung cancer.

\section{Conclusion}
We apply unsupervised clustering to reveal relations of features on cell-gene and single-cell databases. Our analysis consists of three steps: we first conduct GHSOM to cluster samples in a hierarchical structure, such that the self-growth structure of clusters satisfies the required variations between and within. Unlike KNN, K-means, or SOMs that need the number of clusters specified, GHSOM grows as a treemap in a self fashion. Second, we propose a novel $Significant Attributes Identification$ Algorithm to identify significant attributes of clusters. The identified attributes remain the least cluster variation within but most variation between clusters. These significant attributes suppose to affect the clustering result the most. Third, we propose two visualization maps. The \textit{Cluster Feature Map} shows the GHSOM map in grids, coloring each cluster according to the value of a specific feature, such as the value of the significant attribute. The map helps us identify the uniqueness of the features of target clusters. GHSOM clustering result is spatial related. The \textit{Cluster Distribution Map} maps the position of every leaf cluster on it, showing the spatial relations of leaf clusters. Through these two visualization maps, we observe and reveal GHSOM clustering result without dimensional reduction methods. 
\par
Biology data usually have a large amount. Therefore, we need a superb clustering method to categorize and reveal data features. GHSOM is a hierarchical clustering method that we can observe the result in the big picture and details. We apply our analysis to four biology datasets: three single-cell datasets and one CRISPR data (cell-gene database). Then we report our findings accordingly. We also compare GHSOM clustering performance with other clustering methods. We evaluate clustering methods with internal and external evaluation, which are CH and ARI scores. GHSOM performs well even is the best performance in internal evaluation. In external evaluation, GHSOM has the third best performance of all methods. We also improve the visualization of GHSOM clustering results by cluster feature map and cluster distribution map. These visualizations allow us to observe features of clustering results and data more efficiently and clearly.

\nocite{nocite1}
\nocite{nocite2}
\nocite{nocite3}
\nocite{nocite4}
\nocite{nocite5}
\nocite{nocite6}
\nocite{nocite7}
\nocite{nocite8}
\nocite{nocite9}
\nocite{nocite10}
\nocite{nocite11}
\nocite{nocite12}
\nocite{nocite13}
\nocite{nocite14}
\nocite{nocite15}
\nocite{nocite16}

\bibliographystyle{unsrt}
\bibliography{ref}

\begin{thebibliography}{10}

\bibitem{seurat_2015}
Rahul Satija, Jeffrey~A Farrell, David Gennert, Alexander~F Schier, and Aviv Regev.
\newblock Spatial reconstruction of single-cell gene expression data.
\newblock {\em Nature biotechnology}, 33(5):495--502, 2015.

\bibitem{SC3_2017}
Vladimir~Yu Kiselev, Kristina Kirschner, Michael~T Schaub, Tallulah Andrews, Andrew Yiu, Tamir Chandra, Kedar~N Natarajan, Wolf Reik, Mauricio Barahona, Anthony~R Green, and Martin Hemberg.
\newblock {SC3}: consensus clustering of single-cell {RNA-seq} data.
\newblock {\em Nat. Methods}, 14(5):483--486, May 2017.

\bibitem{CIDR_2017}
Peijie Lin, Michael Troup, and Joshua~WK Ho.
\newblock Cidr: Ultrafast and accurate clustering through imputation for single-cell rna-seq data.
\newblock {\em Genome biology}, 18(1):1--11, 2017.

\bibitem{som_1990}
T.~{Kohonen}.
\newblock The self-organizing map.
\newblock {\em Proceedings of the IEEE}, 78(9):1464--1480, 1990.

\bibitem{ghsom_2000}
Michael Dittenbach, Dieter Merkl, and Andreas Rauber.
\newblock Growing hierarchical self-organizing map.
\newblock {\em Proceedings of the International Joint Conference on Neural Networks}, 6:15 -- 19 vol.6, 02 2000.

\bibitem{KNN_1992}
Naomi~S Altman.
\newblock An introduction to kernel and nearest-neighbor nonparametric regression.
\newblock {\em The American Statistician}, 46(3):175--185, 1992.

\bibitem{kmeans_1982}
S.~{Lloyd}.
\newblock Least squares quantization in pcm.
\newblock {\em IEEE Transactions on Information Theory}, 28(2):129--137, 1982.

\bibitem{ACCENSE}
Karthik Shekhar, Petter Brodin, Mark~M Davis, and Arup~K Chakraborty.
\newblock Automatic classification of cellular expression by nonlinear stochastic embedding (accense).
\newblock {\em Proceedings of the National Academy of Sciences}, 111(1):202--207, 2014.

\bibitem{tsne}
Laurens Van~der Maaten and Geoffrey Hinton.
\newblock Visualizing data using t-sne.
\newblock {\em Journal of machine learning research}, 9(11), 2008.

\bibitem{X-shift}
Nikolay Samusik, Zinaida Good, Matthew~H Spitzer, Kara~L Davis, and Garry~P Nolan.
\newblock Automated mapping of phenotype space with single-cell data.
\newblock {\em Nature methods}, 13(6):493--496, 2016.

\bibitem{Phenograph}
Jacob~H Levine, Erin~F Simonds, Sean~C Bendall, Kara~L Davis, D~Amir El-ad, Michelle~D Tadmor, Oren Litvin, Harris~G Fienberg, Astraea Jager, Eli~R Zunder, et~al.
\newblock Data-driven phenotypic dissection of aml reveals progenitor-like cells that correlate with prognosis.
\newblock {\em Cell}, 162(1):184--197, 2015.

\bibitem{louvain}
Vincent~D Blondel, Jean-Loup Guillaume, Renaud Lambiotte, and Etienne Lefebvre.
\newblock Fast unfolding of communities in large networks.
\newblock {\em Journal of statistical mechanics: theory and experiment}, 2008(10):P10008, 2008.

\bibitem{flowmeans}
Nima Aghaeepour, Radina Nikolic, Holger~H Hoos, and Ryan~R Brinkman.
\newblock Rapid cell population identification in flow cytometry data.
\newblock {\em Cytometry Part A}, 79(1):6--13, 2011.

\bibitem{flowmerge}
G~Finak, A~Bashasharti, R~Brinkmann, and R~Gottardo.
\newblock Merging mixture model components for improved cell population identification in high throughput flow cytometry data.
\newblock {\em Advances in Bioinformatics}, 100, 2009.

\bibitem{flowsom}
Sofie Van~Gassen, Britt Callebaut, Mary~J Van~Helden, Bart~N Lambrecht, Piet Demeester, Tom Dhaene, and Yvan Saeys.
\newblock Flowsom: Using self-organizing maps for visualization and interpretation of cytometry data.
\newblock {\em Cytometry Part A}, 87(7):636--645, 2015.

\bibitem{depeche}
Axel Theorell, Yenan~Troi Bryceson, and Jakob Theorell.
\newblock Determination of essential phenotypic elements of clusters in high-dimensional entities—depeche.
\newblock {\em PLoS One}, 14(3):e0203247, 2019.

\bibitem{slm}
Ludo Waltman and Nees~Jan Van~Eck.
\newblock A smart local moving algorithm for large-scale modularity-based community detection.
\newblock {\em The European physical journal B}, 86(11):1--14, 2013.

\bibitem{hgc_2021}
Ziheng Zou, Kui Hua, and Xuegong Zhang.
\newblock Hgc: fast hierarchical clustering for large-scale single-cell data.
\newblock {\em Bioinformatics}, 37(21):3964--3965, 2021.

\bibitem{pairSampling_2018}
Thomas Bonald, Bertrand Charpentier, Alexis Galland, and Alexandre Hollocou.
\newblock Hierarchical graph clustering using node pair sampling.
\newblock 2018.

\bibitem{shinyDepMap_2021}
Kenichi Shimada, John Bachman, Jeremy Muhlich, and Timothy Mitchison.
\newblock shinydepmap, a tool to identify targetable cancer genes and their functional connections from cancer dependency map data.
\newblock {\em eLife}, 10, 02 2021.

\bibitem{CRISPR_2019}
Fiona~M Behan, Francesco Iorio, Gabriele Picco, Emanuel Gon{\c{c}}alves, Charlotte~M Beaver, Giorgia Migliardi, Rita Santos, Yanhua Rao, Francesco Sassi, Marika Pinnelli, et~al.
\newblock Prioritization of cancer therapeutic targets using crispr--cas9 screens.
\newblock {\em Nature}, 568(7753):511--516, 2019.

\bibitem{rnai}
Neema Agrawal, PVN Dasaradhi, Asif Mohmmed, Pawan Malhotra, Raj~K Bhatnagar, and Sunil~K Mukherjee.
\newblock Rna interference: biology, mechanism, and applications.
\newblock {\em Microbiology and molecular biology reviews}, 67(4):657--685, 2003.

\bibitem{depmap_2017}
Aviad Tsherniak, Francisca Vazquez, Phil~G Montgomery, Barbara~A Weir, Gregory Kryukov, Glenn~S Cowley, Stanley Gill, William~F Harrington, Sasha Pantel, John~M Krill-Burger, et~al.
\newblock Defining a cancer dependency map.
\newblock {\em Cell}, 170(3):564--576, 2017.

\bibitem{vienna_ghsom}
Dieter~Merkl M.~D. E. P. Andreas~Rauber.
\newblock The growing hierarchical self- organizing map.

\bibitem{ghsom_algo}
Esteban~J Palomo, Enrique Dom{\'\i}nguez, Rafael~Marcos Luque, and Jos{\'e} Mu{\~n}oz.
\newblock An intrusion detection system based on hierarchical self-organization.
\newblock In {\em Proceedings of the International Workshop on Computational Intelligence in Security for Information Systems CISIS’08}, pages 139--146. Springer, 2009.

\bibitem{SquarifiedTreemap_2000}
Mark Bruls, C.~Huizing, and J.~V. Wijk.
\newblock Squarified treemaps.
\newblock In {\em VisSym}, 2000.

\bibitem{plotly}
Plotly~Technologies Inc.
\newblock Collaborative data science, 2015.

\bibitem{comparisonFramework_2019}
Xiao Liu, Song Weichen, Brandon Wong, Ting Zhang, Shunying Yu, Guan Lin, and Xianting Ding.
\newblock A comparison framework and guideline of clustering methods for mass cytometry data.
\newblock {\em Genome Biology}, 20, 12 2019.

\bibitem{CH}
Ujjwal Maulik and Sanghamitra Bandyopadhyay.
\newblock Performance evaluation of some clustering algorithms and validity indices.
\newblock {\em IEEE Transactions on pattern analysis and machine intelligence}, 24(12):1650--1654, 2002.

\bibitem{ARI}
Jorge~M Santos and Mark Embrechts.
\newblock On the use of the adjusted rand index as a metric for evaluating supervised classification.
\newblock In {\em International conference on artificial neural networks}, pages 175--184. Springer, 2009.

\bibitem{2019DimensionalityReduction}
E.~Becht, L.~McInnes, John Healy, C.~Dutertre, I.~Kwok, L.~Ng, F.~Ginhoux, and E.~Newell.
\newblock Dimensionality reduction for visualizing single-cell data using umap.
\newblock {\em Nature Biotechnology}, 37:38--44, 2019.

\bibitem{STUART20191888}
Tim Stuart, Andrew Butler, Paul Hoffman, Christoph Hafemeister, Efthymia Papalexi, William~M Mauck~III, Yuhan Hao, Marlon Stoeckius, Peter Smibert, and Rahul Satija.
\newblock Comprehensive integration of single-cell data.
\newblock {\em Cell}, 177(7):1888--1902, 2019.

\bibitem{nocite1}
Jelili Oyelade, Itunuoluwa Isewon, Funke Oladipupo, Olufemi Aromolaran, Efosa Uwoghiren, Faridah Ameh, Moses Achas, and Ezekiel Adebiyi.
\newblock Clustering algorithms: Their application to gene expression data.
\newblock {\em Bioinformatics and Biology Insights}, 10:BBI.S38316, 2016.
\newblock PMID: 27932867.

\bibitem{nocite2}
Mayra~Z Rodriguez, Cesar~H Comin, Dalcimar Casanova, Odemir~M Bruno, Diego~R Amancio, Luciano da~F Costa, and Francisco~A Rodrigues.
\newblock Clustering algorithms: A comparative approach.
\newblock {\em PloS one}, 14(1):e0210236, 2019.

\bibitem{nocite3}
Harun Pirim, Burak Ek{\c{s}}io{\u{g}}lu, Andy~D Perkins, and {\c{C}}etin Y{\"u}ceer.
\newblock Clustering of high throughput gene expression data.
\newblock {\em Computers \& operations research}, 39(12):3046--3061, 2012.

\bibitem{nocite4}
Sebastian~J Teran~Hidalgo and Shuangge Ma.
\newblock Clustering multilayer omics data using muncut.
\newblock {\em BMC genomics}, 19(1):1--13, 2018.

\bibitem{nocite5}
Prabhakar Chalise and Brooke~L Fridley.
\newblock Integrative clustering of multi-level ‘omic data based on non-negative matrix factorization algorithm.
\newblock {\em PloS one}, 12(5):e0176278, 2017.

\bibitem{nocite6}
Saket Navlakha and Carl Kingsford.
\newblock The power of protein interaction networks for associating genes with diseases.
\newblock {\em Bioinformatics}, 26(8):1057--1063, 2010.

\bibitem{nocite7}
Elio Masciari, Giuseppe~Massimiliano Mazzeo, and Carlo Zaniolo.
\newblock Analysing microarray expression data through effective clustering.
\newblock {\em Information Sciences}, 262:32--45, 2014.

\bibitem{nocite8}
Diego~H Milone, Georgina Stegmayer, Mariana L{\'o}pez, Laura Kamenetzky, and Fernando Carrari.
\newblock Improving clustering with metabolic pathway data.
\newblock {\em BMC bioinformatics}, 15(1):1--10, 2014.

\bibitem{nocite9}
Deepika Kumar and Usha Batra.
\newblock Clustering algorithms for gene expression data: A review.
\newblock {\em International Journal of Recent Research Aspects}, 4:122--28, 2017.

\bibitem{nocite10}
Shweta Srivastava and Nikita Joshi.
\newblock Clustering techniques analysis for microarray data.
\newblock {\em Int J Comput Sci Mob Comput}, 3:359--364, 2014.

\bibitem{nocite11}
R~Prabahari and V~Thiagarasu.
\newblock Density based clustering using gaussian estimation technique.
\newblock {\em Int J Recent Innovat Trend Comput Commun}, 2:4078--4081, 2014.

\bibitem{nocite12}
Lerato Lerato and Thomas Niesler.
\newblock Clustering acoustic segments using multi-stage agglomerative hierarchical clustering.
\newblock {\em PloS one}, 10(10):e0141756, 2015.

\bibitem{nocite13}
Plamen Angelov, Yannis Manolopoulos, Lazaros Iliadis, Asim Roy, and Marley Vellasco.
\newblock Advances in big data.
\newblock In {\em Proceedings of the 2nd INNS Conference on Big Data}, pages 23--25. Springer, 2016.

\bibitem{nocite14}
M~Sathya Deepa and N~Sujatha.
\newblock Comparative studies of various clustering techniques and its characteristics.
\newblock {\em International Journal of Advanced Networking and Applications}, 5(6):2104, 2014.

\bibitem{nocite15}
Tim Stuart and Rahul Satija.
\newblock Integrative single-cell analysis.
\newblock {\em Nature reviews genetics}, 20(5):257--272, 2019.

\bibitem{nocite16}
Ermelinda Porpiglia, Nikolay Samusik, Andrew Tri~Van Ho, Benjamin~D Cosgrove, Thach Mai, Kara~L Davis, Astraea Jager, Garry~P Nolan, Sean~C Bendall, Wendy~J Fantl, et~al.
\newblock High-resolution myogenic lineage mapping by single-cell mass cytometry.
\newblock {\em Nature cell biology}, 19(5):558--567, 2017.

\end{thebibliography}

\end{document}